%% file: iclr2027_conference.tex
\documentclass{article} 
\usepackage{iclr2027_conference,times}
\iclrfinalcopy
\input{math_commands.tex}

\usepackage{hyperref}
\usepackage{url}
\usepackage{amsthm}
\usepackage{aliascnt}

\title{Nonstationary Multi-Armed Bandits\\with Heterogeneous Arm Classes}

\author{
Orin Levy\thanks{This work was partially conducted while the authors were at Amazon.}\\
Tel Aviv University\\
\And
Noam Touitou$^*$\\
Independent Researcher\\
\And
Aviv Rosenberg$^*$\\
Google Research
}

\usepackage[noorphans,vskip=0ex]{quoting}

\usepackage{xfrac}
\usepackage{microtype}
\usepackage{ifthen}

\PassOptionsToPackage{vlined,ruled,norelsize}{algorithm2e}
\usepackage[algo2e]{algorithm2e}

\SetAlFnt{\small}
\SetAlCapFnt{\small}
\SetAlCapNameFnt{\small}

\LinesNumbered
\DontPrintSemicolon

\usepackage[capitalize,noabbrev]{cleveref}

\RequirePackage{delimset}
\usepackage{stmaryrd}

\def\hideproofs{0}
\usepackage{environ}
\NewEnviron{killcontents}{}
\if 1\hideproofs

\fi

\SetKw{Continue}{continue}

\newcommand{\Prob}{\mathbb{P}}
\newcommand{\Oof}[2][*]{O\brk#1{#2}}

\newcommand{\NumRounds}{T}
\newcommand{\NumArms}{K}
\newcommand{\NumClasses}{M}
\newcommand{\ArmSet}{\mathcal A}
\newcommand{\ClassArms}[1]{\ArmSet_{#1}}
\newcommand{\ChangeBudgetSymbol}{L}

\newcommand{\ChangeRoundsSymbol}{\mathcal L}
\newcommand{\ClassSize}[1]{\NumArms_{#1}}
\newcommand{\ChangeBudget}[1]{\ChangeBudgetSymbol_{#1}}
\newcommand{\ChangeRounds}[1]{\ChangeRoundsSymbol_{#1}}

\newcommand{\MeanRewardSymbol}{\mu}
\newcommand{\ObservedRewardSymbol}{R}
\newcommand{\MeanReward}[2]{\MeanRewardSymbol_{#1}(#2)}
\newcommand{\ObservedReward}[2]{\ObservedRewardSymbol_{#1,#2}}
\newcommand{\PlayedArm}[1]{a_{#1}}
\newcommand{\OptimalArm}[1]{a_{#1}^{*}}
\newcommand{\RepresentativeArm}[1]{\hat a_{#1}}
\newcommand{\RepresentativeOptimalArm}[1]{\hat a_{#1}^{*}}

\newcommand{\BlockingArm}[2]{\bar a_{#1}(#2)}
\newcommand{\ArmCertificateInterval}[2]{I_{#1}(#2)}
\newcommand{\BlockingCertificateInterval}[2]{\overline I_{#1}(#2)}

\newcommand{\NumPullsSymbol}{N}
\newcommand{\EmpMeanSymbol}{\hat{\mu}}
\newcommand{\ConfRadiusSymbol}{\mathrm{CB}}
\newcommand{\NumPulls}[2]{\NumPullsSymbol_{#1}(#2)}
\newcommand{\EmpMean}[2]{\EmpMeanSymbol_{#1}(#2)}
\newcommand{\ConfRadius}[2]{\ConfRadiusSymbol_{#1}(#2)}
\newcommand{\EmpBound}[3]{\EmpMeanSymbol_{#1}^{(#2)}(#3)}
\newcommand{\TaggedNumPulls}[3]{\NumPullsSymbol_{#1}^{#2}(#3)}
\newcommand{\TaggedEmpMean}[3]{\EmpMeanSymbol_{#1}^{#2}(#3)}
\newcommand{\TaggedConfRadius}[3]{\ConfRadiusSymbol_{#1}^{#2}(#3)}
\newcommand{\TaggedEmpBound}[4]{\EmpMeanSymbol_{#1}^{#2,(#3)}(#4)}
\newcommand{\HistoryObservationRounds}[2]{\mathcal O_{#1}^{#2}}

\newcommand{\ArmClassSymbol}{\mathrm{class}}
\newcommand{\PhaseStartSymbol}{\phi}
\newcommand{\GoodArmsSymbol}{\mathcal G}
\newcommand{\BadArmsSymbol}{\mathcal B}
\newcommand{\EligibleArmsSymbol}{\mathcal E}
\newcommand{\BackfillCounterSymbol}{s}
\newcommand{\DominanceGapSymbol}{\Delta}
\newcommand{\DominanceScaleSymbol}{i}
\newcommand{\MaxScale}{J_{\max}}
\newcommand{\LogFactor}{\Lambda}
\newcommand{\ArmClass}[1]{\ArmClassSymbol(#1)}
\newcommand{\PhaseStart}[1]{\PhaseStartSymbol_{#1}}
\newcommand{\GoodArms}[1]{\GoodArmsSymbol_{#1}}
\newcommand{\BadArms}[1]{\BadArmsSymbol_{#1}}
\newcommand{\EligibleArms}[1]{\EligibleArmsSymbol_{#1}}
\newcommand{\BackfillCounter}[2]{\BackfillCounterSymbol_{#1,#2}}
\newcommand{\DominanceGap}[2]{\DominanceGapSymbol_{#1}(#2)}
\newcommand{\DominanceScale}[2]{\DominanceScaleSymbol_{#1}(#2)}
\newcommand{\SelectedClass}[1]{m_{#1}}
\newcommand{\SelectedScale}[1]{j_{#1}}
\newcommand{\PriorityTime}[2]{\tau_{#1}^{#2}}
\newcommand{\CertificateSet}[2]{X_{#1,#2}}

\newcommand{\WidestEligibleArm}[1]{\widetilde a_{#1}}
\newcommand{\TwinArm}[1]{\widetilde a'_{#1}}

\newcommand{\TwinSelectionProbability}{p}
\newcommand{\ArmMSIPairBound}{%
  \sum_{m=1}^{\NumClasses}(2\ChangeBudget{m}+1)\ClassSize{m}}
\newcommand{\DynamicRegret}{\operatorname{Regret}_{\NumRounds}}
\newcommand{\ConcentrationEvent}{\mathcal C}
\newcommand{\HardRounds}{\mathcal U}

\newcommand{\LargeResidualRounds}{\mathcal U_{\mathrm{large}}}

\newcommand{\HardScaleRounds}[2]{\mathcal U_{#1,#2}}
\newcommand{\NoChangeInterval}[2]{J_{#1,#2}}

\newcommand{\ClassBestArm}[2]{a^\star_{#1,#2}}
\newcommand{\BigOTilde}{\widetilde O}
\newcommand{\Indicator}[1]{\mathbf 1\{#1\}}
\newcommand{\ClassDifficultySum}{%
  \sum_{m=1}^{\NumClasses}(\ChangeBudget{m}+1)\ClassSize{m}}

\newcommand{\ServiceTarget}[1]{n_{#1}}
\newcommand{\ClassificationTag}{\mathrm{cls}}
\newcommand{\DetectionTag}{\mathrm{det}}
\newcommand{\ObligationSet}{\mathcal Z}
\newcommand{\ClassificationProposal}[1]{a_{#1}^{\mathrm{cls}}}

\newcommand{\MonitorCandidate}[1]{a_{#1}^{\mathrm M}}
\newcommand{\MonitorClass}[1]{m^{\mathrm M}_{#1}}
\newcommand{\BranchChoice}[1]{\chi_{#1}}
\newcommand{\ClassificationBranch}{\mathrm{cls}}
\newcommand{\MonitoringBranch}{\mathrm{mon}}

\newcommand{\ServiceableObligationSet}[2]{\ObligationSet_{#2}(#1)}
\newcommand{\AdaptiveStoppedRounds}{\mathcal T_{\mathrm A}}
\newcommand{\ServedMonitorRounds}{\mathcal T_{\mathrm M}}
\newcommand{\AdaptiveLaunchProbability}[3]{q_{#1,#2}(#3)}
\newcommand{\AdaptiveEpisodeLaunchProbability}[3]{q_{#1,#2,#3}}
\newcommand{\LaunchSeed}[3]{U_{#1,#2,#3}}
\newcommand{\AdaptiveLaunchIndicator}[3]{B_{#1,#2,#3}}
\newcommand{\UniformUnitInterval}{\operatorname{Unif}[0,1]}
\newcommand{\MonitoringSlotProbability}[1]{p^{\mathrm M}_{#1}}
\newcommand{\DetectedChangeCount}[2]{\widehat{\ChangeBudgetSymbol}_{#1}(#2)}
\newcommand{\DetectedEpisode}[2]{\mathcal P_{#1,#2}}

\newcommand{\AdaptiveStopTime}{\tau_{\mathrm A}}
\newcommand{\PreRoundHistory}[1]{\mathcal F_{#1-1}}
\newcommand{\PreRewardHistory}[1]{\mathcal F_{#1}^{\mathrm{pre}}}

\newcommand{\ClassOptimalRounds}[1]{T_{#1}}
\newcommand{\ClassThreshold}[1]{\zeta_{#1}}
\newcommand{\KnownBudgetOptimalResidual}[1]{%
  \MeanReward{\OptimalArm{#1}}{#1}
  -\EmpBound{\OptimalArm{#1}}{+1}
  {\ArmCertificateInterval{#1}{\OptimalArm{#1}}}}

\newcommand{\CompletionOpportunityCount}[2]{W_{#1,#2}}

\newcommand{\ServiceSuccessEvent}{\mathcal E_{\mathrm{srv}}}

\newcommand{\FrozenHardSequence}[3]{\mathcal H^{\mathrm{aux}}_{#1,#2,#3}}
\newcommand{\CoreOpportunitySequence}[3]{\mathcal H^{\circ}_{#1,#2,#3}}
\newcommand{\CoreWaitingCount}[3]{G_{#1,#2,#3}}

\newcommand{\TotalVariation}[2]{\operatorname{TV}\!\left(#1,#2\right)}
\newcommand{\KLDivergence}[2]{\operatorname{KL}\!\left(#1\middle\|#2\right)}
\newcommand{\BernoulliKL}[2]{\operatorname{kl}\!\left(#1\middle\|#2\right)}

\newcommand{\ClassBlockLength}[1]{b_{#1}}

\newcommand{\TotalDifficulty}{D}

\newtheorem{theorem}{Theorem}[section]

\newaliascnt{proposition}{theorem}
\newtheorem{proposition}[proposition]{Proposition}
\aliascntresetthe{proposition}

\newaliascnt{lemma}{theorem}
\newtheorem{lemma}[lemma]{Lemma}
\aliascntresetthe{lemma}

\newaliascnt{corollary}{theorem}
\newtheorem{corollary}[corollary]{Corollary}
\aliascntresetthe{corollary}

\newaliascnt{definition}{theorem}
\newtheorem{definition}[definition]{Definition}
\aliascntresetthe{definition}

\Crefname{theorem}{Theorem}{Theorems}
\Crefname{proposition}{Proposition}{Propositions}
\Crefname{lemma}{Lemma}{Lemmas}
\Crefname{corollary}{Corollary}{Corollaries}
\Crefname{definition}{Definition}{Definitions}

\Crefname{assumption}{Assumption}{Assumptions}
\Crefname{algorithm}{Algorithm}{Algorithms}

\begin{document}

\maketitle
\lhead{Preprint}

\begin{abstract}
In real-world applications such as online recommendation, nonstationarity is rarely uniform: trending items fluctuate rapidly, while the broader catalog remains stable. 
Existing nonstationary bandit algorithms ignore this structure, incurring regret that scales with the global product of total arms and total switches. 
To address this, we formulate nonstationary multi-armed bandits with heterogeneous arm classes, where $K$ arms are partitioned into $M$ known classes of sizes $K_m$ with individual change budgets $L_m$.
We design an adaptive algorithm requiring no prior knowledge of these budgets that achieves expected dynamic regret of order $\sqrt{MT\sum_m(L_m+1)K_m}$, tightly pairing each class's volatility with only its own size.
Finally, we establish complementary lower bounds showing that this rate is optimal up to logarithmic factors for fixed $M$, and that the $\sqrt{M}$ class-search overhead is fundamentally unavoidable in general.
\end{abstract}

\section{Introduction}
\label{sec:intro}
Multi-armed bandit (MAB) algorithms operating in nonstationary environments provide a foundational framework for sequential decision-making under uncertainty, with widespread applications in online recommendation, ad allocation, and dynamic pricing \citep{garivier2008upper,besbes2014stochastic}. In such domains, the underlying reward distributions of the arms can undergo abrupt switches or continuous drifts over a horizon of $T$ rounds. To protect against arbitrary distribution shifts, the classical literature characterizes nonstationarity through a \emph{global change budget}, typically bounded by $L$ total distribution switches across the entire pool of $K$ arms. Under this formulation, existing minimax-optimal algorithms achieve an expected dynamic regret of $\Theta(\sqrt{(L+1)KT})$ against the sequence of optimal arms at each round \citep{auer2019adaptively, chen2019new}.

Although mathematically elegant, measuring nonstationarity via a single global parameter $L$ overlooks a critical reality: volatility is rarely uniform across actions. In online platforms, a small subset of trending or viral items undergoes rapid shifts in user preference, while the vast majority of the catalog remains essentially stationary. Conventional nonstationary bandit algorithms fail in such asymmetric settings because their change-detection or windowing mechanisms operate globally. A sudden distribution shift in a single viral item triggers a global reset across all arms, needlessly wiping out accumulated observations on stable catalog items.

Consequently, existing algorithms incur dynamic regret that scales with the global product of total arms $K$ and total switches $L$. When frequent changes are confined to a tiny fraction of the action space, this forces the learner to pay for the entire catalog size at the worst-case switching rate. Ideally, an algorithm should isolate nonstationarity to its source, pairing each subset's volatility strictly with its own size. This asymmetry exposes a fundamental theoretical question: \emph{can an online learner achieve regret that decouples localized change budgets from the global arm pool, and what is the unavoidable price of searching across heterogeneous classes?}

To address this challenge, we introduce the framework of \emph{nonstationary multi-armed bandits with heterogeneous arm classes}. Over a horizon of $T$ rounds, a learner interacts with a set of $K$ arms partitioned into $M$ known disjoint classes $\mathcal{A}_1, \dots, \mathcal{A}_M$ with respective sizes $K_m = |\mathcal{A}_m|$. In practical systems, such partitions are readily available from domain metadata, which naturally differentiates volatile items (e.g., breaking news, seasonal promotions) from stable baselines (e.g., evergreen media, staple catalog goods). Rather than imposing a monolithic bound on environmental shifts, each class $\mathcal{A}_m$ is governed by an individual change budget $L_m$, denoting the maximum number of rounds in which reward distributions within class $m$ undergo switches. This formulation captures localized volatility through the cumulative instance difficulty:
$
    D = \sum_{m=1}^M (L_m + 1) K_m.
$
Whenever frequent switches are confined to relatively small subsets of arms, $D$ can be orders of magnitude smaller than the standard global metric $(L+1)K$, establishing a benchmark that reflects the true refined complexity of the environment.

To exploit this structure, we develop \emph{Exponential Backfill for Heterogeneous Arm Classes} (\texttt{EBF-H}). The algorithm is built on two core principles: localized change detection and class-aware backfilling. Rather than resetting globally upon detecting a shift, \texttt{EBF-H} isolates resets strictly to the affected class, preserving sample history for stationary arms. Simultaneously, it uses a careful backfilling schedule to ensure that arms that temporarily appear sub-optimal are re-examined before nonstationarity makes them optimal again. We prove that \texttt{EBF-H} achieves expected dynamic regret of $\BigOTilde ( \sqrt{MTD})$, where $\BigOTilde(\cdot)$ hides logarithmic factors, decoupling the classes up to an explicit factor of $\sqrt{M}$.

We complement this upper bound with two lower bounds confirming its near-optimality. First, a direct-sum bound of $\Omega(\sqrt{TD})$ proves that our rate is optimal up to logarithmic factors whenever $M$ is fixed. Second, an $\Omega(\sqrt{MKT})$ lower bound demonstrates that the $\sqrt{M}$ factor is not an algorithmic artifact, but the fundamental cost of identifying which class currently contains the best arm.

\paragraph{Summary of contributions.}
In summary, this paper provides a near-complete characterization of nonstationary bandits under localized volatility.
We introduce the class-structured nonstationary bandit framework, parameterizing asymmetric volatility via class sizes $K_m$ and change budgets $L_m$, and defining the cumulative instance difficulty $D = \sum_{m=1}^M (L_m + 1)K_m$.
We develop \texttt{EBF-H}, an algorithm achieving expected dynamic regret of  $\BigOTilde(\sqrt{MTD})$ that prevents shifts in volatile classes from contaminating stationary ones. In the supplementary material, we extend this guarantee to an adaptive algorithm (\texttt{AEBF-H}) that requires no prior knowledge of the budgets $\{L_m\}_{m=1}^M$.
Finally, we prove a direct-sum lower bound of $\Omega(\sqrt{TD})$ establishing near-optimality for fixed $M$, alongside an $\Omega(\sqrt{MKT})$ bound confirming the $\sqrt{M}$ factor is the unavoidable price of searching across classes.

\subsection{Related Work}

\paragraph{Global nonstationarity: switching and variation budgets.}
Classical nonstationary bandit formulations characterize environmental shifts through a monolithic global budget, such as total switches $L$ \citep{garivier2008upper, auer2019adaptively, chen2019new} or total variation $V_T$ \citep{besbes2014stochastic}. Because these metrics pool nonstationarity across all actions, a single shifting arm forces algorithms to reset or window observations globally, incurring dynamic regret of order $\sqrt{(L+1)KT}$. Our setting directly complements this line of research: it smoothly recovers the classical switching rate when $M=1$, while offering a structural refinement when nonstationarity is localized, enabling learners to preserve statistical history in stable subsets of arms.

\paragraph{Reward structure vs dynamics structure.}
A rich literature accelerates exploration by exploiting dependencies across arms, including linear and contextual bandits \citep{li2010contextual,abbasi2011improved}, general structured bandits \citep{combes2014unimodal}, and latent or clustered bandits \citep{gentile2014online, maillard2014latent}. These frameworks operate fundamentally in the \emph{reward space}, assuming correlated payoffs or shared latent parameters. Our work explores an orthogonal structural dimension in the \emph{temporal space}: arms within a class share volatility budgets rather than reward values. Because arm means within a class remain independent and unconstrained, this formulation offers a complementary way to incorporate domain structure into online learning.

\paragraph{Specialized temporal dynamics: restless, rotting and corrupted bandits.}
Other important lines of work address structured temporal evolution under specific physical or behavioral dynamics. Restless and rotting bandits \citep{slivkins2008adapting, seznec2020single} model systems governed by continuous stochastic processes (e.g., Brownian motion) or wear-and-tear degradation where rewards diminish with pulls. Similarly, corrupted bandit models \citep{lykouris2018stochastic, gupta2019better} provide robust guarantees for predominantly stationary systems subject to bounded adversarial perturbations. Our piecewise-stationary class model complements these dynamics by capturing abrupt, recurring regime shifts that occur at disparate frequencies across subclasses, without requiring Markovian, monotonic, or small-total-perturbation assumptions.

For additional related work see \Cref{app:related-work}.

\section{Preliminaries}
\label{sec:preliminaries}
\paragraph{Heterogeneous nonstationary bandits.}
We adopt the stochastic switching-bandit setup of
\citet{auer2019adaptively} and introduce a class-structured extension in
which the arms are partitioned into known classes. The online game lasts
$\NumRounds$ rounds and has a finite
set of $\NumArms\ge2$ arms, $\ArmSet=\{1,\ldots,\NumArms\}$. The arms are partitioned
into $\NumClasses$ known, nonempty, disjoint classes
$\ClassArms{1},\ldots,\ClassArms{\NumClasses}$, where
$\bigcup_{m=1}^\NumClasses\ClassArms{m}=\ArmSet$. Class $m$ contains
$\ClassSize{m}:=|\ClassArms{m}|$ arms, so $\NumArms=\sum_{m=1}^\NumClasses \ClassSize{m}$.
For each arm \(a\), \(\ArmClass{a}\) denotes the unique class index \(m\)
such that \(a\in\ClassArms{m}\). Arm $a$ at round $t$
has reward $\ObservedReward{a}{t}\in[0,1]$. Writing \(\E\) for expectation, its mean is
$\MeanReward{a}{t}:=\E[\ObservedReward{a}{t}]$. For each class $m$, define its realized set
of change rounds by
\[
\ChangeRounds{m}
:=\left\{t\in\{2,\ldots,\NumRounds\}:\exists a\in\ClassArms{m}
\text{ such that }\MeanReward{a}{t}\ne\MeanReward{a}{t-1}\right\}.
\]
We assume $|\ChangeRounds{m}|\le \ChangeBudget{m}$ for a class-specific budget
$\ChangeBudget{m}\in\{0,\ldots,\NumRounds-1\}$. A round is counted once for class $m$ whenever
the mean of at least one arm in that class changes, even if several means
change simultaneously. These change rounds partition
$\{1,\ldots,\NumRounds\}$ into
$|\ChangeRounds{m}|+1\le\ChangeBudget{m}+1$ maximal intervals on which every
mean in class $m$ is constant; we call them the \emph{true class-$m$
stationary segments}. Within each maximal interval on which an individual
arm's mean is constant, its rewards are independent and identically
distributed (i.i.d.) from a fixed law; these intervals are the arm's
stationary segments. Thus
$\NumClasses=1$ recovers the classical
switching-bandit model. The budgets may be conservative upper bounds on the
realized numbers of changes.

\paragraph{Online interaction protocol.}
Before interaction begins, the environment fixes the entire mean sequence
$\{\MeanReward{a}{t}\}_{a\in\ArmSet,\ t=1,\ldots,\NumRounds}$. The reward variables are mutually
independent and satisfy $\E[\ObservedReward{a}{t}]=\MeanReward{a}{t}$. At round $t$,
the learner chooses an arm $\PlayedArm{t}$ using
only its internal randomness and the actions and rewards observed before
round $t$, and then observes only $\ObservedReward{\PlayedArm{t}}{t}$. The learner knows $\NumRounds$ and
the class partition. Algorithm \texttt{EBF-H} also receives
$\ChangeBudget{1},\ldots,\ChangeBudget{\NumClasses}$, whereas \texttt{AEBF-H} does not receive these budgets.
Fixing the mean sequence before interaction ensures that it cannot react
to the learner's random choices.
We call any instance satisfying the preceding conditions \emph{admissible}.

\paragraph{Dynamic regret.}
Fix a total order on the arms and choose the first maximizing arm in this
order whenever a tie occurs. Define the optimal arm at round
$t$ by
$
\OptimalArm{t}\in\argmax_{a\in\ArmSet}\MeanReward{a}{t}.
$
The learner's dynamic regret is
\[
\DynamicRegret
:=\sum_{t=1}^\NumRounds\bigl(\MeanReward{\OptimalArm{t}}{t}-\MeanReward{\PlayedArm{t}}{t}\bigr).
\]
Unlike static regret, this benchmark allows the optimal arm to change over
time. Our results bound $\E[\DynamicRegret]$, where the
expectation is over the reward variables and the learner's randomization.

\paragraph{Additional notation.}
For an integer $n\ge0$, write $[n]:=\{1,\ldots,n\}$, with
$[0]:=\emptyset$; for integers
$s\le t$, write $[s,t]:=\{s,s+1,\ldots,t\}$, with
$[t,t-1]:=\emptyset$. $|S|$ denotes the cardinality of  finite set $S$, $\Indicator{E}$ denotes the indicator of event $E$, and \(\Prob\) denotes probability. Unless stated
otherwise, sums indexed by $m$ range over $[\NumClasses]$. Finally, we define the total class
difficulty by
$
\TotalDifficulty:=\ClassDifficultySum.
$

\section{The EBF-H Algorithm and Main Results}
\label{sec:algorithm}
In this section, we present \emph{Exponential Backfill for Heterogeneous Arm Classes} (\texttt{EBF-H}). The primary challenge in heterogeneous nonstationary environments is preventing high-frequency distribution shifts in small, volatile classes from degrading learning progress in stable ones. \texttt{EBF-H} resolves this by decoupling change detection across classes while actively monitoring dormant arms through a multi-scale exponential backfill schedule.

\paragraph{Interval statistics and confidence bounds.}
Because global averages can obscure localized changes by mixing old and new distributions, \texttt{EBF-H} evaluates statistics on sub-intervals within each class's current algorithmic phase. For each class $m \in [\NumClasses]$, the algorithm tracks a phase start round $\PhaseStart{m} \in [\NumRounds]$, initialized to $\PhaseStart{m} = 1$. For an arm $a \in \ClassArms{m}$ and a consecutive sub-interval $I \subseteq [\PhaseStart{m}, t-1]$, we define the pull count $\NumPulls{a}{I} = \sum_{\tau \in I} \Indicator{\PlayedArm{\tau} = a}$, empirical mean $\EmpMean{a}{I} = \frac{\sum_{\tau \in I} \ObservedReward{a}{\tau} \Indicator{\PlayedArm{\tau} = a}}{\max\{\NumPulls{a}{I}, 1\}}$, confidence radius $\ConfRadius{a}{I} = \sqrt{\frac{8 \log(2\NumRounds\NumArms)}{\max\{\NumPulls{a}{I}, 1\}}}$, and empirical bounds $\EmpBound{a}{\pm q}{I} = \EmpMean{a}{I} \pm q\,\ConfRadius{a}{I}$.
With no observations ($I = \emptyset$), we set $\EmpMean{a}{\emptyset} = 0$ and $\ConfRadius{a}{\emptyset} = \sqrt{8\log(2\NumRounds\NumArms)}$.

\paragraph{Mechanism 1: class-isolated change detection.}
At the start of each round $t$, \texttt{EBF-H} tests each class $m \in [\NumClasses]$ for distribution shifts independently. Specifically, it checks whether there exists an arm $a \in \ClassArms{m}$ and consecutive intervals $I_1, I_2 \subseteq [\PhaseStart{m}, t-1]$ whose confidence intervals are disjoint:
\begin{equation}
    \label{eq:reset-condition}
    \EmpBound{a}{-1}{I_1} > \EmpBound{a}{+1}{I_2}.
\end{equation}
If condition~\eqref{eq:reset-condition} is met, a distribution shift is declared in class $m$, triggering an isolated reset: $\PhaseStart{m} \leftarrow t$. Crucially, observations collected for arms in all other classes $m' \neq m$ are strictly preserved, preventing cross-class contamination.

\paragraph{Mechanism 2: arm classification and dominance gaps.}
After evaluating resets, the algorithm classifies each arm $a \in \ArmSet$ as either \emph{good} ($\GoodArms{t}$) or \emph{bad} ($\BadArms{t}$). An arm $a$ is declared bad if another arm $a' \neq a$ provides statistical evidence of superiority across valid phase intervals $I \subseteq [\PhaseStart{\ArmClass{a}}, t-1]$ and $I' \subseteq [\PhaseStart{\ArmClass{a'}}, t-1]$:
\begin{equation}
    \label{eq:bad-condition}
    \EmpBound{a'}{-3}{I'} > \EmpBound{a}{+3}{I}.
\end{equation}
Let $\CertificateSet{t}{a}$ be the set of all triples $(a', I, I')$ satisfying~\eqref{eq:bad-condition}. If $\CertificateSet{t}{a} \neq \emptyset$, $a \in \BadArms{t}$; otherwise, $a \in \GoodArms{t}$. The conservative margin of $3$ times the confidence radii guarantees that the blocking arm in a maximizing certificate,
\begin{equation*}
    (\BlockingArm{t}{a},\ArmCertificateInterval{t}{a},\BlockingCertificateInterval{t}{a}) := \argmax_{(a', I, I') \in \CertificateSet{t}{a}} \left\{ \EmpBound{a'}{-1}{I'} - \EmpBound{a}{+1}{I} \right\},
\end{equation*}
is provably good (\Cref{prop:blocking-is-good}). We record this maximum gap as the \emph{dominance gap} $\DominanceGap{t}{a} := \max_{(a', I, I') \in \CertificateSet{t}{a}} \{ \EmpBound{a'}{-1}{I'} - \EmpBound{a}{+1}{I} \}$ and define the log scale $\DominanceScale{t}{a} := \lceil -\log_2 \DominanceGap{t}{a} \rceil$.

\paragraph{Mechanism 3: Class-Aware Exponential Backfill.}
When an arm's reward distribution changes, it may secretly become optimal while outdated observations from before the shift continue to certify it as bad. If the learner completely ignored bad arms, it would never collect fresh samples to detect this improvement. Conversely, pulling bad arms indiscriminately would incur intolerable regret. To balance this trade-off, \texttt{EBF-H} uses \emph{exponential backfill}, maintaining a set of monitoring counters $\BackfillCounter{a}{j}$ that act as allocated quotas of exploratory pulls.

Each counter $\BackfillCounter{a}{j}$ corresponds to a target gap resolution $2^{-j}$ indexed by log scales $j \in \{0, \dots, \MaxScale\}$, where $\MaxScale := \lceil \log_2 \NumRounds \rceil$. A positive counter value $\BackfillCounter{a}{j} > 0$ represents a pending request to inspect arm $a$ with enough samples to resolve a gap of size $2^{-j}$. At each round, each counter is refreshed to a full quota of $2^{2j+5}\LogFactor$ pulls (where $\LogFactor := \lceil \log(2\NumRounds\NumArms) \rceil$) with independent probability:
\begin{equation*}
    2^{-j} \sqrt{\frac{\ChangeBudget{\ArmClass{a}} + 1}{\NumRounds \ClassSize{\ArmClass{a}}}}.
\end{equation*}
This refresh rule captures the core statistical trade-off of monitoring:
\begin{itemize}
    \item \emph{Resolving finer gaps requires more pulls:} By Hoeffding's inequality, distinguishing a gap of size $2^{-j}$ requires $O(2^{2j})$ observations, which is why the allocated quota quadruples with each increase in scale $j$.
    \item \emph{Finer checks are launched less frequently:} To keep the expected monitoring regret bounded, the probability of initiating a check at scale $j$ halves with each step ($2^{-j}$).
    \item \emph{Rates are calibrated to class volatility:} The refresh probability scales with $\sqrt{(\ChangeBudget{m}+1)/\ClassSize{m}}$, ensuring that arms in volatile classes are monitored frequently while arms in stable classes are queried sparingly.
\end{itemize}

\paragraph{Action Selection Policy.}
Armed with the classified arms and monitoring quotas, the learner decides which arm to sample at round $t$. This process involves three stages:

\begin{enumerate}
    \item \textbf{Forming the eligible pool:} All good arms ($\GoodArms{t}$) have not been ruled out by evidence and are always eligible for selection. In contrast, a bad arm $a \in \BadArms{t}$ is eligible only if it has an active monitoring request ($\BackfillCounter{a}{j} > 0$) at a scale no finer than its estimated gap ($j \le \DominanceScale{t}{a}$):
    \begin{equation}
        \label{eq:eligible-arms}
        \EligibleArms{t} := \GoodArms{t} \cup \big\{a \in \BadArms{t} : \exists j \le \DominanceScale{t}{a} \text{ such that } \BackfillCounter{a}{j} > 0\big\}.
    \end{equation}
    The constraint $j \le \DominanceScale{t}{a}$ is crucial: it prevents a large quota intended for a very subtle gap (large $j$) from being wasted on an arm with a massive performance deficit (small $\DominanceScale{t}{a}$).

    \item \textbf{Class selection:} The learner first samples a class $\SelectedClass{t}$ uniformly at random among all classes that contain at least one eligible arm. Drawing classes uniformly prevents large catalog classes from drowning out small, highly volatile classes, guaranteeing that volatile arms receive sufficient monitoring opportunities regardless of catalog size.

    \item \textbf{Arm selection:} Within the selected class $\SelectedClass{t}$, the learner chooses which eligible arm to pull. Rather than pulling greedily, the algorithm samples a random scale $\SelectedScale{t} \in \{0, \dots, \MaxScale\}$ and identifies how recently each candidate received its $2^{\SelectedScale{t}}$-th most recent pull in the current phase, denoted $\PriorityTime{a}{\SelectedScale{t}}$ (setting $\PriorityTime{a}{\SelectedScale{t}} = -\infty$ if it has fewer pulls). It then plays the arm whose observations are least recent:
    $
        \PlayedArm{t} \leftarrow \argmin_{a \in \ClassArms{\SelectedClass{t}} \cap \EligibleArms{t}} \PriorityTime{a}{\SelectedScale{t}}.
    $
    This priority mechanism operates like a multi-scale round-robin, ensuring that no eligible arm is neglected and that confidence intervals shrink evenly across the class.
\end{enumerate}

\paragraph{Updating Counters.} After playing $\PlayedArm{t}$ and observing its reward $\ObservedReward{\PlayedArm{t}}{t}$, the algorithm updates its empirical records. If the played arm was bad, it decrements its applicable active counters $\BackfillCounter{\PlayedArm{t}}{j}$ by one for all $0 \le j \le \min\{\DominanceScale{t}{\PlayedArm{t}}, \MaxScale\}$, gradually exhausting its monitoring quota until the check concludes. Pulls of good arms leave the counters unchanged.

\begin{algorithm2e}[t]
\caption{\texttt{EBF-H}: Exponential Backfill for Heterogeneous Arm Classes}
\label{alg:ebf-h}
\SetAlgoLined
\DontPrintSemicolon
\KwIn{Horizon $\NumRounds$, classes $\ClassArms{1}, \dots, \ClassArms{\NumClasses}$, budgets $\ChangeBudget{1}, \dots, \ChangeBudget{\NumClasses}$.}
Set $\MaxScale \leftarrow \lceil \log_2 \NumRounds \rceil$, $\LogFactor \leftarrow \lceil \log(2\NumRounds\NumArms) \rceil$, $\PhaseStart{m} \leftarrow 1$ for all $m \in [\NumClasses]$, and $\BackfillCounter{a}{j} \leftarrow 0$.\;
\For{$t = 1, \dots, \NumRounds$}{
    \tcp{1. Class-Isolated Change Detection}
    \ForEach{$m \in [\NumClasses]$}{
        \If{$\exists a \in \ClassArms{m}, I_1, I_2 \subseteq [\PhaseStart{m}, t-1]$ s.t.\ $\EmpBound{a}{-1}{I_1} > \EmpBound{a}{+1}{I_2}$}{
            $\PhaseStart{m} \leftarrow t$.\;
        }
    }
    \tcp{2. Arm Classification (Details in Appendix~\ref{app:additional-algorithms}, Algorithm~\ref{alg:classify})}
    $(\GoodArms{t}, \BadArms{t}, \{\DominanceGap{t}{a}, \DominanceScale{t}{a}\}_{a \in \BadArms{t}}) \leftarrow \texttt{ClassifyArms}(t, \PhaseStart{1}, \dots, \PhaseStart{\NumClasses})$.\;
    \tcp{3. Exponential Backfill Refresh}
    \ForEach{$a \in \ArmSet, j \in \{0, \dots, \MaxScale\}$}{
        With prob.\ $2^{-j}\sqrt{(\ChangeBudget{\ArmClass{a}}+1)/(\NumRounds\ClassSize{\ArmClass{a}})}$, set $\BackfillCounter{a}{j} \leftarrow 2^{2j+5}\LogFactor$.\;
    }
    \tcp{4. Two-Stage Action Selection}
    $\EligibleArms{t} \leftarrow \GoodArms{t} \cup \{a \in \BadArms{t} : \exists j \le \DominanceScale{t}{a} \text{ s.t. } \BackfillCounter{a}{j} > 0\}$.\;
    Draw $\SelectedClass{t}$ uniformly from $\{m : \ClassArms{m} \cap \EligibleArms{t} \neq \emptyset\}$; draw $\SelectedScale{t}$ uniformly from $\{0, \dots, \MaxScale\}$.\;
    Play $\PlayedArm{t} \leftarrow \argmin_{a \in \ClassArms{\SelectedClass{t}} \cap \EligibleArms{t}} \PriorityTime{a}{\SelectedScale{t}}$, observe $\ObservedReward{\PlayedArm{t}}{t}$, and update histories.\;
    \If{$\PlayedArm{t} \in \BadArms{t}$}{
        \ForEach{$j \in \{0, \dots, \min\{\DominanceScale{t}{\PlayedArm{t}}, \MaxScale\}\}$}{
            $\BackfillCounter{\PlayedArm{t}}{j} \leftarrow \max\{0, \BackfillCounter{\PlayedArm{t}}{j} - 1\}$.\;
        }
    }
}
\end{algorithm2e}

The complete execution of \texttt{EBF-H} is summarized in Algorithm~\ref{alg:ebf-h}. By maintaining prefix sums of pull counts and rewards, each interval statistic is computable in $O(1)$ time, yielding an overall runtime of $O(\NumArms^2 \NumRounds^5)$ and $O(\NumArms \NumRounds)$ memory.  The following theorem establishes our main dynamic regret guarantee.

\begin{theorem}[Dynamic regret bound of \texttt{EBF-H}]
\label{thm:main-formal}
For every admissible nonstationary environment, the expected dynamic regret of \texttt{EBF-H} satisfies
\begin{equation*}
    \E[\DynamicRegret] \le 4096 \log^2(2\NumRounds\NumArms) \sqrt{\NumClasses \NumRounds \TotalDifficulty} + 1 = \BigOTilde(\sqrt{\NumClasses \NumRounds \TotalDifficulty}).
\end{equation*}
\end{theorem}

\paragraph{Extension to unknown budgets.}
While \texttt{EBF-H} takes the budgets $\{\ChangeBudget{m}\}_{m=1}^\NumClasses$ as input to calibrate its backfill probabilities, in many realistic settings these volatility budgets are unknown. In Appendix~\ref{app:adaptive-extension}, we show that this requirement can be completely eliminated. We develop \emph{Adaptive Exponential Backfill for Heterogeneous Arm Classes} (\texttt{AEBF-H}), an algorithm that takes only the horizon $\NumRounds$ and the class partition as inputs. Rather than relying on a static budget $\ChangeBudget{m}$, \texttt{AEBF-H} dynamically scales the monitoring rate of class $m$ based on the empirical number of resets detected in that class up to the current round. The proof separates classification and monitoring observations and chooses monitoring candidates independently of obligations. This preserves launch-seed independence. Bounding launch costs by detected resets and summing waiting times across detected phases gives the identical leading dynamic regret bound:
\begin{equation*}
    \E[\DynamicRegret] = \BigOTilde(\sqrt{\NumClasses \NumRounds \TotalDifficulty}).
\end{equation*}
To keep our main presentation focused on the core statistical mechanics, we concentrate on \texttt{EBF-H} in the main text and provide the full pseudocode and analysis of \texttt{AEBF-H} in Appendix~\ref{app:adaptive-extension}.

\section{Dynamic Regret Analysis}
\label{sec:analysis}

In this section, we prove Theorem~\ref{thm:main-formal} by developing the proof architecture for \texttt{EBF-H}. We establish that the algorithm's localized resets strictly isolate nonstationarity, and we decompose the dynamic regret into three distinct statistical penalties. Full proofs of all lemmas are deferred to Appendix~\ref{app:section4-proofs}.

\subsection{Foundations: Concentration, Local Resets, and Representatives}
We begin by establishing uniform confidence bounds and bounding the number of class resets.
For a stationary interval \(I\), let \(\mu_a(I)\) denote the common mean of arm \(a\) on \(I\). The following lemma gives a uniform concentration bound
over all such intervals.
\begin{lemma}[Uniform interval concentration]
\label{lem:confidence-bounds}
Let $\ConcentrationEvent$ be the presampled-block event of Appendix~\ref{app:proof-confidence-bounds}, independent of the algorithm's coins. Then $\Prob(\ConcentrationEvent)\ge1-\NumRounds^{-1}$, and on $\ConcentrationEvent$, $|\EmpMean{a}{I}-\mu_a(I)|\le\ConfRadius{a}{I}$ for every arm $a$ and nonempty interval $I$ within one stationary segment.
\end{lemma}

\Cref{lem:confidence-bounds} follows from Hoeffding's inequality applied to pre-sampled rewards combined with a union bound over all fixed sequence blocks. On $\ConcentrationEvent$, the algorithm's change-detection test is strictly conservative:

\begin{lemma}[Resets are paid by true changes]
\label{lem:phase-resets}
On the concentration event $\ConcentrationEvent$, class $m$ is reset at most $\ChangeBudget{m}$ times, dividing the horizon for class $m$ into at most $\ChangeBudget{m} + 1$ algorithmic phases.
\end{lemma}

If no change occurs in class $m$ during its active phase, the true mean remains identical across all sub-intervals, so their confidence intervals must overlap on $\ConcentrationEvent$. Hence, every reset requires a genuine distribution shift in class $m$.
Next, recall that each bad arm $a \in \BadArms{t}$ is associated with a blocking arm $\BlockingArm{t}{a}$ maximizing the certificate separation. The following structural result ensures that our reference arm is always trustworthy:

\begin{proposition}[Blocking arms are good]
\label{prop:blocking-is-good}
For every round $t \in [\NumRounds]$ and every arm $a \in \BadArms{t}$, the blocking arm satisfies $\BlockingArm{t}{a} \in \GoodArms{t}$.
\end{proposition}

If $\BlockingArm{t}{a}$ were bad, another arm would certify it as inferior; the $3\times$ radius margin in~\eqref{eq:bad-condition} would then imply that this third arm certifies $a$ with a strictly larger gap, contradicting the maximality of $\BlockingArm{t}{a}$.

\subsection{The Three-Term Regret Decomposition}
\label{def:representatives}
Writing $\mu^*(t):=\MeanReward{\OptimalArm{t}}{t}$, we assign a \emph{representative} to both the played arm $\PlayedArm{t}$ and the instantaneous optimal arm $\OptimalArm{t}$:
\begin{equation*}
    \RepresentativeArm{t} := \begin{cases} \PlayedArm{t} & \text{if } \PlayedArm{t} \in \GoodArms{t}, \\ \BlockingArm{t}{\PlayedArm{t}} & \text{if } \PlayedArm{t} \in \BadArms{t}, \end{cases}
    \qquad
    \RepresentativeOptimalArm{t} := \begin{cases} \OptimalArm{t} & \text{if } \OptimalArm{t} \in \GoodArms{t}, \\ \BlockingArm{t}{\OptimalArm{t}} & \text{if } \OptimalArm{t} \in \BadArms{t}. \end{cases}
\end{equation*}
By Proposition~\ref{prop:blocking-is-good}, both $\RepresentativeArm{t}$ and $\RepresentativeOptimalArm{t}$ are guaranteed to be good arms. Adding and subtracting their expected rewards yields the exact instantaneous decomposition:
\begin{equation}
    \label{eq:regret-decomposition}
    \mu^*(t) - \mu_{\PlayedArm{t}}(t) = 
    \underbrace{\mu^*(t) - \mu_{\RepresentativeOptimalArm{t}}(t)}_{\text{\bf Term (iii): delayed detection}} + 
    \underbrace{\mu_{\RepresentativeOptimalArm{t}}(t) - \mu_{\RepresentativeArm{t}}(t)}_{\text{\bf Term (ii): comparison good arms}} + 
    \underbrace{\mu_{\RepresentativeArm{t}}(t) - \mu_{\PlayedArm{t}}(t).}_{\text{\bf Term (i): monitoring bad arms}}
\end{equation}
The terms can be signed; each isolates a distinct statistical challenge:
\begin{itemize}
    \item \textbf{Term (i) (monitoring bad arms):} Measures the cost of exploring an arm currently labeled bad. It is zero when $\PlayedArm{t} \in \GoodArms{t}$, and otherwise bounded by its dominance gap and confidence corrections.
    \item \textbf{Term (ii) (comparing good arms):} Measures the regret between two good representatives. Because neither arm has been ruled out by evidence, their gap is controlled by their remaining confidence radii.
    \item \textbf{Term (iii) (delayed detection):} Measures the loss when the optimal arm $\OptimalArm{t}$ is temporarily misclassified as bad due to outdated observations from before a recent change.
\end{itemize}

\subsection{Controlling the Regret Components}
\label{sec:controlling-terms}

We bound the sums conditional on $\ConcentrationEvent$. Here $\ConfRadius{a}{t}$ uses observations before $t$ since the latest true change or reset of $a$'s class (its current MSI).

\paragraph{Term (i): monitoring bad arms.}
Term~(i) is zero for a good played arm. Otherwise, the certificate and no-reset comparisons give $\mu_{\RepresentativeArm{t}}(t)-\mu_{\PlayedArm{t}}(t)\le2\ConfRadius{\RepresentativeArm{t}}{t}+3\DominanceGap{t}{\PlayedArm{t}}+2\ConfRadius{\PlayedArm{t}}{t}$.

\begin{lemma}[Bounding Term (i)]
\label{lem:term-i}
On the concentration event $\ConcentrationEvent$, the cumulative cost of monitoring bad arms satisfies:
$
    \E\Big[ \sum_{t=1}^\NumRounds \big(\mu_{\RepresentativeArm{t}}(t) - \mu_{\PlayedArm{t}}(t)\big) \mid \ConcentrationEvent \Big] \le \BigOTilde(\sqrt{\NumClasses \NumRounds \TotalDifficulty}).
$
\end{lemma}

\noindent\textbf{Proof intuition.}
Every bad-arm pull charges $3\DominanceGap{t}{\PlayedArm{t}}\le6\cdot2^{-j}$ to the latest refresh of one active counter $\BackfillCounter{\PlayedArm{t}}{j}$ with $j\le\DominanceScale{t}{\PlayedArm{t}}$. A scale-$j$ refresh receives at most $2^{2j+5}\LogFactor$ such charges, costing $O(2^j\LogFactor)$. Multiplying by its refresh probability $2^{-j}\sqrt{(\ChangeBudget{m}+1)/(\NumRounds\ClassSize{m})}$ cancels $2^j$. Summing across scales, arms, and rounds bounds the dominance-gap contribution. Eligible-arm and played-arm confidence sums control the two remaining terms (Appendix~\ref{app:proof-term-i}).

\paragraph{Term (ii): comparing good arms.}
Term~(ii) compares the two good representatives $\RepresentativeOptimalArm{t}$ and $\RepresentativeArm{t}$.

\begin{lemma}[Bounding Term (ii)]
\label{lem:term-ii}
On the concentration event $\ConcentrationEvent$, the cumulative regret between good representatives satisfies:
$
    \E\Big[ \sum_{t=1}^\NumRounds \big(\mu_{\RepresentativeOptimalArm{t}}(t) - \mu_{\RepresentativeArm{t}}(t)\big) \mid \ConcentrationEvent \Big] \le \BigOTilde (\sqrt{\NumClasses \NumRounds \TotalDifficulty}).
$
\end{lemma}

\noindent\textbf{Proof intuition.}
Because both representatives remain good, the three-radius certificate test and concentration on their current MSI histories give $\mu_{\RepresentativeOptimalArm{t}}(t)-\mu_{\RepresentativeArm{t}}(t)\le4\ConfRadius{\RepresentativeOptimalArm{t}}{t}+4\ConfRadius{\RepresentativeArm{t}}{t}$. Summing these confidence radii over the horizon is controlled by the rate at which eligible arms are sampled under our two-stage policy (Appendix~\ref{app:proof-term-ii}).

\paragraph{Term (iii): delayed change detection.}
Term~(iii) is non-zero only when the optimal arm $\OptimalArm{t}$ is misclassified as bad due to stale observations.

\begin{lemma}[Bounding Term (iii)]
\label{lem:term-iii}
On the concentration event $\ConcentrationEvent$, the cumulative regret from delayed detection satisfies:
$
    \E\Big[ \sum_{t=1}^\NumRounds \big(\mu^*(t) - \mu_{\RepresentativeOptimalArm{t}}(t)\big) \mid \ConcentrationEvent \Big] \le \BigOTilde(\sqrt{\NumClasses \NumRounds \TotalDifficulty}).
$
\end{lemma}

\noindent\textbf{Proof intuition.}
For a bad optimal arm, use the certificate residual $r_t:=\KnownBudgetOptimalResidual{t}$, not its regret gap. The no-reset comparison gives Term~(iii)$\le r_t-\DominanceGap{t}{\OptimalArm{t}}+2\ConfRadius{\RepresentativeOptimalArm{t}}{t}$. After bounding the confidence and class-threshold terms, consider hard rounds with $r_t>\DominanceGap{t}{\OptimalArm{t}}$ and $2^{-j}<r_t\le2^{1-j}$. Then $j\le\DominanceScale{t}{\OptimalArm{t}}$. On a fixed true class-$m$ stationary segment, the wait for its fixed best arm's first scale-$j$ refresh has expectation at most $2^j\sqrt{\NumRounds\ClassSize{m}/(\ChangeBudget{m}+1)}$. Multiplication by $2^{1-j}$ bounds the waiting contribution. After refresh, exhausting the counter on such a hard round would force a reset; otherwise the arm is eligible and $r_t\le2\ConfRadius{\OptimalArm{t}}{t}$. A reset eliminates positive residuals for the rest of that segment. Summing over segments and scales, and adding at most one for residuals below $2^{-\MaxScale}$, proves the claim (Appendix~\ref{app:proof-term-iii}).

\subsection{Synthesis: Summing over Maximal Static Intervals}
\label{sec:synthesis-msi}

To complete the proof of Theorem~\ref{thm:main-formal}, we bound the cumulative sum of confidence radii across the horizon. For each class $m \in [\NumClasses]$, we partition the horizon into \emph{Maximal Static Intervals} (MSIs). A new MSI begins whenever: the reward distribution of some arm in $\ClassArms{m}$ changes (at most $\ChangeBudget{m}$ times), or class $m$ undergoes a change-detection reset (at most $\ChangeBudget{m}$ times on $\ConcentrationEvent$ by Lemma~\ref{lem:phase-resets}).
Thus, class $m$ is partitioned into at most $2\ChangeBudget{m} + 1$ MSIs, during which both the true means and the phase start time $\PhaseStart{m}$ remain completely fixed. Across all $M$ classes, the total number of arm-interval pairs $(a, \mathcal{I})$ is bounded by:
\begin{equation*}
    \sum_{m=1}^\NumClasses (2\ChangeBudget{m} + 1) \ClassSize{m} \le 2 \sum_{m=1}^\NumClasses (\ChangeBudget{m} + 1) \ClassSize{m} = 2\TotalDifficulty.
\end{equation*}

Within each MSI $\mathcal{I}$ where arm $a$ is pulled $n_{a, \mathcal{I}}$ times. Standard confidence-sum bounds \citep{lattimore2020bandit} guarantee that the sum of radii over pulls is bounded by $O(\sqrt{\log(\NumRounds\NumArms) n_{a, \mathcal{I}}})$. Applying the Cauchy--Schwarz inequality across all arm-interval pairs yields:
\begin{equation*}
    \sum_{m=1}^\NumClasses \sum_{\mathcal{I}} \sum_{a \in \ClassArms{m}} \sqrt{n_{a, \mathcal{I}}} \le \sqrt{\left(\sum_{m=1}^\NumClasses (2\ChangeBudget{m}+1)\ClassSize{m}\right) \left(\sum_{m=1}^\NumClasses \sum_{\mathcal{I}} \sum_{a \in \ClassArms{m}} n_{a, \mathcal{I}}\right)} \le \sqrt{2\TotalDifficulty \cdot \NumRounds}.
\end{equation*}

\paragraph{Origin of the $\sqrt{\NumClasses}$ overhead.}
For the widest-confidence eligible arm, the recency rule supplies a same-class comparison arm of at least half its radius, played with probability at least $p=1/[\NumClasses(\MaxScale+1)]$. Each arm--MSI pair advances from a given pull count only once. With at most $2\TotalDifficulty$ pairs, the expected number of rounds at any comparison-arm count is at most $2\TotalDifficulty/p$. Splitting the confidence sum at a count threshold gives $O(\sqrt{\TotalDifficulty\NumRounds\log(2\NumRounds\NumArms)/p})$ (\Cref{lem:proof-eligible-cb-sums}), yielding the final bound:
$
    \E[\DynamicRegret \mid \ConcentrationEvent] = \BigOTilde(\sqrt{\NumClasses \NumRounds \TotalDifficulty}).
$

Since $\Prob(\ConcentrationEvent^c) \le \NumRounds^{-1}$ and regret is bounded by $\NumRounds$, the failure event contributes at most $\NumRounds \cdot \NumRounds^{-1} = 1$ in expectation, concluding the proof of Theorem~\ref{thm:main-formal}.

\section{Lower Bound}
\label{sec:lower-bounds}

In this section, we establish the fundamental information-theoretic limits of nonstationary bandits with heterogeneous arm classes. Our upper bound reflects two distinct statistical challenges:
\begin{enumerate}
    \item \emph{Within-class re-learning:} The sample cost of identifying newly optimal arms within each class following localized distribution shifts.
    \item \emph{Cross-class search:} The exploration overhead of identifying which of the $M$ classes currently contains the globally optimal arm.
\end{enumerate}
Below, we present two complementary lower bounds that isolate these two phenomena. Both results hold for randomized learners, even when the algorithm has full prior knowledge of the class budgets $\{L_m\}_{m=1}^M$.

\paragraph{Within-class difficulty: the direct-sum bound.}
Our first result applies to arbitrary class configurations and demonstrates that the cumulative instance difficulty $D$ cannot be circumvented.

\begin{theorem}[Direct-Sum Lower Bound]
\label{thm:direct_sum_lower_bound}
Fix any collection of class sizes $\{K_m\}_{m=1}^M$ and change budgets $\{L_m\}_{m=1}^M$. If the horizon satisfies $T \ge 2D$, then for any policy, there exists an admissible nonstationary environment such that
$
    \mathbb{E}[\DynamicRegret] \ge \frac{1}{256} \sqrt{T D}.
$
\end{theorem}

\noindent\textbf{Intuition.}
Theorem~\ref{thm:direct_sum_lower_bound} establishes that localized nonstationarity is fundamentally additive: an algorithm cannot amortize the cost of changes in one class using samples gathered from another. If $D\le4K$, a stationary $K$-arm instance suffices. Otherwise, class $m$ receives $\lfloor L_m/2\rfloor$ independent learning blocks, each bracketed by baseline rounds and using two changes; their total weight satisfies $\sum_m\lfloor L_m/2\rfloor K_m\ge D/4$. By tuning the sub-interval lengths proportionally to class sizes and calibrating the mean gaps to bound the Kullback--Leibler (KL) divergence of the history, Pinsker's inequality prevents the learner from reliably identifying the best arm in each epoch, yielding the additive rate $\sqrt{TD}$.

\paragraph{Cross-class exploration: the price of class search.}
While Theorem~\ref{thm:direct_sum_lower_bound} captures within-class learning, it does not account for the uncertainty of locating an active class when optimal arms migrate across the partition. Our second result isolates this cross-class search cost in balanced environments with two changes per class.

\begin{theorem}[Class-Search Lower Bound]
\label{thm:class_search_lower_bound}
Let $K$ be divisible by $M$, and suppose all classes are balanced with $K_m = K/M$ with change budgets $L_m = 2$ for all $m \in [M]$. If $T \ge 6MK$, then for any policy, there exists an admissible environment such that
$
    \mathbb{E}[\DynamicRegret] \ge \frac{1}{512} \sqrt{MKT}.
$
\end{theorem}

\noindent\textbf{Intuition.}
In this construction, $M$ active macro-epochs lie between initial and terminal baseline blocks. In each epoch, a single class is chosen according to a uniformly random latent permutation of $[M]$, and a single arm within that class is temporarily elevated above the baseline. Even if an oracle reveals the active class and optimal arm in hindsight at the end of each macro-epoch, the identities of all future active classes remain uniformly distributed among the unvisited classes. Consequently, the learner is forced to repeatedly search over $\Theta(M)$ candidate classes, incurring an unavoidable $\sqrt{M}$ exploration penalty that scales the regret up to $\Omega(\sqrt{MKT})$.

\paragraph{Tightness and Minimax Implications.}
Together, Theorems~\ref{thm:direct_sum_lower_bound} and~\ref{thm:class_search_lower_bound} provide a comprehensive characterization of the minimax regret landscape:
\begin{itemize}
    \item \textbf{Near-optimality for fixed $M$:} For any fixed number of classes ($M = O(1)$), our upper bound $\BigOTilde(\sqrt{MTD})$ matches the direct-sum lower bound $\Omega(\sqrt{TD})$ up to logarithmic factors across all horizon regimes $T \ge 2D$.
    \item \textbf{Necessity of the $\sqrt{M}$ class-search penalty:} In the balanced setting of Theorem~\ref{thm:class_search_lower_bound}, the difficulty parameter evaluates to $D = \sum_{m=1}^M (2+1)(K/M) = 3K$. On this family of instances, our upper bound evaluates to $\BigOTilde(\sqrt{M K T})$, precisely matching the lower bound of Theorem~\ref{thm:class_search_lower_bound} up to logarithmic factors. This resolves the status of the $\sqrt{M}$ factor: it is not an algorithmic defect of \texttt{EBF-H}, but an unavoidable class-search cost on this balanced family.
\end{itemize}

\paragraph{Proof Architecture.}
Both lower bounds rely on a standard information-theoretic change-of-measure argument for bandits \citep{lattimore2020bandit}. We construct a prior distribution over candidate mean vectors and evaluate the trajectory of the learner under a stationary null baseline versus alternative hypotheses where a specific candidate arm is shifted by $\Delta$. The KL divergence between the resulting path measures is bounded by the expected number of pulls allocated to the perturbed arm. In the direct-sum bound, independence across classes allows the conditional regrets to sum additively across blocks via the tower property of expectation. In the class-search bound, the uniform prior over future class permutations ensures that the conditional indistinguishability holds at every macro-epoch, regardless of past feedback. Complete proofs are provided in Appendix~\ref{app:lower-bound-details}.

\section{Discussion and Future Work}
\label{sec:conclusions}
\label{sec:discussion}
This work introduced nonstationary multi-armed bandits with heterogeneous arm classes to model environments with localized volatility. We proposed \texttt{EBF-H}, an algorithm achieving expected dynamic regret of order $\sqrt{MTD}$ by confining distribution shifts strictly to their source classes. Complementary lower bounds establish that this guarantee is optimal up to logarithmic factors for fixed $M$, while confirming that $\sqrt{M}$ factor is an unavoidable cost of searching across classes.

Several compelling directions remain open. First, while our upper and lower bounds match for fixed $M$, pinning down the exact minimax rate when $M$ grows large in highly imbalanced regimes remains an open theoretical question. Second, relaxing the assumption of a known partition to learn latent volatility classes online presents a significant challenge. Third, designing lighter change-detection approximations, such as grid-based or sliding-window change-detection tests, would substantially reduce per-round computational overhead for practical large-scale deployments. Finally, extending localized volatility to linear and contextual bandits offers a natural path toward broader real-world applications.

\subsection*{AI use statement}
Generative AI assistance was used to check and develop mathematical arguments,
verify references, and draft and revise the exposition. The authors are
responsible for the final content.

\section*{Acknowledgment}
This project is supported by the European Research Council (ERC) under the European Union’s
Horizon 2020 research and innovation program (grant agreement No. 882396), by the Israel Science
Foundation (1357/24), the Blavatnik foundation at Tel Aviv University and by a grant from the
Tel Aviv University Center for AI and Data Science (TAD). 
OL is also supported by the Google PhD Fellowship Award
(2025).

\bibliography{iclr2027_conference}
\bibliographystyle{iclr2027_conference}

\clearpage
\appendix

\section{Additional Algorithms}
\label{app:additional-algorithms}

\begin{algorithm2e}[H]
\caption{\texttt{ClassifyArms}$(t,\PhaseStart{1},\ldots,
\PhaseStart{\NumClasses})$\label{alg:classify}}
\SetAlgoLined
\DontPrintSemicolon

\KwIn{Current time $t$, phase start times $\PhaseStart{1},\ldots,
\PhaseStart{\NumClasses}$, and interval statistics from
Section~\ref{sec:algorithm} computed from the rewards before \(t\).}
\KwOut{Good arms $\GoodArms{t}$, bad arms $\BadArms{t}$, and each bad arm's
blocking arm, certificate intervals, dominance gap, and scale.}

Initialize $\BadArms{t} \leftarrow \emptyset$
\tcp{All intervals below are consecutive round intervals; all ties use the fixed rule}

\ForEach{arm $a \in \ArmSet$}{
    $\CertificateSet{t}{a} \leftarrow \{(a', I, I') :
    a' \in \ArmSet\setminus\{a\},
    I \subseteq [\PhaseStart{\ArmClass{a}}, t-1],
    I' \subseteq [\PhaseStart{\ArmClass{a'}}, t-1],
    \EmpBound{a'}{-3}{I'} > \EmpBound{a}{+3}{I}\}$

    \If{$\CertificateSet{t}{a} = \emptyset$}{
        \Continue \tcp*{$a$ is good}
    }

    $\BadArms{t} \leftarrow \BadArms{t} \cup \{a\}$

    $(\BlockingArm{t}{a},\ArmCertificateInterval{t}{a},
    \BlockingCertificateInterval{t}{a}) \leftarrow
    \argmax_{(a',I,I') \in \CertificateSet{t}{a}}
    \left\{\EmpBound{a'}{-1}{I'}-\EmpBound{a}{+1}{I}\right\}$

    $\DominanceGap{t}{a} \leftarrow
    \max_{(a',I,I')\in\CertificateSet{t}{a}}
    \{\EmpBound{a'}{-1}{I'}-\EmpBound{a}{+1}{I}\}$

    $\DominanceScale{t}{a} \leftarrow
    \lceil-\log_2(\DominanceGap{t}{a})\rceil$
}

$\GoodArms{t} \leftarrow \ArmSet \setminus \BadArms{t}$

\Return the good and bad arms and the stored certificates, gaps, and scales
\end{algorithm2e}


\section{Proofs for \Cref{sec:algorithm,sec:analysis} }
\label{app:ebfh-dynamic-regret}
\label{app:section3-proofs}
\label{app:section4-proofs}
\label{app:ebf-details}

Our goal is to prove \Cref{thm:main-formal}, the dynamic-regret upper bound
for Algorithm~\texttt{EBF-H} proving that 
\[
\E[\DynamicRegret]
\leq4096\log^2(2\NumRounds\NumArms)
\sqrt{\NumClasses\NumRounds\TotalDifficulty}+1,
\qquad \text{ where  }
\TotalDifficulty=
\sum_{m=1}^{\NumClasses}(\ChangeBudget{m}+1)\ClassSize{m}.
\]

Our following analysis distinguishes intervals on which a class's means remain
constant from intervals containing a true rewards change. On the former,
concentration bounds define the difference between empirical and true means;
on the latter, reward observations collected before a change may no longer
represent the current means. We therefore partition each class's
timeline at its true change rounds and further split the resulting
stationary segments at algorithmic resets. This gives the
\emph{maximal static intervals} (MSIs), within which both the class
means and its phase start remain fixed. Uniform concentration on
stationary intervals then ensures that a reset cannot occur unless
a true change has occurred since the preceding phase began.
Each reset can consequently be charged to a distinct true change,
bounding the number of MSIs in terms of the class change budget.

We next show that every bad arm has a good blocking arm, which provides
a representative for comparisons even when the played or optimal arm
is classified as bad. Using the MSI partition, we bound the cumulative
confidence radii of played arms and then extend this bound to eligible
arms through the randomized class--scale selection rule. These bounds
control the estimation-error contributions in the regret decomposition.

What remains then is the cost of monitoring bad arms and the delay caused
by certificates that still contain pre-change observations. We charge
monitoring pulls to counter refreshes. To bound the delayed-detection
contribution, we further divide each true stationary segment into
rounds before the relevant counter is refreshed, rounds after that
refresh but before the next reset, and rounds after the reset.
Bounding these contributions and accounting for concentration failure
completes the proof of \Cref{thm:main-formal}.

\subsection{Notation and interval structure}
\label{app:known-budget-statements}
Below we detail the interval structure and definitions used in the proofs.

\paragraph{Maximal static intervals and current-MSI statistics.}
Our analysis compares empirical observations with the current means
using observations from a single algorithmic phase of each arms class. However, a class phase
may contain undetected mean changes. We therefore split each class's
timeline at both true changes and resets. The resulting
\emph{maximal static intervals} (MSIs) are the longest intervals on
which both the class means and its phase start remain fixed.

\begin{definition}[Maximal static intervals]
\label{def:ebfh-msi}
For each class \(m\in[\NumClasses]\), let
\[
\mathcal R_m
:=\{t\in[\NumRounds]:
\text{class }m\text{ is reset at the start of round }t\}.
\]
A \emph{maximal static interval} of class \(m\) is a nonempty
interval \(I=[s,e]\subseteq[\NumRounds]\), maximal under inclusion,
such that
\[
\bigl(\ChangeRounds{m}\cup\mathcal R_m\bigr)
\cap\{s+1,\ldots,e\}=\emptyset.
\]
\end{definition}

For each class, the MSIs partition \([\NumRounds]\). The first MSI
starts at round \(1\), and every subsequent MSI starts at a true
change or reset round. Each ends immediately before the next MSI,
or at \(\NumRounds\) if it is the last.

For an arm \(a\) and round \(t\), let \(\tau\) be the starting round
of the MSI of class \(\ArmClass{a}\) containing \(t\). We evaluate
the current-MSI statistics using only observations before \(t\):
\[
\NumPulls{a}{t}:=\NumPulls{a}{[\tau,t-1]},
\qquad
\EmpMean{a}{t}:=\EmpMean{a}{[\tau,t-1]},
\qquad
\ConfRadius{a}{t}:=\ConfRadius{a}{[\tau,t-1]},
\]
and, for every \(q\geq0\),
\[
\EmpBound{a}{\pm q}{t}
:=\EmpBound{a}{\pm q}{[\tau,t-1]}.
\]
At the start of an MSI, \(\tau=t\), so the empty-history convention
applies. These statistics are used in the analysis; the algorithm's
classification certificates still use phase intervals, which may
contain pre-change observations.

\paragraph{Blocking arms.}
In the regret decomposition, we compare the played arm with the optimal arm through
two intermediate arms classified as good. This separates the comparison
between good arms from the contributions of playing a bad arm and
failing to recognize an optimal arm. A good arm is blocking itself, hence represents itself;
a bad arm is represented by its blocking arm.

For each bad arm \(a\in\BadArms{t}\), the classification rule in
\Cref{alg:ebf-h} guarantees that its certificate set
\(\CertificateSet{t}{a}\) is nonempty. The algorithm selects a maximizing
certificate using its fixed tie-breaking rule, and
\(\BlockingArm{t}{a}\) denotes the other arm appearing in that
certificate, called the \emph{blocking arm} of \(a\). Formally, 
\begin{definition}[Blocking arm]
\label{def:ebfh-blocking-arm}
Fix a round \(t\in[\NumRounds]\) and a bad arm \(a\in\BadArms{t}\).
Let \(\CertificateSet{t}{a}\) be its set of valid classification
certificates:
\[
\CertificateSet{t}{a}
:=
\left\{
(b,I,I'):
\begin{array}{l}
b\in\ArmSet\setminus\{a\},\\
I\subseteq[\PhaseStart{\ArmClass{a}},t-1],\\
I'\subseteq[\PhaseStart{\ArmClass{b}},t-1],\\
\EmpBound{b}{-3}{I'}>\EmpBound{a}{+3}{I}
\end{array}
\right\},
\]
where \(I\) and \(I'\) are intervals of consecutive rounds, and the
phase starts are evaluated after the round-\(t\) reset step.

Since \(a\) is classified as bad, \(\CertificateSet{t}{a}\) is
nonempty. Select the triple
\[
\bigl(
\BlockingArm{t}{a},
\ArmCertificateInterval{t}{a},
\BlockingCertificateInterval{t}{a}
\bigr)
\in
\argmax_{(b,I,I')\in\CertificateSet{t}{a}}
\left\{
\EmpBound{b}{-1}{I'}-\EmpBound{a}{+1}{I}
\right\},
\]
using the algorithm's fixed deterministic tie-breaking rule.
The arm \(\BlockingArm{t}{a}\) is called the \emph{blocking arm}
of \(a\) at round \(t\). The associated intervals
\(\ArmCertificateInterval{t}{a}\) and
\(\BlockingCertificateInterval{t}{a}\) contain the observations
used for \(a\) and its blocking arm, respectively.
\end{definition}

The maximizing triple exists because the certificate set is finite
and nonempty, and the tie-breaking rule makes its selection unique.
By \Cref{prop:blocking-is-good}, the selected blocking arm satisfies
\(\BlockingArm{t}{a}\in\GoodArms{t}\).
This is a property of its classification, not a claim that its
current mean exceeds that of \(a\).

Thus this arm is well defined, and \Cref{prop:blocking-is-good}
establishes that \(\BlockingArm{t}{a}\in\GoodArms{t}\). Using blocking arms, we can define the representative of a bad arm.

\begin{definition}[Representatives]
\label{def:ebfh-representatives}
For each round \(t\in[\NumRounds]\), define the representative of the
played arm \(\PlayedArm{t}\) by
\[
\RepresentativeArm{t}:=
\begin{cases}
\PlayedArm{t},
&\PlayedArm{t}\in\GoodArms{t},\\
\BlockingArm{t}{\PlayedArm{t}},
&\PlayedArm{t}\in\BadArms{t},
\end{cases}
\]
and the representative of the optimal arm \(\OptimalArm{t}\) by
\[
\RepresentativeOptimalArm{t}:=
\begin{cases}
\OptimalArm{t},
&\OptimalArm{t}\in\GoodArms{t},\\
\BlockingArm{t}{\OptimalArm{t}},
&\OptimalArm{t}\in\BadArms{t}.
\end{cases}
\]
\end{definition}

Both representatives therefore belong to
\(\GoodArms{t}\subseteq\EligibleArms{t}\). Adding and subtracting
their mean rewards gives \eqref{eq:regret-decomposition}.
The construction uses historical certificates and does not assume
that a blocking arm's current mean exceeds that of the arm it
represents.

\paragraph{Certificate residuals.}
To analyze rounds on which the optimal arm is classified as bad,
we measure the discrepancy between its current mean and the historical
upper bound used in its classification certificate. We denote the
set of these rounds by \(\HardRounds\), formally defined as
\[
\HardRounds
:=\{t\in[\NumRounds]:\OptimalArm{t}\in\BadArms{t}\}.
\]
For each \(t\in\HardRounds\), define the \emph{certificate residual}
\[
r_t
:=
\MeanReward{\OptimalArm{t}}{t}
-
\EmpBound{\OptimalArm{t}}{+1}
{\ArmCertificateInterval{t}{\OptimalArm{t}}}.
\]
Here \(\ArmCertificateInterval{t}{\OptimalArm{t}}\) is the optimal
arm's selected certificate interval, and
\(\EmpBound{\OptimalArm{t}}{+1}
{\ArmCertificateInterval{t}{\OptimalArm{t}}}\)
is its empirical mean on that interval plus one confidence radius.
Thus a positive residual means that the optimal arm's current mean
exceeds its historical certificate upper bound.

We separate small residual contributions, which can be summed
directly, from larger ones that require the monitoring argument.
For each class \(m\in[\NumClasses]\), let
\[
\ClassOptimalRounds{m}
:=
\left|\{t\in[\NumRounds]:
\OptimalArm{t}\in\ClassArms{m}\}\right|
\]
be the total number of rounds whose optimal arm belongs to class \(m\).
These counts include all such rounds, not only those in
\(\HardRounds\), and satisfy
\(\sum_{m=1}^{\NumClasses}\ClassOptimalRounds{m}=\NumRounds\).
Denote 
\(\LogFactor=\lceil\log(2\NumRounds\NumArms)\rceil\),
where \(\log\) is the natural logarithm. Define the class threshold
\[
\ClassThreshold{m}
:=
\begin{cases}
\displaystyle
\LogFactor
\sqrt{\frac{\NumClasses\ClassSize{m}(\ChangeBudget{m}+1)}
{\ClassOptimalRounds{m}}},
&\ClassOptimalRounds{m}>0,\\[1.2ex]
0,
&\ClassOptimalRounds{m}=0.
\end{cases}
\]

After separating the confidence-radius contribution, the bound for
Term~(iii) involves
\(r_t-\DominanceGap{t}{\OptimalArm{t}}\).
This difference is nonpositive when the residual does not exceed
the dominance gap. If the residual exceeds the dominance gap but
not its class threshold, the difference is at most that threshold.
We therefore reserve the monitoring argument for rounds whose
residual exceeds both:
\[
\LargeResidualRounds
:=
\left\{t\in\HardRounds:
r_t>\max\left\{
\DominanceGap{t}{\OptimalArm{t}},
\ClassThreshold{\ArmClass{\OptimalArm{t}}}
\right\}\right\}.
\]

To match these residuals to the monitoring scales, we group the rounds
in \(\LargeResidualRounds\) by the optimal arm's class and the
magnitude of the residual. For each \(m\in[\NumClasses]\) and integer
\(j\geq1\), define
\begin{equation}
\label{eq:hard-residual-rounds}
\HardScaleRounds{m}{j}
:=
\left\{t\in\LargeResidualRounds:
\OptimalArm{t}\in\ClassArms{m},\
2^{-j}<r_t\leq2^{1-j}
\right\}.
\end{equation}
These sets and thresholds are used only in the analysis, not by
the algorithm.

\subsection{Concentration Bounds}
\label{app:proof-confidence-bounds}

The intervals used in the analysis may be selected based on observed
rewards. We therefore need a concentration bound that holds for all
stationary intervals at once, rather than only for an interval fixed
in advance.

For the analysis, imagine generating a reward sequence for each arm
and each of its stationary segments before interaction, independently
of the algorithm's random coins. Whenever an arm is pulled, it reveals
the next unused reward in the corresponding sequence. We prove
concentration simultaneously for every consecutive block of every
sequence. Since the observations collected within any stationary
round interval form such a block, this also covers intervals chosen
using the observed history. Intervals with no observations are handled
by the empty-history convention.

The resulting event \(\ConcentrationEvent\) depends only on the
presampled reward sequences, not on which rewards the algorithm
chooses to observe. Conditioning on this event therefore leaves
the distribution of the algorithm's independent randomness 
unchanged.

\begin{lemma}[Restatement of \Cref{lem:confidence-bounds}]
\label{lem:ebfh-concentration}
Represent the rewards by presampling one independent sequence for each arm
and each of its stationary segments.  For a segment of length \(d\), with
mean \(\mu\) and sequence \(Y_1,\ldots,Y_d\), let
\(\ConcentrationEvent\) be the event that, simultaneously for every such
arm--segment sequence and all
\(1\le r\le s\le d\),
\[
\left|\frac{1}{s-r+1}\sum_{u=r}^{s}Y_u-\mu\right|
\leq\sqrt{\frac{8\log(2\NumRounds\NumArms)}{s-r+1}}.
\]
Then \(\Prob(\ConcentrationEvent)\ge1-\NumRounds^{-1}\).  Moreover, on
\(\ConcentrationEvent\), every arm \(a\) and every nonempty interval \(I\)
inside one of its stationary segments satisfy
\[
  \bigl|\EmpMean{a}{I}-\mu\bigr|\leq\ConfRadius{a}{I}.
\]
Here \(\mu\) is the segment's common mean.
\end{lemma}

\begin{proof}
The proof idea is first to apply Hoeffding's inequality to fixed blocks of presampled
rewards, obtain simultaneous concentration over all such blocks,
and then show that this event covers the intervals selected by the
algorithm.

\medskip
\noindent\textbf{Step 1: Represent the rewards by independent sequences.}
For each arm \(a\in\ArmSet\), let \(s_a\) be its number of stationary
segments. For \(i\in\{1,\ldots,s_a\}\), denote the length and common
mean of its \(i\)-th segment by \(d_{a,i}\) and \(\mu_{a,i}\),
respectively. Since these segments partition the horizon,
\(
\sum_{i=1}^{s_a}d_{a,i}=\NumRounds.
\)
For each pair \((a,i)\), presample
\(
Y_{a,i,1},\ldots,Y_{a,i,d_{a,i}}
\)
independently from the reward distribution of arm \(a\) in that
segment. All entries are mutually independent and independent of
the algorithm's random coins.

The \(k\)-th pull of arm \(a\) within its \(i\)-th segment reveals
\(Y_{a,i,k}\). Conditional on the observed history and the selected
arm, the next revealed entry has the prescribed reward distribution.
Thus the action--reward history has the same law as in the original
interaction. We note this is the reward-stack representation of
\citet[Section~4.6]{lattimore2020bandit}, applied separately to each
arm's stationary segments.

\medskip
\noindent\textbf{Step 2: Apply Hoeffding's theorem to a fixed block.}
Fix an arm \(a\), a segment \(i\), and deterministic indices
\(1\leq r\leq s\leq d_{a,i}\). Write
\[
n:=s-r+1,
\qquad
\overline Y_{a,i}(r,s)
:=\frac1n\sum_{k=r}^{s}Y_{a,i,k}.
\]
The \(n\) summands are independent, take values in \([0,1]\), and
have common mean \(\mu_{a,i}\). Therefore, Theorem~2 of
\citet{hoeffding1963probability}, with all lower bounds equal to
zero and all upper bounds equal to one, gives
\[
\Prob\!\left(
\overline Y_{a,i}(r,s)-\mu_{a,i}>\varepsilon
\right)
\leq \exp(-2n\varepsilon^2)
\qquad(\varepsilon>0).
\]
Applying the same theorem to the variables \(1-Y_{a,i,k}\) bounds
the lower tail. Adding the two tail probabilities yields
\[
\Prob\!\left(
\left|\overline Y_{a,i}(r,s)-\mu_{a,i}\right|>\varepsilon
\right)
\leq2\exp(-2n\varepsilon^2).
\]
Taking
\(\varepsilon=\sqrt{8\log(2\NumRounds\NumArms)/n}\), we obtain
\[
\Prob\!\left(
\left|\overline Y_{a,i}(r,s)-\mu_{a,i}\right|
>
\sqrt{\frac{8\log(2\NumRounds\NumArms)}{s-r+1}}
\right)
\leq2(2\NumRounds\NumArms)^{-16}.
\]

\medskip
\noindent\textbf{Step 3: Obtain concentration for all blocks simultaneously.}
The event \(\ConcentrationEvent\) in the lemma is the intersection
of the successful concentration events from Step~2 over all arms,
segments, and block endpoints. A sequence of length \(d_{a,i}\)
contains exactly \(d_{a,i}(d_{a,i}+1)/2\) nonempty consecutive
blocks. Hence their total number is at most
\[
\sum_{a\in\ArmSet}\sum_{i=1}^{s_a}
\frac{d_{a,i}(d_{a,i}+1)}2
\leq
\sum_{a\in\ArmSet}
\left(\sum_{i=1}^{s_a}d_{a,i}\right)^2
=
\NumArms\NumRounds^2.
\]
The union bound, together with Step~2, therefore gives
\[
\Prob(\bar{\ConcentrationEvent})
\leq
2\NumArms\NumRounds^2(2\NumRounds\NumArms)^{-16}
\leq\NumRounds^{-1}.
\]
In particular,
\(\Prob(\ConcentrationEvent)\geq1-\NumRounds^{-1}\).
Because the event is defined using only the presampled sequences,
it is independent of the algorithm's random coins.

\medskip
\noindent\textbf{Step 4: Transfer the bound to adaptively selected intervals.}
Fix a realization in \(\ConcentrationEvent\), an arm \(a\), and
a nonempty interval \(I=[u,v]\) contained in the \(i\)-th stationary
segment of that arm. Let \(b\) be the starting round of this segment,
and define
\[
h:=\NumPulls{a}{[b,u-1]},
\qquad
n:=\NumPulls{a}{I}.
\]
Here \(h\) counts the entries revealed from this segment's sequence
before \(I\), while \(n\) counts those revealed during \(I\).

Suppose first that \(n\geq1\). The rewards observed from arm \(a\)
during \(I\) are exactly
\[
Y_{a,i,h+1},\ldots,Y_{a,i,h+n}.
\]
Consequently,
\[
\EmpMean{a}{I}
=\overline Y_{a,i}(h+1,h+n).
\]
Although \(h\) and \(n\) may depend on the observed history,
\(\ConcentrationEvent\) controls every block simultaneously.
Applying its defining inequality to this block gives
\[
\left|\EmpMean{a}{I}-\mu_{a,i}\right|
\leq
\sqrt{\frac{8\log(2\NumRounds\NumArms)}{n}}
= :
\ConfRadius{a}{I}.
\]
This is a pathwise application of the simultaneous event; it does
not condition on the random values of \(h\) or \(n\).

If \(n=0\), the empty-history convention gives
\(\EmpMean{a}{I}=0\) and
\(\ConfRadius{a}{I}=\sqrt{8\log(2\NumRounds\NumArms)}\geq1\).
Since \(\mu_{a,i}\in[0,1]\), the same inequality holds
deterministically.

The arm and interval were arbitrary, so the claimed empirical
concentration bounds hold simultaneously on
\(\ConcentrationEvent\).
\end{proof}

The conclusion is that under \(\ConcentrationEvent\), every current-MSI statistic therefore
concentrates around the arm's current mean. 

\subsection{Reset and classification properties}
\label{app:ebfh-algorithmic-properties}

We first use concentration to bound the number of resets, as formally stated next.


\begin{lemma}[Restatement of \Cref{lem:phase-resets}]
\label{lem:ebfh-phase-resets}
Assume event \(\ConcentrationEvent\) holds, then every class \(m\) is reset at most
\(\ChangeBudget{m}\) times.
\end{lemma}

The proof idea is that under \(\ConcentrationEvent\), if no mean in a class has changed during its current phase, confidence intervals computed from different
parts of that phase for the same arm all contain its common mean.
They therefore overlap, so the reset condition cannot hold. Full proof is given below.

\begin{proof}
Fix a class \(m\) and a round \(t\) at which the algorithm resets
that class. Let \(\PhaseStart{m}\) denote its phase start immediately
before the reset. By the change-detection rule in
\Cref{alg:ebf-h}, line~4, there exist an arm
\(a\in\ClassArms{m}\) and consecutive intervals
\(I_1,I_2\subseteq[\PhaseStart{m},t-1]\) such that
\[
\EmpBound{a}{-1}{I_1}>\EmpBound{a}{+1}{I_2}.
\]
Both witness intervals have non-empty pull counts. Under the empty-history
convention, a zero-pull interval has a negative lower bound and an upper
bound larger than one, whereas every empirical lower bound is at most one
and every empirical upper bound is nonnegative. Thus a zero-pull interval
cannot appear on either side of the strict inequality. In particular, both
intervals are nonempty.

Suppose, toward a contradiction, that class \(m\) has no change round in
\(\{\PhaseStart{m}+1,\ldots,t-1\}\). Then arm \(a\) has the constant mean
\(\MeanReward{a}{\PhaseStart{m}}\) throughout
\([\PhaseStart{m},t-1]\). On \(\ConcentrationEvent\),
\[
\left|\EmpMean{a}{I_r}-\MeanReward{a}{\PhaseStart{m}}\right|
\leq\ConfRadius{a}{I_r},
\qquad r\in\{1,2\}.
\]
Consequently,
\[
\EmpBound{a}{-1}{I_1}
\leq\MeanReward{a}{\PhaseStart{m}}
\leq\EmpBound{a}{+1}{I_2},
\]
contradicting the reset condition. Hence at least one change round of class
\(m\) lies strictly after the phase start and before \(t\).

Charge the reset to any such change round. After the reset at \(t\), the new
phase starts at \(t\), so any later reset must be charged to a change round
strictly after \(t\). Thus distinct resets receive distinct charges. Since
class \(m\) has at most \(\ChangeBudget{m}\) change rounds, it can be reset
at most \(\ChangeBudget{m}\) times.
\end{proof}

On \(\ConcentrationEvent\), every reset of class \(m\) follows a true
change within the phase being reset. Since each reset starts a new
phase, the same change cannot justify a later reset. Class \(m\)
therefore undergoes at most \(\ChangeBudget{m}\) resets.
Its first MSI starts at round \(1\), and every subsequent MSI starts
at a true change or reset round. Together, the at most
\(\ChangeBudget{m}\) true changes and \(\ChangeBudget{m}\) resets
thus partition the horizon into at most \(2\ChangeBudget{m}+1\) MSIs. Formally,


\begin{lemma}[Number of MSIs]
\label{lem:static-intervals}
Under the concentration event \(\ConcentrationEvent\), for each class $m$, the number of MSIs is at most $2\ChangeBudget{m} + 1$.
\end{lemma}

\begin{proof}
Fix a class \(m\). By \Cref{lem:phase-resets}, it has at most
\(\ChangeBudget{m}\) phase resets. Together with its at most
\(\ChangeBudget{m}\) change rounds, these events partition the rounds into at
most \(2\ChangeBudget{m}+1\) MSIs.
\end{proof}

Consequently, on \(\ConcentrationEvent\), the total number of
arm--MSI pairs is at most
\[
\ArmMSIPairBound
\leq2\sum_{m=1}^{\NumClasses}(\ChangeBudget{m}+1)\ClassSize{m}
=2\TotalDifficulty.
\]
This count includes pairs whose arms are never pulled. They will
contribute zero to the played-arm confidence sum.

\subsubsection{Blocking arms are good}
\label{app:proof-blocking-is-good}

We next show that every bad arm's selected blocking arm is classified as good.

\begin{proposition}[Restatement of \Cref{prop:blocking-is-good}]
\label{prop:ebfh-blocking-good}
For every round \(t\) and arm \(a\in\BadArms{t}\), its blocking arm $\BlockingArm{t}{a}$ is good, meaning  
\(\BlockingArm{t}{a}\in\GoodArms{t}\).
\end{proposition}

\begin{proof}
Fix a round \(t\) and an arm \(a\in\BadArms{t}\), and write
\(a':=\BlockingArm{t}{a}\). Suppose, toward a contradiction, that
\(a'\in\BadArms{t}\).
Let the selected maximal certificate for \(a'\) be the tuple
\(
\left(
\BlockingArm{t}{a'},
\ArmCertificateInterval{t}{a'},
\BlockingCertificateInterval{t}{a'}
\right)
\in\CertificateSet{t}{a'}.
\)
Its second component is the observation interval for \(a'\), and its
third component is the observation interval for
\(\BlockingArm{t}{a'}\), the arm blocking \(a'\).
It thus holds true that
\(
\EmpBound{\BlockingArm{t}{a'}}{-3}
{\BlockingCertificateInterval{t}{a'}}
>
\EmpBound{a'}{+3}{\ArmCertificateInterval{t}{a'}}.
\)

Let the selected certificate for \(a\) be the tuple
\(
\left(
a',
\ArmCertificateInterval{t}{a},
\BlockingCertificateInterval{t}{a}
\right)
\in\CertificateSet{t}{a}.
\)
Here the second component is the observation interval for \(a\),
whereas the third component is the observation interval for \(a'\).

The certificate by which \(a'\) blocks \(a\) likewise satisfies
\(
\EmpBound{a'}{-3}{\BlockingCertificateInterval{t}{a}}
>
\EmpBound{a}{+3}{\ArmCertificateInterval{t}{a}}.
\)

\textbf{Step 1: Verify a second blocking certificate for \(a\).}
Let us consider the candidate tuple
\(
\left(
\BlockingArm{t}{a'},
\ArmCertificateInterval{t}{a},
\BlockingCertificateInterval{t}{a'}
\right).
\)
This retains the interval for \(a\) from its selected certificate
and takes the blocking arm and its observation interval from the
selected certificate for \(a'\). To verify this candidate, we connect
the two original certificates through
\(\BlockingCertificateInterval{t}{a}\) and
\(\ArmCertificateInterval{t}{a'}\): both are observation intervals
for \(a'\), although they use in different certificates.

Because \(a'\) is bad after the reset step, its class was not reset at round
\(t\): a newly reset arm has no observations, so its \(+3\) bound
exceeds one, whereas every arm's \(-3\) bound is at most one. Thus the change-detection test is false
on the current phase of \(a'\). Applying that fact to the
two intervals of \(a'\) gives
\begin{align*}
\EmpBound{a'}{-1}{\BlockingCertificateInterval{t}{a}}
\leq
\EmpBound{a'}{+1}{\ArmCertificateInterval{t}{a'}}.
\end{align*}
Equivalently,
\begin{align*}
\EmpMean{a'}{\ArmCertificateInterval{t}{a'}}
\geq{}
\EmpMean{a'}{\BlockingCertificateInterval{t}{a}}
-\ConfRadius{a'}{\BlockingCertificateInterval{t}{a}}
-\ConfRadius{a'}{\ArmCertificateInterval{t}{a'}}.
\end{align*}
Combining this inequality with the two blocking certificates yields
\begin{align*}
&\EmpBound{\BlockingArm{t}{a'}}{-3}
{\BlockingCertificateInterval{t}{a'}}\\
&>
\EmpBound{a'}{+3}{\ArmCertificateInterval{t}{a'}}\\
&\geq
\EmpMean{a'}{\BlockingCertificateInterval{t}{a}}
-\ConfRadius{a'}{\BlockingCertificateInterval{t}{a}}
+2\ConfRadius{a'}{\ArmCertificateInterval{t}{a'}}\\
&\geq
\EmpBound{a'}{-1}{\BlockingCertificateInterval{t}{a}}\\
&=
\EmpBound{a'}{-3}{\BlockingCertificateInterval{t}{a}}
+2\ConfRadius{a'}{\BlockingCertificateInterval{t}{a}}\\
&>
\EmpBound{a}{+3}{\ArmCertificateInterval{t}{a}}.
\end{align*}
Thus
\(
\left(
\BlockingArm{t}{a'},
\ArmCertificateInterval{t}{a},
\BlockingCertificateInterval{t}{a'}
\right)
\)
satisfies the blocking inequality for \(a\). Moreover,
\(\BlockingArm{t}{a'}\neq a\): otherwise the preceding strict separation
would give two intervals of \(a\) in its current phase with disjoint
one-confidence-bound intervals, and the change-detection step would have
reset its class before classification. The displayed triplet is therefore a
valid candidate in \(\CertificateSet{t}{a}\).

\textbf{Step 2: Compare the empirical separations.}
We compare the new candidate tuple from Step~1 with the originally
selected tuple
\(
\left(
a',
\ArmCertificateInterval{t}{a},
\BlockingCertificateInterval{t}{a}
\right).
\)
Both use the same interval \(\ArmCertificateInterval{t}{a}\) for
the blocked arm \(a\), so their one-radius separations subtract
the same upper bound
\(\EmpBound{a}{+1}{\ArmCertificateInterval{t}{a}}\).
It therefore suffices to show that the new candidate's blocking-arm
lower bound is strictly larger.

The inequalities of Step 1 also give
\(
\EmpBound{\BlockingArm{t}{a'}}{-3}
{\BlockingCertificateInterval{t}{a'}}
>
\EmpBound{a'}{-1}{\BlockingCertificateInterval{t}{a}}.
\)
Consequently,
\begin{align*}
&\EmpBound{\BlockingArm{t}{a'}}{-1}
{\BlockingCertificateInterval{t}{a'}}
-\EmpBound{a}{+1}{\ArmCertificateInterval{t}{a}}\\
&=
\EmpBound{\BlockingArm{t}{a'}}{-3}
{\BlockingCertificateInterval{t}{a'}}
+2\ConfRadius{\BlockingArm{t}{a'}}
{\BlockingCertificateInterval{t}{a'}}
-\EmpBound{a}{+1}{\ArmCertificateInterval{t}{a}}\\
&>
\EmpBound{a'}{-1}{\BlockingCertificateInterval{t}{a}}
-\EmpBound{a}{+1}{\ArmCertificateInterval{t}{a}}.
\end{align*}
Hence this candidate has a strictly larger empirical separation than the
certificate defining \(\BlockingArm{t}{a}=a'\), contradicting its maximizing
property. Therefore \(\BlockingArm{t}{a}\in\GoodArms{t}\).
\end{proof}


The preceding proposition ensures that \(\GoodArms{t}\) is nonempty
at every round: if any arm is bad, its blocking arm is good;
otherwise, every arm is good. Since all good arms are eligible,
\(\EligibleArms{t}\) is also nonempty. Moreover, both representatives
in the regret decomposition are good and hence eligible. To control
their confidence contributions, we therefore bound the cumulative
confidence radii of eligible arms.

\subsection{Cumulative confidence-sum bounds}
\label{app:cb-sums}

We first consider the simpler sum over arms actually played.
Within an MSI, each pull increases the corresponding arm's observation
count, allowing us to sum its confidence radii in terms of its total
number of pulls.

\begin{lemma}[Played-arm confidence sum]
\label{lem:played-cb-sum}
Assume that the concentration event \(\ConcentrationEvent\) defined in
\Cref{lem:confidence-bounds} holds. Then, for the realized played-arm
sequence \(\PlayedArm{1},\ldots,\PlayedArm{\NumRounds}\),
\[
\sum_{t=1}^{\NumRounds}\ConfRadius{\PlayedArm{t}}{t}
\leq
3\sqrt{8\log(2\NumRounds\NumArms)\,\NumRounds
       \ArmMSIPairBound}.
\]
In particular, this is
\(\BigOTilde(\sqrt{\NumRounds\TotalDifficulty})\).
\end{lemma}

\begin{proof}
Assume \(\ConcentrationEvent\) holds and fix a realization of the played-arm sequence. We group the pulls by their arm and MSI, bound the contribution
of each group, and then sum these bounds.

\medskip
\noindent\textbf{Step 1: Group the pulls by arm and MSI.}
Let \(\mathcal P\) be the set of pairs \((a,I)\), where \(a\) is an
arm and \(I\) is an MSI of its class. For each pair, define
\(
n_{a,I}
:=\NumPulls{a}{I}
=\sum_{t\in I}\Indicator{\PlayedArm{t}=a}.
\)
Thus \(n_{a,I}\) is the total number of pulls of arm \(a\) during \(I\).

Every round contributes to exactly one pair: the played arm and the
unique MSI of its class containing that round. Consequently,
\(
\sum_{(a,I)\in\mathcal P}n_{a,I}=\NumRounds.
\) 
Also, by \Cref{lem:static-intervals}, class \(m\) has at most
\(2\ChangeBudget{m}+1\) MSIs on \(\ConcentrationEvent\).
Each such MSI is paired with the \(\ClassSize{m}\) arms in that
class, so
\(
|\mathcal P|\leq\ArmMSIPairBound.
\)

\medskip
\noindent\textbf{Step 2: Bound the contribution of one arm within one MSI.}
Fix \((a,I)\in\mathcal P\), and write \(n:=n_{a,I}\).
If \(n=0\), this pair contributes nothing. Otherwise, let
\(t_1<\cdots<t_n\) be the rounds in \(I\) at which arm \(a\) is
played. Here \(t_k\) is the global round of the \(k\)-th pull of
arm \(a\) within \(I\).

Immediately before round \(t_k\), arm \(a\) has exactly \(k-1\)
observations in this MSI. Since the current-MSI statistics exclude
the current round's observation, for every \(k\in\{1,\ldots,n\}\),
\[
\NumPulls{a}{t_k}=k-1,
\qquad
\ConfRadius{a}{t_k}
=
\sqrt{\frac{8\log(2\NumRounds\NumArms)}
           {\max\{k-1,1\}}}.
\]

To sum these contributions, write \(q:=k-1\) for the number of
preceding pulls of arm \(a\) in \(I\). As the pull index \(k\)
ranges from \(1\) to \(n\), this count \(q\) ranges from \(0\) to
\(n-1\). Therefore,
\begin{align*}
\sum_{\substack{t\in I\\\PlayedArm{t}=a}}\ConfRadius{a}{t}
&=\sum_{k=1}^{n}\ConfRadius{a}{t_k}\\
&=\sqrt{8\log(2\NumRounds\NumArms)}
  \sum_{q=0}^{n-1}\frac{1}{\sqrt{\max\{q,1\}}}\\
&=\sqrt{8\log(2\NumRounds\NumArms)}
  \left(1+\sum_{q=1}^{n-1}\frac{1}{\sqrt q}\right).
\end{align*}
The separate term \(1\) corresponds to \(q=0\), namely the first
pull, before which no observation of arm \(a\) has been collected
in this MSI. The remaining sum accounts for the second through
the \(n\)-th pulls.

We bound the remaining sum by an integral. For each integer
\(q\geq2\) and every \(x\in[q-1,q]\), we have \(x\leq q\), so
\(x^{-1/2}\geq q^{-1/2}\). Integrating over this interval of length
one gives
\[
\frac{1}{\sqrt q}
\leq\int_{q-1}^{q}\frac{dx}{\sqrt x}.
\]
For \(n\geq2\), separate the term \(q=1\) and apply this inequality
to the other terms:
\begin{align*}
\sum_{q=1}^{n-1}\frac{1}{\sqrt q}
&=1+\sum_{q=2}^{n-1}\frac{1}{\sqrt q}\\
&\leq1+\int_1^{n-1}\frac{dx}{\sqrt x}\\
&=2\sqrt{n-1}-1\\
&\leq2\sqrt n.
\end{align*}
When \(n=2\), the sum from \(q=2\) to \(n-1\) and the integral
are both zero. When \(n=1\), the entire sum from \(q=1\) to
\(n-1\) is empty, so the bound by \(2\sqrt n\) also holds.
Consequently, for every \(n\geq1\),
\[
1+\sum_{q=1}^{n-1}\frac{1}{\sqrt q}
\leq1+2\sqrt n
\leq3\sqrt n,
\]
where the final inequality uses \(1\leq\sqrt n\).
Substituting this into the confidence sum gives
\[
\sum_{\substack{t\in I\\\PlayedArm{t}=a}}\ConfRadius{a}{t}
\leq
3\sqrt{8\log(2\NumRounds\NumArms)}\,\sqrt{n_{a,I}}.
\]
This inequality also holds when \(n_{a,I}=0\), since both sides
are zero.

\medskip
\noindent\textbf{Step 3: Sum over all arm--MSI pairs.}
Using the bound from Step~2,
\begin{align*}
\sum_{t=1}^{\NumRounds}\ConfRadius{\PlayedArm{t}}{t}
&=
\sum_{(a,I)\in\mathcal P}
\sum_{\substack{t\in I\\\PlayedArm{t}=a}}\ConfRadius{a}{t}\\
&\leq
3\sqrt{8\log(2\NumRounds\NumArms)}
\sum_{(a,I)\in\mathcal P}\sqrt{n_{a,I}}.
\end{align*}
Apply Cauchy--Schwarz to the sequences
\((1)_{(a,I)\in\mathcal P}\) and
\((\sqrt{n_{a,I}})_{(a,I)\in\mathcal P}\):
\begin{align*}
\sum_{(a,I)\in\mathcal P}\sqrt{n_{a,I}}
&\leq
\sqrt{
\left(\sum_{(a,I)\in\mathcal P}1\right)
\left(\sum_{(a,I)\in\mathcal P}n_{a,I}\right)}\\
&=\sqrt{|\mathcal P|\,\NumRounds}\\
&\leq\sqrt{\ArmMSIPairBound\,\NumRounds}.
\end{align*}
The equality uses the fact that the pull counts sum to
\(\NumRounds\), and the last inequality uses the bound on
\(|\mathcal P|\) from Step~1. Hence
\[
\sum_{t=1}^{\NumRounds}\ConfRadius{\PlayedArm{t}}{t}
\leq
3\sqrt{8\log(2\NumRounds\NumArms)\,\NumRounds
       \ArmMSIPairBound}.
\]

Finally, since
\(2\ChangeBudget{m}+1\leq2(\ChangeBudget{m}+1)\) for every class
\(m\),
\[
\ArmMSIPairBound
\leq
2\sum_{m=1}^{\NumClasses}
(\ChangeBudget{m}+1)\ClassSize{m}
=
2\TotalDifficulty.
\]
Substituting this inequality and using \(3\sqrt{8\cdot2}=12\)
gives
\[
\sum_{t=1}^{\NumRounds}\ConfRadius{\PlayedArm{t}}{t}
\leq
12\sqrt{\log(2\NumRounds\NumArms)\,
         \NumRounds\TotalDifficulty},
\]
which is \(\BigOTilde(\sqrt{\NumRounds\TotalDifficulty})\).
\end{proof}

We note that since every term corresponds to an arm that was
actually played, this bound has no additional \(\sqrt{\NumClasses}\) factor.

In the previous proof, the bound on the number of arm--MSI pairs then
controls the total contribution across the horizon. For eligible
arms that are not played, this direct counting argument does not
apply. We instead use the randomized class--scale selection and
recency rule to relate their radii to those of arms the algorithm
samples. This yields a bound on the sum of the largest eligible-arm
radius at each round, which applies in particular to both
representatives, as formally stated next.

\begin{lemma}[Eligible-arm confidence sum]\label{lem:proof-eligible-cb-sums}
Let \(\ConcentrationEvent\) be the concentration event from
\Cref{lem:confidence-bounds}. For every admissible instance,
Algorithm~\texttt{EBF-H} satisfies
\[
\E\!\left[
\left.
\sum_{t=1}^{\NumRounds}
\max_{a\in\EligibleArms{t}}\ConfRadius{a}{t}
\,\right|\ConcentrationEvent
\right]
\leq
64\,\log(2\NumRounds\NumArms)
\sqrt{\NumClasses\NumRounds\TotalDifficulty}.
\]

Consequently, for any possibly random sequence
\(a'_1,\ldots,a'_{\NumRounds}\) satisfying
\(a'_t\in\EligibleArms{t}\) at every round,
\[
\E\!\left[
\left.
\sum_{t=1}^{\NumRounds}\ConfRadius{a'_t}{t}
\,\right|\ConcentrationEvent
\right]
\leq
64\,\log(2\NumRounds\NumArms)
\sqrt{\NumClasses\NumRounds\TotalDifficulty}.
\]
In particular, both expectations are
\(\widetilde{O}\!\left(\sqrt{\NumClasses\NumRounds\TotalDifficulty}\right)\).
\end{lemma}

The proof idea is that an eligible arm need not be played when its confidence radius is
counted in the sum. We therefore cannot simply count its successive
pulls as in the preceding lemma. Instead, at each round we identify
a comparison arm whose radius is at least half the largest
eligible-arm radius and which the algorithm plays with a controlled
probability. Whenever this comparison arm is played, its MSI
observation count increases. We use these increases to bound the
number of rounds with small observation counts. On the remaining
rounds, the counts are large enough that the confidence radii
themselves are small. This is formally given next.

\begin{proof}
Let 
\(
\ell:=\log(2\NumRounds\NumArms)
\) and recall that \(\NumPulls{a}{t}\) counts observations of arm \(a\)
before round \(t\) in its current MSI, and
\(
\ConfRadius{a}{t}
=
\sqrt{\frac{8\ell}{\max\{\NumPulls{a}{t},1\}}}.
\)
We retain the randomness of the interaction throughout the proof;
the bound is on a conditional expectation, rather than on every
realized sequence of plays.

\medskip
\noindent\textbf{Step 1: Choose a comparison arm with a comparable radius.}
At each round \(t\), after the counter refreshes and before the
class and scale draws, choose
\(
\WidestEligibleArm{t}
\in\argmax_{a\in\EligibleArms{t}}\ConfRadius{a}{t},
\)
using the fixed tie-breaking rule. Such an arm exists because
\(\EligibleArms{t}\) is nonempty.

Let
\(
k_t:=\NumPulls{\WidestEligibleArm{t}}{t}
\) 
be the number of observations this arm has already received in
its current MSI. Choose the scale
\(
h_t:=\left\lceil\log_2(k_t+1)\right\rceil.
\)
This scale is used only in the analysis; it is not necessarily
the scale drawn by the algorithm. Since
\(0\leq k_t\leq\NumRounds-1\), we have
\(h_t\in\{0,\ldots,\MaxScale\}\). The definition also gives
\(
k_t<2^{h_t}\leq2(k_t+1).
\)

Define the comparison arm by
\(
\TwinArm{t}
\in
\argmin_{a\in
\EligibleArms{t}\cap
\ClassArms{\ArmClass{\WidestEligibleArm{t}}}}
\PriorityTime{a}{h_t},
\)
using the same rule as in line 14 of \Cref{alg:ebf-h}.
Thus \(\TwinArm{t}\) is the arm the algorithm would play if it
selected the class of \(\WidestEligibleArm{t}\) and scale \(h_t\).

We now compare the two arms' observation counts. Let \(u_t\) be
the starting round of their common current MSI. They share this
MSI because they belong to the same class. Recall that
\(\PriorityTime{a}{h_t}\) is the round of arm \(a\)'s
\(2^{h_t}\)-th most recent pull in its current phase, or
\(-\infty\) if that many pulls do not exist.

Arm \(\WidestEligibleArm{t}\) has only \(k_t<2^{h_t}\)
observations in its current MSI. Its \(2^{h_t}\)-th most recent
phase observation must therefore precede \(u_t\), if it exists.
By the minimizing choice of \(\TwinArm{t}\),
\[
\PriorityTime{\TwinArm{t}}{h_t}
\leq
\PriorityTime{\WidestEligibleArm{t}}{h_t}
<u_t.
\]
Consequently, \(\TwinArm{t}\) also has fewer than \(2^{h_t}\)
observations in this MSI:
\[
\NumPulls{\TwinArm{t}}{t}
<2^{h_t}\leq2(k_t+1).
\]

If \(k_t=0\), then \(h_t=0\), so this inequality implies
\(\NumPulls{\TwinArm{t}}{t}=0\).
If \(k_t\geq1\), then \(2(k_t+1)\leq4k_t\).
In either case,
\[
\max\{\NumPulls{\TwinArm{t}}{t},1\}
\leq4\max\{k_t,1\}.
\]
Substituting into the confidence-radius formula gives
\[
\frac{\ConfRadius{\WidestEligibleArm{t}}{t}}
     {\ConfRadius{\TwinArm{t}}{t}}
=
\sqrt{
\frac{\max\{\NumPulls{\TwinArm{t}}{t},1\}}
     {\max\{k_t,1\}}}
\leq2.
\]
Therefore, at every round,
\[
\max_{a\in\EligibleArms{t}}\ConfRadius{a}{t}
\leq2\ConfRadius{\TwinArm{t}}{t}.
\]
It remains to bound the sum of the comparison arms' radii.

\medskip
\noindent\textbf{Step 2: Bound the probability of playing the comparison arm.}
For the conditioning argument, let \(\mathcal H_t\) be the
sigma-field generated by all presampled reward sequences, all
algorithmic random draws before round \(t\), and the counter-refresh
coins at round \(t\). These variables determine the eligible set,
the two arms from Step~1, and the scale \(h_t\).
The round-\(t\) class and scale draws have not been included.

Conditional on \(\mathcal H_t\), the algorithm selects uniformly
among at most \(\NumClasses\) classes containing an eligible arm.
The class of \(\WidestEligibleArm{t}\) is one of these classes.
Independently, the algorithm selects uniformly among the
\(\MaxScale+1\) maintained scales. Thus
\begin{align*}
\Prob\!\left(
\SelectedClass{t}=\ArmClass{\WidestEligibleArm{t}},
\ \SelectedScale{t}=h_t
\,\middle|\,\mathcal H_t
\right)
\geq
\frac{1}{\NumClasses(\MaxScale+1)}
=:\TwinSelectionProbability.
\end{align*}
When these two choices occur, the algorithm plays \(\TwinArm{t}\).
It may also play this arm under other scale choices, so
\[
\Prob\!\left(
\PlayedArm{t}=\TwinArm{t}
\,\middle|\,\mathcal H_t
\right)
\geq\TwinSelectionProbability.
\]

The event \(\ConcentrationEvent\) depends only on the presampled
reward sequences, which are included in \(\mathcal H_t\).
The fresh class and scale draws therefore retain these probabilities
when we subsequently condition on \(\ConcentrationEvent\).

\medskip
\noindent\textbf{Step 3: Count rounds with a given number of preceding observations.}
Fix a nonnegative integer \(n\). Here \(n\) is a possible
observation count, not a round number or the total number of pulls
over the horizon. Define the random counts
\begin{align*}
V_n
&:=
\sum_{t=1}^{\NumRounds}
\Indicator{\NumPulls{\TwinArm{t}}{t}=n},\\
S_n
&:=
\sum_{t=1}^{\NumRounds}
\Indicator{
\NumPulls{\TwinArm{t}}{t}=n,\
\PlayedArm{t}=\TwinArm{t}}.
\end{align*}
The first count, \(V_n\), includes every round when the comparison
arm has exactly \(n\) preceding observations in its current MSI.
The second, \(S_n\), includes only those rounds when the algorithm
actually plays that comparison arm. Both counts are zero for
\(n\geq\NumRounds\).

First, we bound \(S_n\). Every round counted by \(S_n\) is the
\((n+1)\)-st pull of a particular arm within a particular MSI.
The same arm--MSI pair cannot have two different
\((n+1)\)-st pulls. Hence each pair contributes at most once to
\(S_n\), even though the comparison arm may change between rounds.
By \Cref{lem:static-intervals}, on \(\ConcentrationEvent\),
\[
S_n
\leq\ArmMSIPairBound
\leq2\TotalDifficulty.
\]

Next, we relate \(V_n\) to \(S_n\).
The comparison arm and its preceding observation count are
determined by \(\mathcal H_t\). Therefore, Step~2 gives
\begin{align*}
&\E\!\left[
\Indicator{
\NumPulls{\TwinArm{t}}{t}=n,\
\PlayedArm{t}=\TwinArm{t}}
\,\middle|\,\mathcal H_t
\right]\\
&\quad=
\Indicator{\NumPulls{\TwinArm{t}}{t}=n}
\Prob\!\left(
\PlayedArm{t}=\TwinArm{t}
\,\middle|\,\mathcal H_t
\right)\\
&\quad\geq
\TwinSelectionProbability\,
\Indicator{\NumPulls{\TwinArm{t}}{t}=n}.
\end{align*}
Taking conditional expectations given \(\ConcentrationEvent\)
and summing over \(t\), the tower property
\citep[Theorem~2.12(5)]{lattimore2020bandit} yields
\[
\TwinSelectionProbability\,
\E[V_n\mid\ConcentrationEvent]
\leq
\E[S_n\mid\ConcentrationEvent]
\leq2\TotalDifficulty.
\]
Thus, for every nonnegative integer \(n\),
\[
\E[V_n\mid\ConcentrationEvent]
\leq\frac{2\TotalDifficulty}{\TwinSelectionProbability}.
\]
This bounds the number of rounds with observation count \(n\),
including rounds when the comparison arm is not played.

\medskip
\noindent\textbf{Step 4: Sum the confidence radii over observation counts.}
Recall that
\(\ell=\log(2\NumRounds\NumArms)\), and that \(V_n\) counts the
rounds at which the comparison arm has exactly \(n\) preceding
observations in its current MSI. At every such round, its radius
equals
\(
\sqrt{\frac{8\ell}{\max\{n,1\}}}.
\)
The combined contribution of these rounds is therefore
\(
V_n\sqrt{\frac{8\ell}{\max\{n,1\}}}.
\)
The comparison arms may differ between these rounds, but their
radii are equal because their observation counts are equal.

Every round has exactly one observation count
\(n\in\{0,\ldots,\NumRounds-1\}\). Thus each round is counted
in exactly one of \(V_0,\ldots,V_{\NumRounds-1}\), and
\(
\sum_{n=0}^{\NumRounds-1}V_n=\NumRounds.
\)
Adding the contributions for these different counts gives, on
every realization,
\[
\sum_{t=1}^{\NumRounds}\ConfRadius{\TwinArm{t}}{t}
=
\sqrt{8\ell}
\sum_{n=0}^{\NumRounds-1}
\frac{V_n}{\sqrt{\max\{n,1\}}}.
\]
Taking conditional expectations and using linearity, we obtain
\[
\E\!\left[
\left.
\sum_{t=1}^{\NumRounds}\ConfRadius{\TwinArm{t}}{t}
\,\right|\ConcentrationEvent
\right]
=
\sqrt{8\ell}
\sum_{n=0}^{\NumRounds-1}
\frac{\E[V_n\mid\ConcentrationEvent]}
     {\sqrt{\max\{n,1\}}}.
\]

Small observation counts give large radii, but Step~3 limits how
often each such count occurs:
\(
\E[V_n\mid\ConcentrationEvent]
\leq\frac{2\TotalDifficulty}{\TwinSelectionProbability}.
\) 
Large observation counts may occur many times, but each corresponding
radius is small. We use these two bounds on opposite sides of the
threshold
\(
\beta:=
\frac{\NumRounds\TwinSelectionProbability}{2\TotalDifficulty}.
\)
The threshold is a positive real number; the observation counts
compared with it are nonnegative integers.

\medskip
\noindent\emph{First handle the case \(\beta<1\).}
In this case,
\(
\NumRounds<
\frac{2\TotalDifficulty}{\TwinSelectionProbability}.
\)
Every arm's confidence radius is at most \(\sqrt{8\ell}\).
Hence we can bound the largest eligible-arm radius directly,
without using the comparison arm:
\begin{align*}
\E\!\left[
\left.
\sum_{t=1}^{\NumRounds}
\max_{a\in\EligibleArms{t}}\ConfRadius{a}{t}
\,\right|\ConcentrationEvent
\right]
\leq\NumRounds\sqrt{8\ell}
\leq
\sqrt{8\ell\NumRounds\frac{2\TotalDifficulty}{\TwinSelectionProbability}}
=
4\sqrt{
\frac{\ell\NumRounds\TotalDifficulty}
     {\TwinSelectionProbability}}.
\end{align*}
The second inequality uses
\(\NumRounds^2\leq
2\NumRounds\TotalDifficulty/\TwinSelectionProbability\).
This case already gives a smaller bound than the one obtained
below. We therefore assume \(\beta\geq1\) for the next two parts.

\medskip
\noindent\emph{Small counts: fewer than \(\beta\) preceding observations.}
The contribution of rounds with
\(\NumPulls{\TwinArm{t}}{t}<\beta\) can be written as
\begin{align*}
\E\!\left[
\left.
\sum_{\substack{t\in[\NumRounds]
               \NumPulls{\TwinArm{t}}{t}<\beta}}
\ConfRadius{\TwinArm{t}}{t}
\,\right|\ConcentrationEvent
\right]
&=
\sqrt{8\ell}
\sum_{\substack{0\leq n\leq\NumRounds-1\\n<\beta}}
\frac{\E[V_n\mid\ConcentrationEvent]}
     {\sqrt{\max\{n,1\}}}\\
&\leq
\sqrt{8\ell}\,
\frac{2\TotalDifficulty}{\TwinSelectionProbability}
\sum_{\substack{0\leq n\leq\NumRounds-1\\n<\beta}}
\frac{1}{\sqrt{\max\{n,1\}}}
\end{align*}
where the inequality applies the bound from Step~3 to each count
\(n\). All sums over \(n\) below also range over integers.

To bound the remaining sum, the terms \(n=0\) and \(n=1\),
when present, contribute at most two. For each integer \(n\geq2\),
the function \(x^{-1/2}\) is at least \(n^{-1/2}\) throughout
\([n-1,n]\). Therefore,
\[
\frac{1}{\sqrt n}
\leq\int_{n-1}^{n}\frac{dx}{\sqrt x}.
\]
For the integers \(n<\beta\), these integration intervals have
disjoint interiors and lie inside \([1,\beta]\). Since the
integrand is nonnegative,
\begin{align*}
\sum_{\substack{0\leq n\leq\NumRounds-1\\n<\beta}}
\frac{1}{\sqrt{\max\{n,1\}}}
\leq2+\int_1^\beta\frac{dx}{\sqrt x}
=2+2(\sqrt\beta-1)
=2\sqrt\beta.
\end{align*}
This remains valid when \(\beta=1\): only the term \(n=0\)
is included, and the upper bound is two.

Substituting this estimate and the definition of \(\beta\) gives
\begin{align*}
\E\!\left[
\left.
\sum_{\substack{t\in[\NumRounds]\\
               \NumPulls{\TwinArm{t}}{t}<\beta}}
\ConfRadius{\TwinArm{t}}{t}
\,\right|\ConcentrationEvent
\right]
\leq
\sqrt{8\ell}\,
\frac{2\TotalDifficulty}{\TwinSelectionProbability}
\cdot2\sqrt\beta
=
\sqrt{8\ell}\,
\frac{4\TotalDifficulty}{\TwinSelectionProbability}
\sqrt{\frac{\NumRounds\TwinSelectionProbability}
           {2\TotalDifficulty}}
=
8\sqrt{
\frac{\ell\NumRounds\TotalDifficulty}
     {\TwinSelectionProbability}}.
\end{align*}

\medskip
\noindent\emph{Large counts: at least \(\beta\) preceding observations.}
On a round with
\(\NumPulls{\TwinArm{t}}{t}\geq\beta\), the confidence-radius
formula gives
\[
\ConfRadius{\TwinArm{t}}{t}
=
\sqrt{\frac{8\ell}
{\max\{\NumPulls{\TwinArm{t}}{t},1\}}}
\leq\sqrt{\frac{8\ell}{\beta}}.
\]
There are at most \(\NumRounds\) such rounds. Thus, on every
realization,
\begin{align*}
\sum_{\substack{t\in[\NumRounds]\\
               \NumPulls{\TwinArm{t}}{t}\geq\beta}}
\ConfRadius{\TwinArm{t}}{t}
\leq
\NumRounds\sqrt{\frac{8\ell}{\beta}}
=
\NumRounds
\sqrt{\frac{16\ell\TotalDifficulty}
           {\NumRounds\TwinSelectionProbability}}
=
4\sqrt{
\frac{\ell\NumRounds\TotalDifficulty}
     {\TwinSelectionProbability}}.
\end{align*}
The same bound holds after taking conditional expectations given
\(\ConcentrationEvent\).

\medskip
\noindent\emph{Combine the two contributions.}
Each round has either fewer than \(\beta\) observations or at least
\(\beta\) observations, so the two sums cover all rounds without
overlap. Adding their conditional bounds yields
\[
\E\!\left[
\left.
\sum_{t=1}^{\NumRounds}\ConfRadius{\TwinArm{t}}{t}
\,\right|\ConcentrationEvent
\right]
\leq
(8+4)\sqrt{
\frac{\ell\NumRounds\TotalDifficulty}
     {\TwinSelectionProbability}}.
\]
By Step~1,
\[
\max_{a\in\EligibleArms{t}}\ConfRadius{a}{t}
\leq2\ConfRadius{\TwinArm{t}}{t}
\]
at every round. Consequently,
\[
\E\!\left[
\left.
\sum_{t=1}^{\NumRounds}
\max_{a\in\EligibleArms{t}}\ConfRadius{a}{t}
\,\right|\ConcentrationEvent
\right]
\leq
24\sqrt{
\frac{\ell\NumRounds\TotalDifficulty}
     {\TwinSelectionProbability}}.
\]
This bound also covers the case \(\beta<1\), for which we
obtained the smaller coefficient four.

Recall that
\[
\TwinSelectionProbability
=\frac{1}{\NumClasses(\MaxScale+1)}.
\]
Moreover, since
\(\MaxScale=\lceil\log_2\NumRounds\rceil\),
\(\NumRounds\geq1\), and \(\NumArms\geq2\),
\[
\MaxScale+1
\leq\log_2\NumRounds+2
\leq3\log(2\NumRounds\NumArms)
=3\ell.
\]
Substituting these relations gives
\begin{align*}
\E\!\left[
\left.
\sum_{t=1}^{\NumRounds}
\max_{a\in\EligibleArms{t}}\ConfRadius{a}{t}
\,\right|\ConcentrationEvent
\right]
&\leq
24\sqrt{\ell\NumClasses(\MaxScale+1)
             \NumRounds\TotalDifficulty}\\
&\leq
24\sqrt{3}\,\ell
\sqrt{\NumClasses\NumRounds\TotalDifficulty}\\
&\leq
32\sqrt{3}\,\log(2\NumRounds\NumArms)
\sqrt{\NumClasses\NumRounds\TotalDifficulty}.
\end{align*}
The last inequality uses \(24\leq32\), giving the constant stated
in the lemma. In particular, the expectation is
\(
\widetilde{O}\!\left(
\sqrt{\NumClasses\NumRounds\TotalDifficulty}
\right).
\)

Finally, let \(a'_1,\ldots,a'_{\NumRounds}\) be any possibly random
sequence satisfying \(a'_t\in\EligibleArms{t}\) at every round.
On every realization,
\[
\sum_{t=1}^{\NumRounds}\ConfRadius{a'_t}{t}
\leq
\sum_{t=1}^{\NumRounds}
\max_{a\in\EligibleArms{t}}\ConfRadius{a}{t}.
\]
Taking conditional expectations therefore gives
\[
\E\!\left[
\left.
\sum_{t=1}^{\NumRounds}\ConfRadius{a'_t}{t}
\,\right|\ConcentrationEvent
\right]
\leq
64\,\log(2\NumRounds\NumArms)
\sqrt{\NumClasses\NumRounds\TotalDifficulty}.
\]
No independence assumption on the choices \(a'_t\) is needed,
because the comparison with the maximum holds on every realization.
\end{proof}

\subsection{Regret decomposition and its three contributions}
\label{app:regret-terms}

The confidence bounds now apply to both representative arms. It remains
to separate the comparisons they can control directly from the cost of
sampling bad arms and the residual left by an outdated optimal-arm
certificate.

\begin{lemma}[Three-term regret decomposition]
\label{lem:ebfh-regret-decomposition}
For the representatives in \Cref{def:ebfh-representatives}, define
\begin{align*}
R_1&:=\sum_{t=1}^{\NumRounds}
\bigl(\MeanReward{\RepresentativeArm{t}}{t}
-\MeanReward{\PlayedArm{t}}{t}\bigr),\\
R_2&:=\sum_{t=1}^{\NumRounds}
\bigl(\MeanReward{\RepresentativeOptimalArm{t}}{t}
-\MeanReward{\RepresentativeArm{t}}{t}\bigr),\\
R_3&:=\sum_{t=1}^{\NumRounds}
\bigl(\MeanReward{\OptimalArm{t}}{t}
-\MeanReward{\RepresentativeOptimalArm{t}}{t}\bigr).
\end{align*}
Then \(\DynamicRegret=R_1+R_2+R_3\) on every realization.
The three sums correspond to Terms~(i), (ii), and (iii) of
\eqref{eq:regret-decomposition}; their summands need not all be
nonnegative.
\end{lemma}

\begin{proof}
For every round \(t\), add and subtract the means of the two
representatives:
\begin{align*}
\MeanReward{\OptimalArm{t}}{t}-\MeanReward{\PlayedArm{t}}{t}
={}\bigl(\MeanReward{\RepresentativeArm{t}}{t}
-\MeanReward{\PlayedArm{t}}{t}\bigr)
+\bigl(\MeanReward{\RepresentativeOptimalArm{t}}{t}
-\MeanReward{\RepresentativeArm{t}}{t}\bigr)
+\bigl(\MeanReward{\OptimalArm{t}}{t}
-\MeanReward{\RepresentativeOptimalArm{t}}{t}\bigr).
\end{align*}
Summing over \(t\in[\NumRounds]\) proves the identity. 
\end{proof}

We bound each contribution from above. The good representatives are
already covered by the eligible-arm confidence bound. For a bad arm,
the historical dominance gap must be accompanied by confidence
corrections before it can be compared with current means.

\subsubsection{Term~(i): monitoring bad arms}
\label{app:proof-term-i}

\begin{lemma}[Restatement of \Cref{lem:term-i}]
For the concentration event \(\ConcentrationEvent\),
\[
\E\!\left[\left.\sum_{t=1}^\NumRounds
 [\MeanReward{\RepresentativeArm{t}}{t}-\MeanReward{\PlayedArm{t}}{t}]\,\right|\,\ConcentrationEvent\right]
\leq \BigOTilde\!\left(
\sqrt{\NumClasses\NumRounds\TotalDifficulty}\right).
\]
\end{lemma}

\begin{proof}
If the played arm is good, then
\(\RepresentativeArm{t}=\PlayedArm{t}\) and the summand is zero. We therefore
fix a round \(t\) on which \(\PlayedArm{t}\in\BadArms{t}\). By
\Cref{def:ebfh-representatives},
\(\RepresentativeArm{t}=\BlockingArm{t}{\PlayedArm{t}}\), and
\Cref{prop:blocking-is-good} gives
\(\RepresentativeArm{t}\in\GoodArms{t}\).

The blocking certificate implies
\begin{align}
\DominanceGap{t}{\PlayedArm{t}}
={}&
\EmpBound{\RepresentativeArm{t}}{-1}
{\BlockingCertificateInterval{t}{\PlayedArm{t}}}
-\EmpBound{\PlayedArm{t}}{+1}
{\ArmCertificateInterval{t}{\PlayedArm{t}}}\notag\\
\geq{}&
2\ConfRadius{\RepresentativeArm{t}}
{\BlockingCertificateInterval{t}{\PlayedArm{t}}}
+2\ConfRadius{\PlayedArm{t}}
{\ArmCertificateInterval{t}{\PlayedArm{t}}}.
\label{eq:term-i-gap-lower}
\end{align}
Using the first line of this display, we have the exact decomposition
\begin{align}
&\MeanReward{\RepresentativeArm{t}}{t}
-\MeanReward{\PlayedArm{t}}{t}\notag\\
={}&
\underbrace{\MeanReward{\RepresentativeArm{t}}{t}
-\EmpBound{\RepresentativeArm{t}}{-1}
{\BlockingCertificateInterval{t}{\PlayedArm{t}}}
-\DominanceGap{t}{\PlayedArm{t}}}_{\text{(1a)}}\notag\\
&\quad+\underbrace{3\DominanceGap{t}{\PlayedArm{t}}}_{\text{(1b)}}\notag\\
&\quad+\underbrace{\EmpBound{\PlayedArm{t}}{+1}
{\ArmCertificateInterval{t}{\PlayedArm{t}}}
-\DominanceGap{t}{\PlayedArm{t}}
-\MeanReward{\PlayedArm{t}}{t}}_{\text{(1c)}}.
\label{eq:term-i-decomp}
\end{align}

The certificate and the unit reward range imply
\[
0\le
\EmpBound{\PlayedArm{t}}{+1}{\ArmCertificateInterval{t}{\PlayedArm{t}}}
<\EmpBound{\RepresentativeArm{t}}{-1}
{\BlockingCertificateInterval{t}{\PlayedArm{t}}}
\le1.
\]
Thus each of (1a) and (1c) is at most one, since
\(\DominanceGap{t}{\PlayedArm{t}}>0\) and both true means lie in
\([0,1]\). If the corresponding arm's MSI history is empty, its
confidence radius is at least one, proving the claimed bound by twice
that radius. Below we therefore consider nonempty MSI histories.

\textbf{Bounding Term (1a).}
The concentration event on the MSI of \(\RepresentativeArm{t}\) gives
\(\MeanReward{\RepresentativeArm{t}}{t}
\leq\EmpBound{\RepresentativeArm{t}}{+1}{t}\). Hence
\begin{align*}
\text{(1a)}
&\leq
\EmpBound{\RepresentativeArm{t}}{+1}{t}
-\EmpBound{\RepresentativeArm{t}}{-1}
{\BlockingCertificateInterval{t}{\PlayedArm{t}}}
-\DominanceGap{t}{\PlayedArm{t}}\\
&\leq
\EmpBound{\RepresentativeArm{t}}{+1}{t}
-\EmpBound{\RepresentativeArm{t}}{-1}
{\BlockingCertificateInterval{t}{\PlayedArm{t}}}
-2\ConfRadius{\RepresentativeArm{t}}
{\BlockingCertificateInterval{t}{\PlayedArm{t}}}\\
&=
\EmpBound{\RepresentativeArm{t}}{-1}{t}
-\EmpBound{\RepresentativeArm{t}}{+1}
{\BlockingCertificateInterval{t}{\PlayedArm{t}}}
+2\ConfRadius{\RepresentativeArm{t}}{t}\\
&\leq2\ConfRadius{\RepresentativeArm{t}}{t}.
\end{align*}
The final inequality is the contrapositive of the change-detection rule,
applied to the two displayed intervals in the current phase of
\(\RepresentativeArm{t}\). Since
\(\RepresentativeArm{t}\in\GoodArms{t}\subseteq\EligibleArms{t}\),
\Cref{lem:proof-eligible-cb-sums} bounds the conditional expectation of the sum of
(1a) by
\(
\BigOTilde\!\left(\sqrt{\NumClasses\NumRounds\TotalDifficulty}\right).
\)

\textbf{Bounding Term (1c).}
The same argument for \(\PlayedArm{t}\), using
\eqref{eq:term-i-gap-lower}, gives
\begin{align*}
\text{(1c)}
&\leq
\EmpBound{\PlayedArm{t}}{+1}
{\ArmCertificateInterval{t}{\PlayedArm{t}}}
-\DominanceGap{t}{\PlayedArm{t}}
-\EmpBound{\PlayedArm{t}}{-1}{t}\\
&\leq
\EmpBound{\PlayedArm{t}}{-1}
{\ArmCertificateInterval{t}{\PlayedArm{t}}}
-\EmpBound{\PlayedArm{t}}{+1}{t}
+2\ConfRadius{\PlayedArm{t}}{t}\\
&\leq2\ConfRadius{\PlayedArm{t}}{t}.
\end{align*}
The played-arm confidence-sum bound, \Cref{lem:played-cb-sum}, therefore
bounds the conditional expectation of the sum of (1c) by
\(
\BigOTilde\!\left(\sqrt{\NumRounds\TotalDifficulty}\right).
\)

\textbf{Bounding Term (1b).}
Because the bad arm \(\PlayedArm{t}\) is eligible, let
\(j\in\{0,\ldots,\MaxScale\}\) be the smallest scale such
that \(j\leq\DominanceScale{t}{\PlayedArm{t}}\) and
\(\BackfillCounter{\PlayedArm{t}}{j}>0\) after the round-\(t\) refreshes
and before the play. By the
definition of the dominance scale,
\(
\DominanceGap{t}{\PlayedArm{t}}
\leq2^{-\DominanceScale{t}{\PlayedArm{t}}+1}
\leq2^{-j+1},
\)
so \(3\DominanceGap{t}{\PlayedArm{t}}\leq6\cdot2^{-j}\). Assign this
amount to the decrease of \(\BackfillCounter{\PlayedArm{t}}{j}\) caused by
the play.

Every such decrease has a unique most recent update of the same counter. An
update sets the counter to
$2^{2j+5}\LogFactor$, so it can be followed by at
most this many decreases before the next update. If a later update overwrites
a positive value, unused counter units are discarded, which can only reduce
the number of assigned decreases. Therefore, for every realization of the
algorithm's random choices, the total dominance-gap charge assigned to one
update at scale $j$ is at most
\(
2^{2j+5}\LogFactor
\cdot 6\cdot2^{-j}.
\)

The algorithm maintains only scales
\(j\in\{0,\ldots,\MaxScale\}\). Because the counter-update
coins are independent of \(\ConcentrationEvent\), their probabilities are
unchanged after conditioning on that event. Thus the conditional expected charge generated by the updates at any one
round is at most
\begin{align*}
&\sum_{m=1}^{\NumClasses}\sum_{a\in\ClassArms{m}}\sum_{j=0}^{\MaxScale}
2^{-j}\sqrt{\frac{\ChangeBudget{m} + 1}{\NumRounds \ClassSize{m}}}
\cdot 2^{2j+5}\LogFactor
\cdot 6 \cdot 2^{-j} \\
&= 6 \cdot 2^5 \LogFactor \cdot (\MaxScale + 1) \sum_{m\in[\NumClasses]} \ClassSize{m} \sqrt{\frac{\ChangeBudget{m} + 1}{\NumRounds \ClassSize{m}}} \\
&= 6 \cdot 2^5 \LogFactor \cdot (\MaxScale + 1) \sum_{m\in[\NumClasses]} \sqrt{\frac{(\ChangeBudget{m} + 1) \ClassSize{m}}{\NumRounds}} \\
&\leq 6 \cdot 2^5 \LogFactor \cdot (\MaxScale + 1) \cdot \sqrt{\frac{\NumClasses\TotalDifficulty}{\NumRounds}},
\end{align*}
where the last step is Cauchy--Schwarz. Summing this bound over all rounds
gives
\begin{align*}
\E\!\left[
\left.
\sum_{\substack{t\in[\NumRounds]:\\\PlayedArm{t}\in\BadArms{t}}}
3\DominanceGap{t}{\PlayedArm{t}}
\right|\,\ConcentrationEvent
\right]
\leq
6\cdot2^5\LogFactor
(\MaxScale+1)
\sqrt{\NumClasses\NumRounds\TotalDifficulty}
=\BigOTilde\!\left(
\sqrt{\NumClasses\NumRounds\TotalDifficulty}\right).
\end{align*}

Combining Terms (1a), (1b), (1c) via \eqref{eq:term-i-decomp} completes the proof.
\end{proof}

Only the dominance-gap part was charged to counter refreshes. The
comparison with the blocker's current mean used the eligible-arm
confidence sum, and the comparison with the played arm's current mean
used the played-arm sum. The next term has no counter charge, because
both arms in the comparison are good.

\subsubsection{Term~(ii): comparing good representatives}
\label{app:proof-term-ii}

\begin{lemma}[Restatement of \Cref{lem:term-ii}]
For the concentration event \(\ConcentrationEvent\),
\[
\E\!\left[\left.\sum_{t=1}^\NumRounds
 [\MeanReward{\RepresentativeOptimalArm{t}}{t}-\MeanReward{\RepresentativeArm{t}}{t}]\,\right|\,\ConcentrationEvent\right]
\leq \BigOTilde\!\left(
\sqrt{\NumClasses\NumRounds\TotalDifficulty}\right).
\]
\end{lemma}

\begin{proof}
For every round \(t\), both \(\RepresentativeArm{t}\) and
\(\RepresentativeOptimalArm{t}\) are good by
\Cref{prop:blocking-is-good}. If they are equal, the summand is zero.
Otherwise, the concentration event applied to the MSI of each arm gives
\begin{align*}
\MeanReward{\RepresentativeOptimalArm{t}}{t}
-\MeanReward{\RepresentativeArm{t}}{t}
&\leq
\EmpBound{\RepresentativeOptimalArm{t}}{+1}{t}
-\EmpBound{\RepresentativeArm{t}}{-1}{t}\\
&=
\EmpMean{\RepresentativeOptimalArm{t}}{t}
-\EmpMean{\RepresentativeArm{t}}{t}
+\ConfRadius{\RepresentativeOptimalArm{t}}{t}
+\ConfRadius{\RepresentativeArm{t}}{t}\\
&\leq
4\ConfRadius{\RepresentativeOptimalArm{t}}{t}
+4\ConfRadius{\RepresentativeArm{t}}{t}.
\end{align*}
The final inequality follows because the goodness of
\(\RepresentativeArm{t}\) precludes the two MSI-history intervals from
certifying it as bad:
\[
\EmpBound{\RepresentativeOptimalArm{t}}{-3}{t}
\leq\EmpBound{\RepresentativeArm{t}}{+3}{t}.
\]
If either MSI history is empty, the conclusion follows instead from the unit
reward range and the fact that the corresponding confidence bound is at
least one.

Both representatives belong to
\(\GoodArms{t}\subseteq\EligibleArms{t}\). Taking conditional expectations
and applying \Cref{lem:proof-eligible-cb-sums} therefore gives
\begin{align*}
\E\!\left[\left.
\sum_{t=1}^{\NumRounds}
\bigl[
\MeanReward{\RepresentativeOptimalArm{t}}{t}
-\MeanReward{\RepresentativeArm{t}}{t}
\bigr]
\right|\,\ConcentrationEvent\right]
\leq4\E\!\left[\left.
\sum_{t=1}^{\NumRounds}
\bigl[
\ConfRadius{\RepresentativeOptimalArm{t}}{t}
+\ConfRadius{\RepresentativeArm{t}}{t}
\bigr]
\right|\,\ConcentrationEvent\right]
=\BigOTilde\!\left(
\sqrt{\NumClasses\NumRounds\TotalDifficulty}\right).
\end{align*}
\end{proof}

The remaining term is nonzero only when the optimal arm is classified
as bad. Its blocker is good, but the optimal arm need not be eligible.
We separate the blocker's estimation error and the class-threshold
contribution, then follow the monitoring counter to bound the remaining
certificate residuals.

\subsubsection{Term~(iii): delayed detection}
\label{app:proof-term-iii}
\label{app:proof-term-iii-hard}

\begin{lemma}[Restatement of \Cref{lem:term-iii}]
For the concentration event \(\ConcentrationEvent\),
\[
\E\!\left[\left.\sum_{t=1}^\NumRounds
 [\MeanReward{\OptimalArm{t}}{t}-\MeanReward{\RepresentativeOptimalArm{t}}{t}]
 \,\right|\,\ConcentrationEvent\right]
\leq \BigOTilde\!\left(
\sqrt{\NumClasses\NumRounds\TotalDifficulty}\right).
\]
\end{lemma}

\begin{proof}
The difficulty in bounding this term is that an optimal arm can still be classified as bad:
its certificate may rely on observations collected before its mean
changed. Its representative is good, but the optimal arm itself need
not be eligible. We therefore cannot apply the eligible-arm confidence
bound directly to both arms.

We first separate the error in the blocking arm's historical estimate
from the discrepancy in the optimal arm's certificate. The former is
controlled by the eligible-arm confidence sum. We then split the latter
into small contributions, bounded by class thresholds, and large
residuals that require monitoring. For the large residuals, we fix a
true stationary segment and a monitoring scale, and consider three
periods: before the first counter refresh, after that refresh but before
the next reset, and after the reset. We bound the waiting time in the
first period, prove eligibility on the relevant rounds in the second,
and show that no positive residual remains in the third. Finally, we
sum these contributions over classes, segments, and scales.

Throughout, pointwise inequalities are proved on
\(\ConcentrationEvent\); expectations retain the randomness of the
interaction conditional on this event.

\medskip
\noindent\textbf{Step 1: Express the loss through the selected certificate.}
Recall that
\(
\HardRounds:=\{t\in[\NumRounds]:\OptimalArm{t}\in\BadArms{t}\}
\)
is the set of rounds on which the optimal arm is classified as bad.
Outside this set, \(\RepresentativeOptimalArm{t}=\OptimalArm{t}\),
so Term~(iii) is zero.

For \(t\in\HardRounds\), the representative is the blocking arm
\(\RepresentativeOptimalArm{t}=\BlockingArm{t}{\OptimalArm{t}}\).
By \Cref{prop:blocking-is-good}, it belongs to
\(\GoodArms{t}\subseteq\EligibleArms{t}\). Let the selected certificate be
\(
\bigl(
\RepresentativeOptimalArm{t},
\ArmCertificateInterval{t}{\OptimalArm{t}},
\BlockingCertificateInterval{t}{\OptimalArm{t}}
\bigr).
\)
The second interval contains the observations used for the optimal arm;
the third contains those used for its blocking arm. Define the
certificate residual by
\[
r_t:=\MeanReward{\OptimalArm{t}}{t}
-\EmpBound{\OptimalArm{t}}{+1}
 {\ArmCertificateInterval{t}{\OptimalArm{t}}}
\qquad(t\in\HardRounds).
\]
This compares the optimal arm's current mean with its historical
certificate upper bound, not with another arm's current mean.

By the definition of the dominance gap,
\begin{equation}
\DominanceGap{t}{\OptimalArm{t}}
=
\EmpBound{\RepresentativeOptimalArm{t}}{-1}
 {\BlockingCertificateInterval{t}{\OptimalArm{t}}}
-
\EmpBound{\OptimalArm{t}}{+1}
 {\ArmCertificateInterval{t}{\OptimalArm{t}}}
\geq0.
\label{eq:optimal-blocking-gap}
\end{equation}
Subtracting this identity from the definition of \(r_t\) gives
\begin{equation}
\MeanReward{\OptimalArm{t}}{t}
-\EmpBound{\RepresentativeOptimalArm{t}}{-1}
 {\BlockingCertificateInterval{t}{\OptimalArm{t}}}
=
r_t-\DominanceGap{t}{\OptimalArm{t}}.
\label{eq:optimal-residual-identity}
\end{equation}
Adding the difference between the blocker's certificate lower bound
and its current mean to both sides recovers the Term~(iii) summand.
This separates the blocking-arm comparison from the residual difference
\(r_t-\DominanceGap{t}{\OptimalArm{t}}\).

\medskip
\noindent\textbf{Step 2: Separate small residuals from those requiring monitoring.}
For each class \(m\in[\NumClasses]\), recall the count
\[
\ClassOptimalRounds{m}
:=\bigl|\{t\in[\NumRounds]:\OptimalArm{t}\in\ClassArms{m}\}\bigr|.
\]
These counts include all rounds, not only those in \(\HardRounds\),
and satisfy \(\sum_{m=1}^{\NumClasses}\ClassOptimalRounds{m}=\NumRounds\).
Recall also \(\LogFactor=\lceil\log(2\NumRounds\NumArms)\rceil\),
where \(\log\) is the natural logarithm, and the thresholds
\[
\ClassThreshold{m}:=
\begin{cases}
\displaystyle
\LogFactor\sqrt{
\frac{\NumClasses\ClassSize{m}(\ChangeBudget{m}+1)}
     {\ClassOptimalRounds{m}}},
&\ClassOptimalRounds{m}>0,\\[1ex]
0,&\ClassOptimalRounds{m}=0.
\end{cases}
\]
If \(r_t\leq\DominanceGap{t}{\OptimalArm{t}}\), the residual
difference is nonpositive. If \(r_t\) exceeds that gap but not its
class threshold, the difference is at most the threshold. We therefore
need the monitoring argument only for the remaining residuals, indexed by
\[
\LargeResidualRounds
:=\left\{t\in\HardRounds:
r_t>\max\left\{
\DominanceGap{t}{\OptimalArm{t}},
\ClassThreshold{\ArmClass{\OptimalArm{t}}}
\right\}\right\}.
\]
On this set, the residual difference is at most \(r_t\).
There are at most \(\ClassOptimalRounds{m}\) threshold contributions
from class \(m\). Summing \eqref{eq:optimal-residual-identity} after
adding the blocking-arm comparison therefore yields
\begin{align}
&\sum_{t=1}^{\NumRounds}
\bigl[\MeanReward{\OptimalArm{t}}{t}
      -\MeanReward{\RepresentativeOptimalArm{t}}{t}\bigr]
\notag\\
&\quad\leq
\underbrace{\sum_{t\in\HardRounds}
\left[
\EmpBound{\RepresentativeOptimalArm{t}}{-1}
 {\BlockingCertificateInterval{t}{\OptimalArm{t}}}
-\MeanReward{\RepresentativeOptimalArm{t}}{t}
\right]}_{\text{(3a): blocking-arm comparison}}
\notag\\
&\qquad+
\underbrace{\sum_{m=1}^{\NumClasses}
\ClassOptimalRounds{m}\ClassThreshold{m}}_{\text{(3b): class thresholds}}
+
\underbrace{\sum_{t\in\LargeResidualRounds}r_t}
 _{\text{(3c): large residuals}}.
\label{eq:term-iii-decomp}
\end{align}
We next bound these three contributions.

\medskip
\noindent\textbf{Step 3: Bound the blocking-arm comparison and thresholds.}
For brevity, denote the conditional expected eligible-arm confidence
sum by
\[
S_{\mathrm{elig}}
:=\E\!\left[\left.
\sum_{t=1}^{\NumRounds}\max_{a\in\EligibleArms{t}}\ConfRadius{a}{t}
\,\right|\ConcentrationEvent\right].
\]
As throughout the analysis, a statistic evaluated at \(t\) uses only
observations before \(t\) in the arm's current MSI.

\emph{Term (3a).}
Fix \(t\in\HardRounds\). The blocker's class was not reset at
round \(t\): after a reset its phase history would be empty, making
its three-radius lower bound negative and hence unable to exceed the
optimal arm's nonnegative three-radius upper bound.
The reset test in \eqref{eq:reset-condition} is therefore false for
this class. Applied to its certificate interval and current MSI
history, it gives
\[
\EmpBound{\RepresentativeOptimalArm{t}}{-1}
 {\BlockingCertificateInterval{t}{\OptimalArm{t}}}
\leq
\EmpBound{\RepresentativeOptimalArm{t}}{+1}{t}
\leq
\MeanReward{\RepresentativeOptimalArm{t}}{t}
+2\ConfRadius{\RepresentativeOptimalArm{t}}{t}.
\]
The second inequality follows from concentration on the current MSI.
If that history has no observations, its upper bound is its confidence
radius, which is at least one. The first inequality then follows from
the certificate lower bound being at most one, and the second follows
from the mean being nonnegative. Thus the same comparison holds in
the zero-observation case.

Subtracting the blocker's mean, summing, and using its eligibility,
we obtain
\[
\E\!\left[\left.
\sum_{t\in\HardRounds}
\left[
\EmpBound{\RepresentativeOptimalArm{t}}{-1}
 {\BlockingCertificateInterval{t}{\OptimalArm{t}}}
-\MeanReward{\RepresentativeOptimalArm{t}}{t}
\right]
\,\right|\ConcentrationEvent\right]
\leq2S_{\mathrm{elig}}.
\]

\emph{Term (3b).}
The threshold definition and Cauchy--Schwarz give
\begin{align*}
\sum_{m=1}^{\NumClasses}\ClassOptimalRounds{m}\ClassThreshold{m}
&=\LogFactor\sqrt{\NumClasses}
\sum_{m=1}^{\NumClasses}
\sqrt{\ClassOptimalRounds{m}\ClassSize{m}(\ChangeBudget{m}+1)}\\
&\leq\LogFactor\sqrt{\NumClasses}
\sqrt{\sum_{m=1}^{\NumClasses}\ClassOptimalRounds{m}}
\sqrt{\sum_{m=1}^{\NumClasses}\ClassSize{m}(\ChangeBudget{m}+1)}\\
&=\LogFactor\sqrt{\NumClasses\NumRounds\TotalDifficulty}.
\end{align*}
The zero-count classes contribute zero in the first equality.
The last equality uses the total count \(\NumRounds\) and the
definition of \(\TotalDifficulty\). It remains to bound Term~(3c).

\medskip
\noindent\textbf{Step 4: Partition the large-residual rounds.}
Recall the scale sets from \eqref{eq:hard-residual-rounds}:
\[
\HardScaleRounds{m}{j}
=\{t\in\LargeResidualRounds:
\OptimalArm{t}\in\ClassArms{m},\quad 2^{-j}<r_t\leq2^{1-j}\},
\qquad j\geq1.
\]
On \(\LargeResidualRounds\), we have \(0<r_t\leq1\): the true
mean is at most one and the certificate upper bound is nonnegative.
Each such round therefore belongs to exactly one class and one
residual scale. The algorithm maintains scales up to
\(\MaxScale=\lceil\log_2\NumRounds\rceil\).
Residuals at most \(2^{-\MaxScale}\) have total contribution at most
\(\NumRounds\,2^{-\MaxScale}\leq1\). We treat the remaining
scales \(j\in\{1,\ldots,\MaxScale\}\) below; this set is empty
when \(\NumRounds=1\).

For class \(m\), write its true stationary segments as
\[
\NoChangeInterval{m}{i}=[b_{m,i},e_{m,i}],
\qquad
 i\in\{0,\ldots,|\ChangeRounds{m}|\},
\qquad
 d_{m,i}:=e_{m,i}-b_{m,i}+1.
\]
Thus \(b_{m,i}\), \(e_{m,i}\), and \(d_{m,i}\) are the segment's
first round, last round, and length. These segments have constant
class means but, unlike MSIs, may contain algorithmic resets.
Let \(\ClassBestArm{m}{i}\) be the best arm in class \(m\) on this
segment, using the same fixed arm order as \(\OptimalArm{t}\).
The segment and its best arm are fixed by the instance. In particular,
\[
t\in\NoChangeInterval{m}{i},\quad
\OptimalArm{t}\in\ClassArms{m}
\quad\Longrightarrow\quad
\OptimalArm{t}=\ClassBestArm{m}{i}.
\]
The global optimal arm may leave this class as other classes change;
whenever it belongs to this class, it is the same class-best arm.

Fix a class \(m\), segment index \(i\), and maintained scale \(j\).
For each \(t\in\NoChangeInterval{m}{i}\), define
\(X_{m,i,j,t}\) to be the indicator that the counter
\(\BackfillCounter{\ClassBestArm{m}{i}}{j}\) is refreshed at round
\(t\). The refresh probability is
\[
p_{m,j}:=2^{-j}
\sqrt{\frac{\ChangeBudget{m}+1}{\NumRounds\ClassSize{m}}}.
\]
These indicators are independent across rounds and independent of
\(\ConcentrationEvent\), which depends only on the presampled reward
sequences. Each refresh sets the counter to \(2^{2j+5}\LogFactor\),
whether the arm is good or bad. Since
\(\ChangeBudget{m}+1\leq\NumRounds\) and \(\ClassSize{m}\geq1\),
we have \(0<p_{m,j}\leq1\).

Let the first refresh round in this segment be
\[
\sigma_{m,i,j}
:=\min\bigl(\{t\in\NoChangeInterval{m}{i}:X_{m,i,j,t}=1\}
              \cup\{e_{m,i}+1\}\bigr).
\]
The value \(e_{m,i}+1\) means that no refresh occurs in the segment.
Let \(\eta_{m,i,j}\) be the first class-\(m\) reset round in the
segment strictly after \(\sigma_{m,i,j}\), or \(e_{m,i}+1\) if
there is no such reset. Define
\begin{align*}
\mathcal R^{(1)}_{m,i,j}
&:=\{t\in\NoChangeInterval{m}{i}:t<\sigma_{m,i,j}\},\\
\mathcal R^{(2)}_{m,i,j}
&:=\{t\in\NoChangeInterval{m}{i}:
        \sigma_{m,i,j}\leq t<\eta_{m,i,j}\},\\
\mathcal R^{(3)}_{m,i,j}
&:=\{t\in\NoChangeInterval{m}{i}:\eta_{m,i,j}\leq t\}.
\end{align*}
These regions are, respectively, before refresh, after refresh but
before the next reset, and after that reset. They partition the
segment. If no refresh occurs, only Region~1 is nonempty; if no
subsequent reset occurs, Region~3 is empty. A reset at the start
of the refresh round occurs before the refresh and does not end
Region~2.

For \(k\in\{1,2,3\}\), define the residual sum in Region~\(k\) by
\[
Z^{(k)}_{m,i,j}
:=\sum_{t\in\HardScaleRounds{m}{j}\cap\mathcal R^{(k)}_{m,i,j}}r_t.
\]
All these sums are nonnegative, and empty sums are zero. Every
large-residual round at a maintained scale has a unique class,
segment, scale, and region, so
\[
\sum_{m=1}^{\NumClasses}\sum_{j=1}^{\MaxScale}
\sum_{t\in\HardScaleRounds{m}{j}}r_t
=
\sum_{m=1}^{\NumClasses}\sum_{i=0}^{|\ChangeRounds{m}|}
\sum_{j=1}^{\MaxScale}\sum_{k=1}^{3}Z^{(k)}_{m,i,j}.
\]
We now bound these region sums in chronological order.

\medskip
\noindent\textbf{Step 5: Bound the three monitoring periods.}

\emph{Region 1: before the first refresh.}
Define the waiting count
\begin{equation}
W_{m,i,j}:=|\mathcal R^{(1)}_{m,i,j}|
          =\sigma_{m,i,j}-b_{m,i}.
\label{eq:ebfh-region-one-wait}
\end{equation}
This counts all rounds before the first refresh, excluding the
refresh round. It is zero if the first round refreshes the counter,
and equals \(d_{m,i}\) if the counter is never refreshed in the segment.
For an integer \(q\in\{1,\ldots,d_{m,i}\}\), at least \(q\)
rounds precede the refresh exactly when the first \(q\) coins fail:
\[
\{W_{m,i,j}\geq q\}
=
\bigcap_{u=b_{m,i}}^{b_{m,i}+q-1}\{X_{m,i,j,u}=0\}.
\]
Independence of the refresh coins, also after conditioning on
\(\ConcentrationEvent\), gives
\begin{equation}
\Prob(W_{m,i,j}\geq q\mid\ConcentrationEvent)
=
\begin{cases}
(1-p_{m,j})^q,&1\leq q\leq d_{m,i},\\
0,&q>d_{m,i},
\end{cases}
\label{eq:ebfh-region-one-tail}
\end{equation}
for every integer \(q\geq1\). Using the tail-sum identity
\citep[Proposition~2.8]{lattimore2020bandit},
\begin{align*}
\E[W_{m,i,j}\mid\ConcentrationEvent]
&=\sum_{q=1}^{d_{m,i}}(1-p_{m,j})^q\\
&\leq\sum_{q=1}^{\infty}(1-p_{m,j})^q
=\frac{1-p_{m,j}}{p_{m,j}}
\leq\frac{1}{p_{m,j}}.
\end{align*}
This estimate does not require a refresh to occur before the segment
ends. Each summand of \(Z^{(1)}_{m,i,j}\) is at most \(2^{1-j}\),
and there are at most \(W_{m,i,j}\) such summands. Hence, pathwise,
\(Z^{(1)}_{m,i,j}\leq2^{1-j}W_{m,i,j}\), and therefore
\[
\E[Z^{(1)}_{m,i,j}\mid\ConcentrationEvent]
\leq\frac{2^{1-j}}{p_{m,j}}
=2\sqrt{\frac{\NumRounds\ClassSize{m}}{\ChangeBudget{m}+1}}.
\]
This uses no independence between the large-residual rounds and
the refresh coins; the bound follows by counting all waiting rounds.
Summing over at most \(\ChangeBudget{m}+1\) segments in class \(m\)
and the \(\MaxScale\) maintained residual scales gives
\begin{align*}
&\E\!\left[\left.
\sum_{m=1}^{\NumClasses}\sum_{i=0}^{|\ChangeRounds{m}|}
\sum_{j=1}^{\MaxScale}Z^{(1)}_{m,i,j}
\,\right|\ConcentrationEvent\right]\\
&\quad\leq2\MaxScale\sum_{m=1}^{\NumClasses}
\sqrt{\NumRounds\ClassSize{m}(\ChangeBudget{m}+1)}\\
&\quad\leq2\MaxScale\sqrt{\NumClasses\NumRounds\TotalDifficulty}.
\end{align*}
The last inequality is Cauchy--Schwarz over the classes, using
\(\sum_m(\ChangeBudget{m}+1)\ClassSize{m}=\TotalDifficulty\).

\medskip
\noindent\emph{Region 2: after refresh and before the next reset.}
Fix a round
\(t\in\HardScaleRounds{m}{j}\cap\mathcal R^{(2)}_{m,i,j}\),
and write \(s:=\sigma_{m,i,j}\). The optimal arm at \(t\) is
\(\ClassBestArm{m}{i}\), whose scale-\(j\) counter was refreshed
at \(s\). We show first that this counter is still positive, then
that the arm is eligible, and finally that its residual is bounded
by its current-MSI confidence radius.

\emph{The counter cannot be exhausted on this round.}
Counter values here are evaluated after the round-\(t\) refreshes
and before the play. If \(t=s\), the counter is positive by its
refresh. Otherwise, suppose it were zero. Since the most recent
refresh at or after \(s\) set it to \(2^{2j+5}\LogFactor\), it
must have been decremented at least this many times before \(t\).
Each decrement requires a pull of this arm. Thus
\[
\ConfRadius{\OptimalArm{t}}{[s,t-1]}
\leq
\sqrt{\frac{8\log(2\NumRounds\NumArms)}{2^{2j+5}\LogFactor}}
\leq2^{-j-1},
\]
where the second inequality uses
\(\LogFactor\geq\log(2\NumRounds\NumArms)\).
There is no class-\(m\) change on \([s,t-1]\). Concentration and
\(r_t>2^{-j}\) therefore imply
\begin{align*}
\EmpBound{\OptimalArm{t}}{-1}{[s,t-1]}
&\geq\MeanReward{\OptimalArm{t}}{t}
      -2\ConfRadius{\OptimalArm{t}}{[s,t-1]}\\
&\geq\MeanReward{\OptimalArm{t}}{t}-2^{-j}\\
&>\EmpBound{\OptimalArm{t}}{+1}
 {\ArmCertificateInterval{t}{\OptimalArm{t}}}.
\end{align*}
Both intervals lie in the current class phase: no reset occurs
strictly after \(s\) and before or at this round in Region~2.
The displayed inequality would therefore trigger the reset test
at \(t\), contradicting membership in Region~2. Consequently,
\(\BackfillCounter{\OptimalArm{t}}{j}>0\).

\emph{The positive counter makes the optimal arm eligible.}
Membership in \(\HardScaleRounds{m}{j}\) gives
\[
0<\DominanceGap{t}{\OptimalArm{t}}<r_t\leq2^{1-j}.
\]
Taking negative base-two logarithms gives
\(-\log_2\DominanceGap{t}{\OptimalArm{t}}>j-1\), so
\[
\DominanceScale{t}{\OptimalArm{t}}
=\left\lceil-\log_2\DominanceGap{t}{\OptimalArm{t}}\right\rceil
\geq j.
\]
The arm is bad, its scale-\(j\) counter is positive, and \(j\)
is a maintained scale not exceeding its dominance scale.
These are exactly the conditions for its inclusion in
\(\EligibleArms{t}\).

\emph{The residual is bounded by a confidence radius.}
Because the optimal arm is bad at \(t\), its class was not reset
at the start of that round: a newly reset arm has no observations
to support a bad-arm certificate. In particular, a reset at \(s\)
is impossible when \(t=s\).
If \(\NumPulls{\OptimalArm{t}}{t}>0\), concentration on the
current MSI and the absence of a reset witness give, respectively,
\[
\MeanReward{\OptimalArm{t}}{t}
\leq\EmpBound{\OptimalArm{t}}{+1}{t},
\qquad
\EmpBound{\OptimalArm{t}}{-1}{t}
\leq\EmpBound{\OptimalArm{t}}{+1}
 {\ArmCertificateInterval{t}{\OptimalArm{t}}}.
\]
Subtracting the certificate upper bound from the first inequality
and using the second gives
\[
r_t\leq
\EmpBound{\OptimalArm{t}}{+1}{t}
-\EmpBound{\OptimalArm{t}}{-1}{t}
=2\ConfRadius{\OptimalArm{t}}{t}.
\]
If the current-MSI count is zero, the same conclusion follows from
\(r_t\leq1\) and \(\ConfRadius{\OptimalArm{t}}{t}\geq1\).
Since this arm is eligible, each Region~2 residual is at most twice
the largest eligible-arm radius at that round. Every relevant
round has a unique class, segment, and residual scale, so
\[
\sum_{m=1}^{\NumClasses}\sum_{i=0}^{|\ChangeRounds{m}|}
\sum_{j=1}^{\MaxScale}Z^{(2)}_{m,i,j}
\leq2\sum_{t=1}^{\NumRounds}
\max_{a\in\EligibleArms{t}}\ConfRadius{a}{t}.
\]
There is no additional factor for the number of scales: no round
is counted at two different residual scales. Taking conditional
expectations bounds this contribution by \(2S_{\mathrm{elig}}\).

\medskip
\noindent\emph{Region 3: after the reset.}
Suppose that
\(t\in\HardScaleRounds{m}{j}\cap\mathcal R^{(3)}_{m,i,j}\).
Class \(m\) has already reset within its current true stationary
segment. The optimal arm's current certificate interval must
therefore lie entirely inside that segment:
\[
\ArmCertificateInterval{t}{\OptimalArm{t}}
\subseteq[\PhaseStart{m},t-1]
\subseteq\NoChangeInterval{m}{i}.
\]
Concentration now gives
\[
\MeanReward{\OptimalArm{t}}{t}
\leq\EmpBound{\OptimalArm{t}}{+1}
 {\ArmCertificateInterval{t}{\OptimalArm{t}}},
\qquad r_t\leq0.
\]
This contradicts \(r_t>0\) on \(\HardScaleRounds{m}{j}\).
Hence the intersection is empty and \(Z^{(3)}_{m,i,j}=0\).

\medskip
\noindent\textbf{Step 6: Combine the bounds.}
The Region~1 contribution is at most
\(2\MaxScale\sqrt{\NumClasses\NumRounds\TotalDifficulty}\)
in conditional expectation, Region~2 contributes at most
\(2S_{\mathrm{elig}}\), and Region~3 contributes zero.
Adding the at-most-one contribution of residuals below the
smallest maintained scale proves
\begin{equation}
\E\!\left[\left.
\sum_{t\in\LargeResidualRounds}r_t
\,\right|\ConcentrationEvent\right]
\leq2S_{\mathrm{elig}}
+2\MaxScale\sqrt{\NumClasses\NumRounds\TotalDifficulty}+1.
\label{prop:term-iii-hard}
\end{equation}
This completes the bound on Term~(3c). Substituting it and the two
bounds from Step~3 into \eqref{eq:term-iii-decomp}, we obtain
\begin{align*}
\E\!\left[\left.
\sum_{t=1}^{\NumRounds}
\bigl[\MeanReward{\OptimalArm{t}}{t}
      -\MeanReward{\RepresentativeOptimalArm{t}}{t}\bigr]
\,\right|\ConcentrationEvent\right]
\leq
4S_{\mathrm{elig}}
+(\LogFactor+2\MaxScale)
\sqrt{\NumClasses\NumRounds\TotalDifficulty}+1.
\end{align*}
The coefficient four consists of two eligible-arm confidence sums
from the blocking-arm comparison and two from Region~2.
Finally, \Cref{lem:proof-eligible-cb-sums} bounds
\(S_{\mathrm{elig}}\) by
\(O(\log(2\NumRounds\NumArms)
\sqrt{\NumClasses\NumRounds\TotalDifficulty})\).
Since \(\LogFactor\) and \(\MaxScale\) are also
\(O(\log(2\NumRounds\NumArms))\), the preceding display proves
the claimed \(\BigOTilde(\sqrt{\NumClasses\NumRounds\TotalDifficulty})\)
bound.
\end{proof}

\subsection{Proof of the main regret theorem}

The three contributions are controlled by two confidence sums, the
counter-refresh charge, the class thresholds, and the first-refresh
waiting bound. We retain their explicit constants below and then
remove the conditioning on concentration.

\begin{theorem}[Restatement of \Cref{thm:main-formal}]
\label{app:proof-main-formal}
For every admissible instance, Algorithm~\texttt{EBF-H} satisfies
\[
\E[\DynamicRegret]
\leq
4096\log^2(2\NumRounds\NumArms)
\sqrt{\NumClasses\NumRounds\TotalDifficulty}+1.
\]
Equivalently, its expected dynamic regret is
\(\BigOTilde(\sqrt{\NumClasses\NumRounds\TotalDifficulty})\).
\end{theorem}

\begin{proof}
Fix an admissible instance. Recall that \(\NumRounds\) is the horizon,
\(\NumArms\geq2\) is the number of arms, and \(\NumClasses\geq1\) is the
number of classes. The arm set of class \(m\) is \(\ClassArms{m}\),
with \(\ClassSize{m}\) arms and change budget \(\ChangeBudget{m}\). Write
\[
\TotalDifficulty
=\sum_{m=1}^{\NumClasses}(\ChangeBudget{m}+1)\ClassSize{m},
\qquad
\ell:=\log(2\NumRounds\NumArms),
\qquad
\Gamma:=\sqrt{\NumClasses\NumRounds\TotalDifficulty}.
\]
Here and below, \(\log\) denotes the natural logarithm. Recall also the
algorithm's parameters
\[
\LogFactor=\lceil\ell\rceil,
\qquad
\MaxScale=\lceil\log_2\NumRounds\rceil.
\]
If \(\NumRounds=1\), the claim follows immediately from
\(0\leq\DynamicRegret\leq1\). We therefore assume
\(\NumRounds\geq2\).

Recall \(\ConcentrationEvent\) is the concentration event
of \Cref{lem:confidence-bounds}. We first bound regret conditional on
\(\ConcentrationEvent\), and then account for its complement.

\medskip
\noindent\textbf{Step 1: Collect the two confidence-sum bounds.}
For an arm \(a\) and round \(t\), \(\ConfRadius{a}{t}\) denotes its
confidence radius computed using observations before \(t\) in the current
\emph{maximal static interval} (MSI) of its class. An MSI starts at round
one and whenever the class undergoes a true mean change or an algorithmic
reset. Thus both the class means and its phase start remain fixed within
an MSI; see \Cref{app:known-budget-statements}.

Let \(\EligibleArms{t}\) be the algorithm's eligible set at round \(t\),
and let \(\PlayedArm{t}\) be the arm played at that round. Define the two
\emph{conditional expected} confidence sums
\[
S_{\mathrm{elig}}
:=\E\!\left[
\left.\sum_{t=1}^{\NumRounds}
\max_{a\in\EligibleArms{t}}\ConfRadius{a}{t}
\,\right|\ConcentrationEvent\right],
\qquad
S_{\mathrm{play}}
:=\E\!\left[
\left.\sum_{t=1}^{\NumRounds}\ConfRadius{\PlayedArm{t}}{t}
\,\right|\ConcentrationEvent\right].
\]
On \(\ConcentrationEvent\), \Cref{lem:static-intervals} bounds the total
number of arm--MSI pairs by
\[
\sum_{m=1}^{\NumClasses}(2\ChangeBudget{m}+1)\ClassSize{m}
\leq2\TotalDifficulty.
\]
The explicit bound proved in \Cref{lem:proof-eligible-cb-sums} therefore
gives
\[
S_{\mathrm{elig}}
\leq32\sqrt{\ell(\MaxScale+1)}\,\Gamma.
\]
For the played arms, \Cref{lem:played-cb-sum} gives, on every realization
in \(\ConcentrationEvent\),
\[
\sum_{t=1}^{\NumRounds}\ConfRadius{\PlayedArm{t}}{t}
\leq3\sqrt{8\ell\NumRounds
\sum_{m=1}^{\NumClasses}(2\ChangeBudget{m}+1)\ClassSize{m}}
\leq12\sqrt{\ell\NumRounds\TotalDifficulty}.
\]
Taking conditional expectations and using \(\NumClasses\geq1\) yields
\[
S_{\mathrm{play}}\leq12\sqrt{\ell}\,\Gamma.
\]

\medskip
\noindent\textbf{Step 2: Separate the three regret contributions.}
Write \(\OptimalArm{t}\) for the optimal arm at round \(t\), and
\(\MeanReward{a}{t}\) for the mean reward of arm \(a\) at that round.
Recall that \(\RepresentativeArm{t}\) equals \(\PlayedArm{t}\) when the
played arm is good and equals its blocking arm otherwise.
The representative \(\RepresentativeOptimalArm{t}\) is defined in the
same way for \(\OptimalArm{t}\). By \Cref{prop:blocking-is-good}, both
representatives are good and hence belong to \(\EligibleArms{t}\).

Recall the cumulative versions of Terms~(i), (ii), and (iii):
\begin{align*}
R_1&:=\sum_{t=1}^{\NumRounds}
\bigl(\MeanReward{\RepresentativeArm{t}}{t}
      -\MeanReward{\PlayedArm{t}}{t}\bigr),\\
R_2&:=\sum_{t=1}^{\NumRounds}
\bigl(\MeanReward{\RepresentativeOptimalArm{t}}{t}
      -\MeanReward{\RepresentativeArm{t}}{t}\bigr),\\
R_3&:=\sum_{t=1}^{\NumRounds}
\bigl(\MeanReward{\OptimalArm{t}}{t}
      -\MeanReward{\RepresentativeOptimalArm{t}}{t}\bigr).
\end{align*}
\Cref{lem:ebfh-regret-decomposition} gives the pathwise
identity
\[
\DynamicRegret=R_1+R_2+R_3.
\]
The individual contributions need not all be nonnegative; we bound each
one from above.

\medskip
\noindent\textbf{Step 3: Bound Term~(i), the cost of playing bad arms.}
Let \(\BadArms{t}\) denote the bad-arm set, and let
\(\DominanceGap{t}{a}\) be the dominance gap of a bad arm \(a\), defined
by its maximizing classification certificate in \Cref{sec:algorithm}.
Term~(i) is zero when \(\PlayedArm{t}\notin\BadArms{t}\).
Otherwise, the pointwise bound proved in \Cref{app:proof-term-i} is
\[
\MeanReward{\RepresentativeArm{t}}{t}
-\MeanReward{\PlayedArm{t}}{t}
\leq
2\ConfRadius{\RepresentativeArm{t}}{t}
+3\DominanceGap{t}{\PlayedArm{t}}
+2\ConfRadius{\PlayedArm{t}}{t}.
\]
The first confidence radius is at most the largest eligible-arm radius,
because \(\RepresentativeArm{t}\in\EligibleArms{t}\).

The counter-assignment argument in \Cref{app:proof-term-i} bounds the
remaining dominance-gap contribution by
\begin{align*}
\E\!\left[
\left.\sum_{\substack{1\leq t\leq\NumRounds\\
                     \PlayedArm{t}\in\BadArms{t}}}
3\DominanceGap{t}{\PlayedArm{t}}
\,\right|\ConcentrationEvent\right]
&\leq6\cdot2^5\LogFactor(\MaxScale+1)
\sum_{m=1}^{\NumClasses}
\sqrt{\NumRounds(\ChangeBudget{m}+1)\ClassSize{m}}\\
&\leq192\LogFactor(\MaxScale+1)\Gamma.
\end{align*}
The last inequality is Cauchy--Schwarz over the classes. Summing the
pointwise bound and taking conditional expectations therefore yields
\[
\E[R_1\mid\ConcentrationEvent]
\leq2S_{\mathrm{elig}}+2S_{\mathrm{play}}
+192\LogFactor(\MaxScale+1)\Gamma.
\]

\medskip
\noindent\textbf{Step 4: Bound Term~(ii), the comparison of good representatives.}
The pointwise comparison proved in \Cref{app:proof-term-ii} gives
\[
\MeanReward{\RepresentativeOptimalArm{t}}{t}
-\MeanReward{\RepresentativeArm{t}}{t}
\leq
4\ConfRadius{\RepresentativeOptimalArm{t}}{t}
+4\ConfRadius{\RepresentativeArm{t}}{t}.
\]
Both representatives are eligible. Each radius is therefore at most
\(\max_{a\in\EligibleArms{t}}\ConfRadius{a}{t}\), and hence
\[
\E[R_2\mid\ConcentrationEvent]\leq8S_{\mathrm{elig}}.
\]

\medskip
\noindent\textbf{Step 5: Bound Term~(iii), the cost of delayed detection.}
We recall the residual notation from \Cref{app:known-budget-statements}.
Let
\[
\HardRounds:=\{t\in\{1,\ldots,\NumRounds\}:
               \OptimalArm{t}\in\BadArms{t}\}.
\]
For \(t\in\HardRounds\), let
\(\ArmCertificateInterval{t}{\OptimalArm{t}}\) be the optimal arm's
certificate interval. For any arm \(a\) and interval \(I\),
\(\EmpBound{a}{+1}{I}\) denotes the empirical mean of arm \(a\) on
\(I\) plus its confidence radius. Define
\[
r_t:=\MeanReward{\OptimalArm{t}}{t}
-\EmpBound{\OptimalArm{t}}{+1}
 {\ArmCertificateInterval{t}{\OptimalArm{t}}}.
\]
Thus \(r_t\) is the difference between the optimal arm's current mean and
its historical certificate upper bound, not its instantaneous regret.

For each class \(m\), recall the optimal-class count and threshold
\[
\ClassOptimalRounds{m}
:=\bigl|\{t\in\{1,\ldots,\NumRounds\}:
                  \OptimalArm{t}\in\ClassArms{m}\}\bigr|,
\qquad
\ClassThreshold{m}:=
\begin{cases}
\displaystyle
\LogFactor\sqrt{
\frac{\NumClasses\ClassSize{m}(\ChangeBudget{m}+1)}
     {\ClassOptimalRounds{m}}},
&\ClassOptimalRounds{m}>0,\\[1ex]
0,&\ClassOptimalRounds{m}=0.
\end{cases}
\]
Each round has a unique optimal arm under the fixed tie-breaking rule, so
\(\sum_{m=1}^{\NumClasses}\ClassOptimalRounds{m}=\NumRounds\).
Write \(\ArmClass{a}\) for the class containing arm \(a\), and define
\[
\LargeResidualRounds
:=\left\{t\in\HardRounds:
 r_t>\max\left\{
 \DominanceGap{t}{\OptimalArm{t}},
 \ClassThreshold{\ArmClass{\OptimalArm{t}}}
 \right\}\right\}.
\]

For \(t\in\HardRounds\), the blocking-certificate comparison in
\Cref{app:proof-term-iii} gives
\[
\MeanReward{\OptimalArm{t}}{t}
-\MeanReward{\RepresentativeOptimalArm{t}}{t}
\leq
r_t-\DominanceGap{t}{\OptimalArm{t}}
+2\ConfRadius{\RepresentativeOptimalArm{t}}{t}.
\]
Outside \(\HardRounds\), Term~(iii) is zero. If
\(r_t\leq\DominanceGap{t}{\OptimalArm{t}}\), the residual difference in
this display is nonpositive. If the residual exceeds the dominance gap
but not the class threshold, it is bounded by that threshold. All other
rounds belong to \(\LargeResidualRounds\), where we upper-bound the
residual difference by \(r_t\). Consequently,
\[
\E[R_3\mid\ConcentrationEvent]
\leq2S_{\mathrm{elig}}
+\sum_{m=1}^{\NumClasses}\ClassOptimalRounds{m}\ClassThreshold{m}
+\E\!\left[
\left.\sum_{t\in\LargeResidualRounds}r_t
\,\right|\ConcentrationEvent\right].
\]

\emph{The threshold contribution.}
Using the threshold definition and Cauchy--Schwarz,
\begin{align*}
\sum_{m=1}^{\NumClasses}\ClassOptimalRounds{m}\ClassThreshold{m}
&=\LogFactor\sqrt{\NumClasses}
\sum_{m=1}^{\NumClasses}
\sqrt{\ClassOptimalRounds{m}\ClassSize{m}(\ChangeBudget{m}+1)}\\
&\leq\LogFactor\sqrt{\NumClasses}
\sqrt{\sum_{m=1}^{\NumClasses}\ClassOptimalRounds{m}}
\sqrt{\sum_{m=1}^{\NumClasses}\ClassSize{m}(\ChangeBudget{m}+1)}\\
&=\LogFactor\Gamma.
\end{align*}

\emph{The large-residual contribution.}
We use \eqref{prop:term-iii-hard}, proved within \Cref{app:proof-term-iii}.
Group rounds in \(\LargeResidualRounds\) by their residual scale
\(j\), defined by \(2^{-j}<r_t\leq2^{1-j}\). For each maintained
scale \(j\in\{1,\ldots,\MaxScale\}\) and true class-stationary
segment, the proof separates rounds before the first scale-\(j\) counter
refresh, rounds after that refresh but before the next class reset, and
rounds after that reset.

Before the first refresh, the conditional expected residual contribution
of each class-\(m\) segment and scale is at most
\[
2\sqrt{\frac{\NumRounds\ClassSize{m}}{\ChangeBudget{m}+1}}.
\]
Since class \(m\) has at most
\(\ChangeBudget{m}+1\) true stationary segments, summing gives
\[
2\MaxScale\sum_{m=1}^{\NumClasses}
\sqrt{\NumRounds\ClassSize{m}(\ChangeBudget{m}+1)}
\leq2\MaxScale\Gamma.
\]
On the large-residual rounds at this scale after the refresh and before a
class reset, the optimal arm is eligible and its residual is at most twice
its current MSI confidence radius. Each large-residual round has a unique
residual scale, so the total conditional expected contribution of these
rounds is at most \(2S_{\mathrm{elig}}\), with
\emph{no additional scale factor}. After a class reset within the segment,
concentration on the new phase makes the residual nonpositive whenever the
optimal arm belongs to this class. Thus no large-residual contribution
from this class remains in that segment. Finally, residuals at most
\(2^{-\MaxScale}\) sum to at most
\(\NumRounds\,2^{-\MaxScale}\leq1\).
Combining these contributions gives
\[
\E\!\left[
\left.\sum_{t\in\LargeResidualRounds}r_t
\,\right|\ConcentrationEvent\right]
\leq2S_{\mathrm{elig}}+2\MaxScale\Gamma+1.
\]
Substituting the threshold and residual bounds into the bound for
\(R_3\), we obtain
\[
\E[R_3\mid\ConcentrationEvent]
\leq4S_{\mathrm{elig}}+(\LogFactor+2\MaxScale)\Gamma+1.
\]
The coefficient four consists of two confidence-sum contributions from
the blocking-arm comparison and two from the post-refresh residuals.

\medskip
\noindent\textbf{Step 6: Combine the three bounds and simplify the constants.}
Adding the bounds for \(R_1,R_2,R_3\) gives
\begin{align*}
\E[\DynamicRegret\mid\ConcentrationEvent]
&\leq(2+8+4)S_{\mathrm{elig}}+2S_{\mathrm{play}}\\
&\quad+
\bigl[192\LogFactor(\MaxScale+1)+\LogFactor+2\MaxScale\bigr]\Gamma+1.
\end{align*}
Using Step~1 and the identities \(14\cdot32=448\) and
\(2\cdot12=24\), this becomes
\begin{align*}
\E[\DynamicRegret\mid\ConcentrationEvent]
\leq{}\Bigl[
448\sqrt{\ell(\MaxScale+1)}+24\sqrt{\ell}
+192\LogFactor(\MaxScale+1)+\LogFactor+2\MaxScale
\Bigr]\Gamma+1.
\end{align*}
Since \(\NumArms\geq2\), \(\NumRounds\geq1\), and
\(\TotalDifficulty\geq\NumArms\), we have
\[
\ell\geq\log4>1,
\qquad
\LogFactor\leq2\ell,
\qquad
\MaxScale+1\leq3\ell,
\qquad
\Gamma\geq1.
\]
Hence,
\(192\LogFactor(\MaxScale+1)\leq1152\ell^2\).
For the remaining terms, we use
\(\sqrt{\ell}\leq\ell\leq\ell^2\) and
\(2\MaxScale\leq6\ell\). The additive one can also be absorbed because
\(1\leq\ell^2\Gamma\). Therefore,
\begin{align*}
\E[\DynamicRegret\mid\ConcentrationEvent]
&\leq
\bigl(448\sqrt3+24+1152+2+6\bigr)\ell^2\Gamma+1\\
&\leq\bigl(448\sqrt3+1185\bigr)\ell^2\Gamma\\
&\leq4096\ell^2\Gamma.
\end{align*}

\medskip
\noindent\textbf{Step 7: Account for failure of concentration.}
By \Cref{lem:confidence-bounds},
\(\Prob(\ConcentrationEvent^c)\leq\NumRounds^{-1}\), where
\(\ConcentrationEvent^c\) denotes the complementary event.
Every instantaneous regret lies in \([0,1]\), so
\(0\leq\DynamicRegret\leq\NumRounds\) on every realization.
Splitting the expectation according to whether
\(\ConcentrationEvent\) occurs gives
\begin{align*}
\E[\DynamicRegret]
&\leq\Prob(\ConcentrationEvent)
       \E[\DynamicRegret\mid\ConcentrationEvent]
       +\NumRounds\Prob(\ConcentrationEvent^c)\\
&\leq4096\ell^2\Gamma+1\\
&=4096\log^2(2\NumRounds\NumArms)
\sqrt{\NumClasses\NumRounds\TotalDifficulty}+1,
\end{align*}
as stated.
\end{proof}

\section{Proofs for Section~\ref{sec:lower-bounds}}
\label{app:lower-bound-details}

We prove the two lower bounds from Section~\ref{sec:lower-bounds}.
Both are built from the same basic statistical obstacle: over a block of
\(b\) rounds, identifying one slightly better arm among \(k\) candidates
costs \(\Omega(\sqrt{kb})\) regret. The direct-sum lower bound repeats
this difficulty independently across classes, while the class-search lower
bound additionally hides which class contains the improved arm.

All constructions below use Bernoulli rewards and are sampled before
interaction, and hence satisfy the stochastic model of
\Cref{sec:preliminaries}. The results apply to arbitrary randomized
learners, even when the class budgets are known.

\subsection{Auxiliary Lower-Bound Tools}
\label{app:lower-bound-tools}

We first prove the block-level lower bound used in both constructions.
For \(p,q\in(0,1)\), write
\[
\BernoulliKL{p}{q}
:=
p\log\frac pq+(1-p)\log\frac{1-p}{1-q}.
\]
For \(0<\epsilon\le1/4\), we use the elementary bounds
\[
\BernoulliKL{1/2}{1/2+\epsilon}\le4\epsilon^2,
\qquad
\BernoulliKL{1/2-\epsilon}{1/2+\epsilon}\le16\epsilon^2.
\]

\subsubsection{Axillary Lower Bounds}
\label{app:proof-conditional-block-lower-bound}

We first prove a conditional Bernoulli version of the standard minimax
lower bound for stochastic multi-armed bandits
\citep[Chapter~15]{lattimore2020bandit}; see also
\citet{auer2002nonstochastic}. We include it because we need
a version that can be applied conditionally after an arbitrary preceding
history and, for singleton classes, relative to an outside baseline arm.

\begin{lemma}[Conditional block lower bound]
\label{lem:conditional-block-lower-bound}
Fix any candidate set \(S\) of \(k\ge1\) arms and a block of \(b\ge k\)
rounds. If \(k=1\), assume that the bandit contains at least one arm outside
\(S\). There is a prior over Bernoulli mean assignments during the block
for which every possibly randomized learner has Bayes regret at least
\[
\frac{1}{32\sqrt2}\sqrt{kb}.
\]
The statement remains valid conditional on any preceding history
independent of the random parameter assigned to the block.

Moreover, if \(S\) is one class and the block is bracketed by baseline
rounds on which every arm has mean \(1/2\), every assignment in the prior
creates exactly two change rounds in that class.
\end{lemma}

\begin{proof}
We first consider \(k\ge2\). Choose one arm \(J\) uniformly from \(S\).
During the block, arm \(J\) has mean \(1/2+\epsilon\), while every other
arm has mean \(1/2\).

For each \(j\in[k]\), let \(P_j\) be the block-transcript law when arm
\(j\) is improved, let \(Z_j\in[0,b]\) be its number of pulls, and let
\(\E_j\) denote expectation under \(P_j\). Let \(P_0\) be the
all-baseline law and let \(\E_0\) denote its expectation.

Once the transcript preceding the block is fixed, the learner follows the
same decision rule under \(P_0\) and under every \(P_j\). The standard
bandit divergence decomposition
\citep[Lemma~15.1]{lattimore2020bandit} therefore gives
\[
\KLDivergence{P_0}{P_j}
=
\E_0 Z_j\,\BernoulliKL{1/2}{1/2+\epsilon}
\le
4\epsilon^2\E_0Z_j.
\]
Since \(0\le Z_j\le b\), Pinsker's inequality and the standard
expectation--total-variation bound
\citep[Eq.~(14.12)]{lattimore2020bandit} imply
\[
\E_jZ_j
\le
\E_0Z_j+b\TotalVariation{P_j}{P_0}
\le
\E_0Z_j+b\epsilon\sqrt{2\E_0Z_j}.
\]
Because only one arm is pulled in each round,
\[
\sum_{j=1}^k\E_0Z_j\le b.
\]
Averaging over \(j\) and applying Cauchy--Schwarz gives
\[
\frac1k\sum_{j=1}^k\E_jZ_j
\le
\frac bk+b\epsilon\sqrt{\frac{2b}{k}}.
\]
Choose
\[
\epsilon
=
\frac{1}{8\sqrt2}\sqrt{\frac{k}{b}}.
\]
Since \(b\ge k\), we have \(\epsilon\le1/4\), and since \(k\ge2\),
the preceding display is at most \(5b/8\). Hence the Bayes regret is
\[
\frac1k\sum_{j=1}^k
\epsilon\bigl(b-\E_jZ_j\bigr)
\ge
\frac{\epsilon b}{4}
=
\frac{1}{32\sqrt2}\sqrt{kb}.
\]

It remains to handle \(k=1\). Let the unique arm in \(S\) have mean
\(1/2+\epsilon\) or \(1/2-\epsilon\), each with probability \(1/2\),
while every arm outside \(S\) remains at mean \(1/2\). Set
\[
\epsilon=\frac{1}{8\sqrt{2b}}.
\]
Let \(P_+\) and \(P_-\) be the corresponding transcript laws, and let
\(Z\) denote the number of pulls of the candidate arm. Under \(P_+\),
the expected regret is \(\epsilon(b-\E_+Z)\), whereas under \(P_-\)
it is \(\epsilon\E_-Z\). Moreover,
\[
\KLDivergence{P_-}{P_+}
\le
16\epsilon^2\E_-Z
\le
16\epsilon^2b.
\]
Pinsker's inequality therefore gives
\[
\TotalVariation{P_+}{P_-}\le\frac14.
\]
Using again that \(0\le Z\le b\), the Bayes regret is at least
\begin{align*}
\frac{\epsilon}{2}
\bigl(b-\E_+Z+\E_-Z\bigr)
&\ge
\frac{\epsilon b}{2}
\bigl(1-\TotalVariation{P_+}{P_-}\bigr)\\
&\ge
\frac{3\epsilon b}{8}\\
&\ge
\frac{1}{32\sqrt2}\sqrt b.
\end{align*}

The argument is unchanged after conditioning on any preceding history
independent of the random parameter assigned to the block: the learner
may enter the block in an arbitrary state, but the block parameter retains
the prior above.

Finally, when \(S\) is a single class and the block is bracketed by
baseline rounds, that class changes once when the block begins and once
when it returns to the baseline. Thus the block uses exactly two change
rounds in that class.
\end{proof}



Next, we add uncertainty over the identity of the active
class, as stated in the next lemma. 
The argument is a variant of the standard minimax lower-bound construction
for switching bandits, which partitions the horizon into hard stationary
epochs and randomizes the identity of the improved arm
\citep{auer2002nonstochastic,besbes2014stochastic,auer2019adaptively}.
We use a random permutation of the classes so that the same construction
also respects the class-specific change budgets.

\begin{lemma}[Random-permutation block construction]
\label{lem:random-permutation-blocks}
Consider \(M\ge1\) classes with \(k\ge1\) designated arms in each class,
where \(Mk\ge2\). For every horizon \(h\ge6M^2k\) and every learner,
there exists an instance with at most two change rounds in each class such
that the expected regret over these \(h\) rounds is at least
\[
\frac{1}{512}M\sqrt{kh}.
\]
The same Bayes-regret bound holds conditional on any preceding history that
is independent of the random parameters of the construction.
\end{lemma}

\begin{proof}
Partition the \(h\) rounds into an initial baseline block, \(M\) active
blocks, and a terminal baseline block, each of length
\[
B:=\left\lfloor\frac{h}{M+2}\right\rfloor.
\]
All remaining rounds are also baseline. On baseline rounds, every arm has
mean \(1/2\).

Before interaction, draw a uniformly random permutation of the \(M\)
classes and independently choose one designated arm uniformly from each
class. In active block \(u\), the \(u\)-th class in the permutation is
active. Let
\[
s:=M-u+1
\]
be the number of classes that have not yet appeared, and set
\[
\epsilon_u
:=
\frac{1}{8\sqrt2}\sqrt{\frac{sk}{B}}.
\]
During this block, the selected arm of the active class has mean
\(1/2+\epsilon_u\), while every other arm has mean \(1/2\).

For the analysis, reveal to the learner the active class and improved arm
after every completed active block. This can only make the problem easier.
Conditional on the entire past and these revelations, the next active class
is uniform among the \(s\) unused classes, and its improved arm is uniform
among their \(sk\) designated arms. 
%

Whenever $sk\ge2$, this conditional block law is exactly the
multiple-candidate construction of Lemma C.1 with $sk$ candidate arms.
When $sk=1$, we use only the nonnegativity of regret and do not
invoke the singleton construction of that lemma.

Since \(h\ge6M^2k\),
\[
\frac{h}{M+2}\ge2Mk,
\qquad\text{and hence}\qquad
B\ge Mk\ge sk.
\]
Therefore, whenever \(sk\ge2\), the conditional expected regret in block
\(u\) is at least
\[
\frac{1}{32\sqrt2}\sqrt{skB}.
\]

Suppose first that \(M\ge2\). Keeping only the stages with \(s>M/2\),
\[
\sum_{\substack{s\in[M]\\sk\ge2}}\sqrt{s}
\ge
\frac M2\sqrt{\frac M2}
\ge
\frac{M^{3/2}}4.
\]
Also,
\[
B
=
\left\lfloor\frac{h}{M+2}\right\rfloor
\ge
\frac{h}{2(M+2)}
\ge
\frac{h}{6M}.
\]
Hence the total expected regret is at least
\begin{align*}
\frac{1}{32\sqrt2}\sqrt{kB}
\sum_{\substack{s\in[M]\\sk\ge2}}\sqrt{s}
\ge
\frac{1}{128\sqrt{12}}M\sqrt{kh}
\ge
\frac{1}{512}M\sqrt{kh}.
\end{align*}

If \(M=1\), then \(k\ge2\), and the single active block gives
\[
\frac{1}{32\sqrt2}\sqrt{kB}
\ge
\frac{1}{32\sqrt{12}}\sqrt{kh}
\ge
\frac{1}{512}\sqrt{kh}.
\]

Finally, each class is active in at most one block. Its selected arm moves
from the baseline at the beginning of that block and returns to the
baseline afterward, so every class undergoes at most two change rounds.
Thus every instance in the prior is admissible.

Averaging over the prior yields a fixed admissible instance with at least
the displayed expected regret. The same argument applies conditional on
any preceding history independent of the permutation and improved-arm
identities.
\end{proof}


\subsection{Proofs of Our Main Lower Bounds}
\label{app:lower-bound}

We now apply the two auxiliary constructions to prove the lower bounds stated in
Section~\ref{sec:lower-bounds}.

\subsubsection{Direct-sum construction for arbitrary class parameters}
\label{app:proof-direct-sum-lower-bound}

\begin{theorem}[Restatement of \Cref{thm:direct_sum_lower_bound}]
Fix any collection of class sizes $\{K_m\}_{m=1}^M$ and change budgets $\{L_m\}_{m=1}^M$. If the horizon satisfies $T \ge 2D$, then for any policy, there exists an admissible nonstationary environment such that
\begin{equation*}
    \mathbb{E}[\DynamicRegret] \ge \frac{1}{256} \sqrt{T D}.
\end{equation*}
\end{theorem}

\begin{proof}
Recall that
\(
\TotalDifficulty
=
\sum_{m=1}^{\NumClasses}
(\ChangeBudget{m}+1)\ClassSize{m}.
\)
The construction uses independent learning blocks inside the different
classes. Each such block is surrounded by baseline rounds, so it spends
exactly two changes and presents the learner with a fresh instance of
\Cref{lem:conditional-block-lower-bound}.

We distinguish two cases.

\paragraph{Case 1: \(\TotalDifficulty\le4\NumArms\).}
In this regime a stationary instance is already sufficiently hard.
Apply \Cref{lem:conditional-block-lower-bound} with all \(\NumArms\)
arms as candidates over the full game length $T$. Since
\[
\NumRounds
\ge
2\TotalDifficulty
\ge
2\NumArms,
\]
the lemma applies and gives a stationary instance with
\[
\E[\DynamicRegret]
\ge
\frac{1}{32\sqrt2}\sqrt{\NumArms\NumRounds}.
\]
Every stationary instance is admissible for every collection of change
budgets. Moreover,
\[
\frac{1}{32\sqrt2}\sqrt{\NumArms\NumRounds}
\ge
\frac{1}{64\sqrt2}\sqrt{\NumRounds\TotalDifficulty}
\ge
\frac{1}{256}\sqrt{\NumRounds\TotalDifficulty}.
\]

\paragraph{Case 2: \(\TotalDifficulty>4\NumArms\).}
For each class \(m\), create
\[
q_m
:=
\left\lfloor\frac{\ChangeBudget{m}}2\right\rfloor
\]
learning blocks, each of length
\[
\ClassBlockLength{m}
:=
\left\lfloor
\frac{\NumRounds\ClassSize{m}}{\TotalDifficulty}
\right\rfloor.
\]
Arrange all these blocks in any fixed order, with one baseline round before
the first block, between consecutive blocks, and after the last block.
All arms have mean \(1/2\) on baseline rounds and on any unused rounds.

Since \(\NumRounds\ge2\TotalDifficulty\),
\[
\frac{\NumRounds\ClassSize{m}}{\TotalDifficulty}
\ge
2\ClassSize{m}\ge2.
\]
Hence the floor loses at most a factor of two:
\[
\ClassBlockLength{m}
\ge
\frac{\NumRounds\ClassSize{m}}{2\TotalDifficulty}
\ge
\ClassSize{m}.
\]
Thus the conditional block lower bound applies to every class-\(m\) block.

We next verify that the construction fits within the horizon. The learning
blocks occupy at most
\begin{align*}
\sum_m q_m\ClassBlockLength{m}
&\le
\frac{\NumRounds}{\TotalDifficulty}
\sum_m q_m\ClassSize{m}\\
&\le
\frac{\NumRounds}{2},
\end{align*}
where the last inequality uses
\(2q_m\le\ChangeBudget{m}\).

The number of required baseline separator rounds is at most
\[
1+\sum_mq_m
\le
1+\frac{\TotalDifficulty}{2}
\le
\TotalDifficulty
\le
\frac{\NumRounds}{2}.
\]
Here we used \(\ClassSize{m}\ge1\) and
\(\TotalDifficulty\ge\NumArms\ge2\).
Thus all blocks and separators fit within the game length $T$.

Independently for every class-\(m\) block, sample the mean assignment from
\Cref{lem:conditional-block-lower-bound} with the \(\ClassSize{m}\) arms
of that class as candidates. If \(\ClassSize{m}\ge2\), one uniformly
chosen arm is elevated above the baseline. If \(\ClassSize{m}=1\), its
unique arm receives a positive or negative shift; an outside baseline arm
exists because \(\NumArms\ge2\).

Every class-\(m\) block creates exactly two changes in class \(m\), and
no changes in any other class. Therefore class \(m\) uses at most
\[
2q_m
=
2\left\lfloor\frac{\ChangeBudget{m}}2\right\rfloor
\le
\ChangeBudget{m}
\]
changes, so every instance in the prior is admissible.

It remains to lower-bound the total weight of the blocks. For every
nonnegative integer \(\ChangeBudget{m}\),
\[
\ChangeBudget{m}+1
\le
2+
2\left\lfloor\frac{\ChangeBudget{m}}2\right\rfloor.
\]
Multiplying by \(\ClassSize{m}\) and summing gives
\[
\sum_m q_m\ClassSize{m}
\ge
\frac{\TotalDifficulty}{2}-\NumArms.
\]
Since \(\TotalDifficulty>4\NumArms\),
\[
\sum_m q_m\ClassSize{m}
\ge
\frac{\TotalDifficulty}{4}.
\]

Finally, condition on the complete transcript before each learning block.
Its random mean assignment is independent of that transcript, so
\Cref{lem:conditional-block-lower-bound} applies conditionally. By the
tower property,
\begin{align*}
\E[\DynamicRegret]
&\ge
\frac{1}{32\sqrt2}
\sum_m q_m
\sqrt{\ClassSize{m}\ClassBlockLength{m}}\\
&\ge
\frac{1}{64}
\sqrt{\frac{\NumRounds}{\TotalDifficulty}}
\sum_m q_m\ClassSize{m}\\
&\ge
\frac{1}{256}
\sqrt{\NumRounds\TotalDifficulty}.
\end{align*}
Averaging over the product prior yields a fixed admissible mean sequence
with at least this expected regret.
\end{proof}

We conclude the quantity
\(
\TotalDifficulty
=
\sum_m(\ChangeBudget{m}+1)\ClassSize{m}
\) 
is therefore unavoidable. Each class contributes its own relearning cost,
and observations collected from other classes cannot remove this cost.
Consequently every algorithm suffers
\(
\Omega\!\left(
\sqrt{\NumRounds\TotalDifficulty}
\right)
\)
dynamic regret in general.

The direct-sum construction does not, however, determine whether the
additional dependence on \(\NumClasses\) can be necessary when the number
of classes grows. The next lower bound isolates this cross-class search
difficulty.

\begin{theorem}[Restatement of \Cref{thm:class_search_lower_bound}]
Let $K$ be divisible by $M$, and suppose all classes are balanced with $K_m = K/M$ with change budgets $L_m = 2$ for all $m \in [M]$. If $T \ge 6MK$, then for any policy, there exists an admissible environment such that
\begin{equation*}
    \mathbb{E}[\DynamicRegret] \ge \frac{1}{512} \sqrt{MKT}.
\end{equation*}
\end{theorem}

\begin{proof}
Apply \Cref{lem:random-permutation-blocks} with
\[
n=\NumClasses,
\qquad
k=\frac{\NumArms}{\NumClasses},
\qquad
h=\NumRounds.
\]
The requirement \(nk\ge2\) is exactly the assumption
\(\NumArms\ge2\), and the horizon requirement becomes
\[
\NumRounds
\ge
6n^2k
=
6\NumClasses\NumArms.
\]
The lemma therefore gives
\[
\E[\DynamicRegret]
\ge
\frac{1}{512}\NumClasses
\sqrt{
\frac{\NumArms}{\NumClasses}\NumRounds
}
=
\frac{1}{512}
\sqrt{
\NumClasses\NumArms\NumRounds
}.
\]

Each class is active in at most one block in the construction and therefore
undergoes at most two changes. Hence the instance is admissible when
\(\ChangeBudget{m}=2\) for every \(m\in[\NumClasses]\).
\end{proof}

Note that for this balanced family,
\[
\ClassSize{m}=\frac{\NumArms}{\NumClasses},
\qquad
\ChangeBudget{m}=2,
\]
and therefore
\[
\TotalDifficulty
=
\sum_{m=1}^{\NumClasses}
3\frac{\NumArms}{\NumClasses}
=
3\NumArms.
\]
Thus the direct-sum difficulty is only of order \(\NumArms\), yet every
algorithm must incur
\[
\Omega\!\left(
\sqrt{
\NumClasses\NumArms\NumRounds
}
\right)
=
\Omega\!\left(
\sqrt{
\NumClasses\NumRounds\TotalDifficulty
}
\right)
\]
dynamic regret.

Hence the additional factor \(\sqrt{\NumClasses}\) therefore cannot in general
be removed. Even with only two changes per class, the learner must
repeatedly determine which class contains the improved arm. The
\(\sqrt{\NumClasses}\) dependence in our upper bound is thus an
unavoidable class-search cost on this balanced family.

The two lower bounds capture different difficulties: the direct-sum bound
measures the cumulative cost of relearning arms within classes, while the
class-search bound measures the additional cost of locating the relevant
class. The following corollary records both statements on the balanced
two-change family.

\begin{corollary}[Combined lower bound]
\label{cor:combined-lower-bound}
Suppose \(\NumArms\ge2\) and \(\NumArms\) is divisible by
\(\NumClasses\). If
\[
\ClassSize{m}=\frac{\NumArms}{\NumClasses},
\qquad
\ChangeBudget{m}=2
\]
for every \(m\in[\NumClasses]\), and
\(\NumRounds\ge6\NumClasses\NumArms\), then every algorithm has an
admissible instance on which
\[
\E[\DynamicRegret]
\ge
\max\!\left\{
\frac{1}{512}
\sqrt{\NumClasses\NumArms\NumRounds},
\frac{1}{256}
\sqrt{\NumRounds\TotalDifficulty}
\right\}.
\]
\end{corollary}

\begin{proof}
For these parameters,
\[
\TotalDifficulty=3\NumArms.
\]
Moreover,
\[
\NumRounds
\ge
6\NumClasses\NumArms
\ge
6\NumArms
=
2\TotalDifficulty.
\]
Hence both
\Cref{thm:class_search_lower_bound,thm:direct_sum_lower_bound}
apply to the same class sizes and budgets.

Fix an algorithm. Each theorem supplies an admissible hard instance,
possibly a different one. Choose the instance supplied by whichever
theorem gives the larger of the two displayed lower bounds.
\end{proof}

\section{Extension to Unknown Budgets}
\label{app:adaptive-extension}

In this section, we remove the assumption that the class change budgets
\(\ChangeBudget{1},\ldots,\ChangeBudget{\NumClasses}\) are known.
We introduce Adaptive Exponential Backfill for Heterogeneous Arm Classes
(\texttt{AEBF-H}), which receives only the horizon \(\NumRounds\) and the
class partition and achieves the same leading regret rate as
\texttt{EBF-H}.

The main challenge is that the backfill probability of \texttt{EBF-H}
depends explicitly on \(\ChangeBudget{m}\). A natural alternative is to
replace this budget by the number of changes detected so far in class
\(m\). Doing so introduces a new feedback loop: monitoring affects the
observations, these observations affect classification and resets, and
these in turn determine future monitoring rates. \texttt{AEBF-H} breaks
this feedback by separating classification observations from additional
monitoring observations and by servicing monitoring requests through an
independent monitoring branch.

Apart from these adaptations, the algorithm preserves the phase-based
classification mechanism of \texttt{EBF-H}: it uses the same
\texttt{ClassifyArms} procedure, blocking arms, dominance gaps and scales,
and the same class--scale priority rule from \Cref{sec:algorithm}.

\subsection{The \texttt{AEBF-H} Algorithm}
\label{app:adaptive-algorithm}

We retain all notation from \Cref{sec:algorithm}. In particular,
\[
\MaxScale:=\lceil\log_2\NumRounds\rceil,
\qquad
\LogFactor:=\lceil\log(2\NumRounds\NumArms)\rceil.
\]

\paragraph{Classification and detection statistics.}
Unlike \texttt{EBF-H}, \texttt{AEBF-H} separates the observations used
for \emph{arm classification} from those used for \emph{change detection}.
For every arm \(a\), it maintains two histories:
$
\HistoryObservationRounds{a}{\ClassificationTag}
\subseteq
\HistoryObservationRounds{a}{\DetectionTag}.
$

The \emph{classification history}
\(\HistoryObservationRounds{a}{\ClassificationTag}\) contains only rewards
collected on the classification branch. These observations are used by
\texttt{ClassifyArms} to form certificates, classify arms as good or bad,
compute dominance gaps and scales, and determine the class--scale priority
rule. Thus, with respect to this history, the classification mechanism is
exactly the one from \Cref{sec:algorithm}.

The \emph{detection history}
\(\HistoryObservationRounds{a}{\DetectionTag}\) contains all classification
observations together with the additional observations collected while
servicing monitoring obligations. It is used only for detecting distribution
changes and deciding when a class should be reset.

This separation is necessary in the unknown-budget setting. Monitoring
obligations are launched randomly at rates that depend on the previously
detected changes. If the resulting monitoring observations were also used
for classification, the random launch decisions could change future
certificates and hence determine which obligations remain relevant. This
would create a feedback loop between launching a monitoring request and
the future classification trajectory. On the other hand, ignoring the
monitoring observations altogether would defeat their purpose: they are
precisely the additional samples used to reveal that a previously bad arm
has changed. Using two histories resolves both issues: monitoring
observations contribute to change detection without directly affecting
classification.

For \(H\in\{\ClassificationTag,\DetectionTag\}\) and an interval \(I\),
let

$$
\TaggedNumPulls{a}{H}{I},\qquad
\TaggedEmpMean{a}{H}{I},\qquad
\TaggedConfRadius{a}{H}{I},\qquad
\TaggedEmpBound{a}{H}{\pm q}{I}
$$

denote the pull count, empirical mean, confidence radius, and empirical
confidence bounds of \Cref{sec:algorithm}, computed using only observations
from history \(H\). Throughout this appendix, unmarked statistics refer to
the classification history, while detection statistics retain the
\(\DetectionTag\) superscript. In particular, \texttt{ClassifyArms} from
\Cref{alg:classify} can be applied without modification.

\paragraph{Mechanism 1: Adaptive class-isolated change detection.}
As in \texttt{EBF-H}, every class has its own phase start
\(\PhaseStart{m}\), and resets are performed independently across classes.
The difference is that change detection may now use either observation
history.

Let \(\DetectedChangeCount{m}{t}\) be the number of resets of class \(m\)
completed through the reset step at the beginning of round \(t\), with
\(\DetectedChangeCount{m}{0}=0\). At round \(t\), class \(m\) is reset if
there exist \(a\in\ClassArms{m}\),
\(H,H'\in\{\ClassificationTag,\DetectionTag\}\), and intervals
\(I_1,I_2\subseteq[\PhaseStart{m},t-1]\) such that
\begin{equation}
\label{eq:adaptive-reset-test}
\TaggedEmpBound{a}{H}{-1}{I_1}
>
\TaggedEmpBound{a}{H'}{+1}{I_2}.
\end{equation}
This is exactly the change-detection test of \Cref{sec:algorithm}, except
that its two intervals may use either history.

The intuition behind the design is that the two sides of the reset test need not use the same observation history.
Indeed, suppose arm \(a\) has not changed during the current phase and has
constant mean \(\mu\). On the concentration event, for either
\(H\in\{\ClassificationTag,\DetectionTag\}\) and every stationary interval
\(I\),
\[
\TaggedEmpBound{a}{H}{-1}{I}
\le
\mu
\le
\TaggedEmpBound{a}{H}{+1}{I}.
\]
Consequently, for any two histories \(H,H'\) and any two phase intervals
\(I_1,I_2\),
\[
\TaggedEmpBound{a}{H}{-1}{I_1}
\le
\mu
\le
\TaggedEmpBound{a}{H'}{+1}{I_2},
\]
so the strict inequality in the reset test cannot hold without a true
change.

Allowing cross-history comparisons is essential because the two histories
play different roles. Classification observations determine certificates
and therefore the future classification trajectory, whereas additional
monitoring observations are recorded only in the detection history. Thus,
after a bad arm changes, fresh monitoring observations can eventually
become inconsistent with its stale classification certificate, e.g.,
\[
\TaggedEmpBound{a}{\DetectionTag}{-1}{I_{\mathrm{new}}}
>
\TaggedEmpBound{a}{\ClassificationTag}{+1}{I_{\mathrm{old}}},
\]
triggering a reset without allowing monitoring samples to directly affect
classification. The four choices
\((\ClassificationTag,\ClassificationTag)\),
\((\ClassificationTag,\DetectionTag)\),
\((\DetectionTag,\ClassificationTag)\), and
\((\DetectionTag,\DetectionTag)\)
therefore preserve the original detector while allowing fresh monitoring
data to detect changes in either direction.

When a reset occurs, we set
\[
\PhaseStart{m}\leftarrow t,
\qquad
\DetectedChangeCount{m}{t}
\leftarrow
\DetectedChangeCount{m}{t}+1,
\]
and discard all outstanding monitoring requests for arms in
\(\ClassArms{m}\). Observations need not be deleted, since all subsequent
statistics are restricted to the new phase.

As in the known-budget algorithm, the reset rule is conservative:
\Cref{lem:adaptive-reset-counting} shows that, under simultaneous
concentration, every reset can be charged to a true change. In particular,
\[
\DetectedChangeCount{m}{t}\le\ChangeBudget{m}.
\]
Thus, the detected reset count can therefore serve as an online proxy for the
unknown budget.

\paragraph{Mechanism 2: Arm classification and dominance gaps.}
Classification is unchanged from \texttt{EBF-H}. After the reset step, we
run
\(
\texttt{ClassifyArms}
(t,\PhaseStart{1},\ldots,\PhaseStart{\NumClasses})
\) 
from \Cref{alg:classify}, using the classification observations only.
This produces exactly the same objects as in \Cref{sec:algorithm}:
\(\GoodArms{t}\), \(\BadArms{t}\), blocking arms, certificate intervals,
dominance gaps \(\DominanceGap{t}{a}\), and dominance scales
\(\DominanceScale{t}{a}\).
The structural property that blocking arms are good also continues to hold.

\paragraph{Mechanism 3: Budget-free adaptive backfill.}
The main difference from \texttt{EBF-H} is the monitoring schedule.
At scale \(j\), a monitoring request requires
\(
\ServiceTarget{j}:=2^{2j+7}\LogFactor
\)
observations. Instead of using the unknown budget
\(\ChangeBudget{m}\), class \(m\) launches monitoring requests at rate
\begin{equation}
\label{eq:adaptive-launch-rate}
\AdaptiveLaunchProbability{m}{j}{t}
:=
\min\!\left\{
1,\;
2^{-j}
\sqrt{
\frac{\DetectedChangeCount{m}{t}+1}
{\NumRounds\ClassSize{m}\LogFactor}
}
\right\}.
\end{equation}

This rule has the same qualitative behavior as the backfill rule of
\texttt{EBF-H}. Finer scales require more samples and are therefore
launched less frequently, while classes in which more changes have already
been detected are monitored more aggressively. The crucial difference is
that the rate is now completely data-driven.

A successful launch creates a \emph{monitoring obligation}
\(
(a,j,s,r),
\) 
where \(a\) is the arm, \(j\) is the scale, \(s\) is the launch round,
and \(r\) is the number of required monitoring observations remaining.
A scale-\(j\) launch at round \(s\) creates
\(
(a,j,s,\ServiceTarget{j}).
\)
Let \(\ObligationSet\) denote the set of live obligations.

Unlike the single counters of \texttt{EBF-H}, different launches are kept
separately rather than overwriting one another. This is important in the
adaptive analysis because each launch has its own independent random seed
and can be charged separately.


For every bad arm $a\in\mathcal A_m\cap\mathcal B_t$ and scale
$j\in\{0,\ldots,\min\{i_t(a),J_{\max}\}\}$, an obligation is launched
with probability $q_{m,j}(t)$. All launch decisions use mutually
independent uniform random seeds, indexed by arm, scale, and round.
Conditional on the history before round $t$, the launch decisions
at that round are independent with their stated history-dependent
probabilities.

\paragraph{Mechanism 4: Independent monitoring.}
Once a monitoring obligation is created, the algorithm must collect the
additional samples it requests. The difficulty is that these samples
cannot be scheduled using the obligation set itself: whether an obligation
exists depends on the random coin used to launch it, so choosing future
actions based directly on the outstanding obligations would make those
launch coins influence the subsequent sampling process and, potentially,
the classification trajectory.

To avoid this dependence, \texttt{AEBF-H} separates ordinary
classification from monitoring. At every round, it independently chooses
between a \emph{classification branch}, which continues the usual
\texttt{EBF-H} classification process, and a \emph{monitoring branch},
which provides opportunities to service outstanding obligations. On the
monitoring branch, the candidate arm is sampled independently of which
obligations are currently active. Thus an obligation determines only
whether a monitoring opportunity is used, not which arm receives that
opportunity.

An obligation \((a,j,s,r)\) is \emph{serviceable} at round \(t\) if
either \(a\) is currently good, or \(a\) is bad and its scale is still
relevant:
\[
\ServiceableObligationSet{a}{t}
:=
\left\{
(a,j,s,r)\in\ObligationSet:
a\in\GoodArms{t}
\ \text{or}\
\bigl(
a\in\BadArms{t},
j\le\DominanceScale{t}{a}
\bigr)
\right\}.
\]
An obligation that is temporarily not serviceable remains live.

At every round, the algorithm first constructs a
\emph{classification proposal}. It draws a class uniformly among those
containing a good arm and a scale uniformly from
\(\{0,\ldots,\MaxScale\}\), and then applies the same recency rule as
\texttt{EBF-H}, restricted to good arms and classification observations.

The algorithm then chooses uniformly between two branches:

\begin{enumerate}
\item \emph{Classification branch.}
Play the classification proposal and add its reward to both the
classification and detection histories.

\item \emph{Monitoring branch.}
Draw a class uniformly from \([\NumClasses]\), and then draw an arm
uniformly from that class, independently of the outstanding obligations.
If the selected arm has a serviceable obligation, play it, add its reward
only to the detection history, and decrease the remaining requirement of
every serviceable obligation of that arm by one. Otherwise, play the
classification proposal but do not add this reward to either history.
\end{enumerate}

The last convention is deliberate. Whether the monitoring candidate has a
serviceable obligation depends on previous launch coins. If a fallback
reward were added to the classification history, these launch coins could
change the future classification trajectory. Discarding the fallback
observation prevents this feedback.

Consequently, every arm \(a\in\ClassArms{m}\) receives an independent
monitoring opportunity with probability
\[
\MonitoringSlotProbability{m}
=
\frac{1}{2\NumClasses\ClassSize{m}}
\]
at every round. This predictable service rate is the key property used in
the adaptive analysis.

\begin{algorithm2e}[t!]
\caption{\texttt{AEBF-H}: Adaptive Exponential Backfill for Heterogeneous Arm Classes
\label{alg:adaptive-ebf-h}}
\SetAlgoLined
\DontPrintSemicolon
\KwIn{Horizon \(\NumRounds\) and classes
\(\ClassArms{1},\ldots,\ClassArms{\NumClasses}\); no change budgets.}

Set
\(\MaxScale\leftarrow\lceil\log_2\NumRounds\rceil\),
\(\LogFactor\leftarrow\lceil\log(2\NumRounds\NumArms)\rceil\), and
\(\ServiceTarget{j}\leftarrow2^{2j+7}\LogFactor\) for
\(j\in\{0,\ldots,\MaxScale\}\)\;

Initialize \(\PhaseStart{m}\leftarrow1\) and
\(\DetectedChangeCount{m}{0}\leftarrow0\) for every class \(m\);
initialize both observation histories and
\(\ObligationSet\) to \(\emptyset\)\;

\For{\(t=1,\ldots,\NumRounds\)}{

  \tcp{1. Adaptive class-isolated change detection}
  \ForEach{\(m\in[\NumClasses]\)}{
    \(\DetectedChangeCount{m}{t}
    \leftarrow\DetectedChangeCount{m}{t-1}\)\;

    \If{\(\exists a\in\ClassArms{m}\),
    \(H,H'\in\{\ClassificationTag,\DetectionTag\}\),
    \(I_1,I_2\subseteq[\PhaseStart{m},t-1]\) such that
    \[
    \TaggedEmpBound{a}{H}{-1}{I_1}
    >
    \TaggedEmpBound{a}{H'}{+1}{I_2}
    \]}{
      \(\PhaseStart{m}\leftarrow t\)\;
      \(\DetectedChangeCount{m}{t}
      \leftarrow\DetectedChangeCount{m}{t}+1\)\;
      Delete all obligations whose arm belongs to \(\ClassArms{m}\)\;
    }
  }

  \tcp{2. Arm classification}
  Run
  \(\texttt{ClassifyArms}
  (t,\PhaseStart{1},\ldots,\PhaseStart{\NumClasses})\)
  from \Cref{alg:classify} using classification observations only\;

  \tcp{3. Adaptive backfill launches}
  \ForEach{\(a\in\BadArms{t}\)}{
    \(m\leftarrow\ArmClass{a}\)\;
    \ForEach{\(j=0,\ldots,
    \min\{\DominanceScale{t}{a},\MaxScale\}\)}{
      With probability
      \(\AdaptiveLaunchProbability{m}{j}{t}\),
      add \((a,j,t,\ServiceTarget{j})\) to
      \(\ObligationSet\)\;
    }
  }

  \tcp{4. Classification proposal}
  Draw \(\SelectedClass{t}\) uniformly from
  \(\{m:\ClassArms{m}\cap\GoodArms{t}\neq\emptyset\}\)
  and \(\SelectedScale{t}\) uniformly from
  \(\{0,\ldots,\MaxScale\}\)\;

  Set
  \[
  \ClassificationProposal{t}
  \leftarrow
  \argmin_{
  a\in\ClassArms{\SelectedClass{t}}\cap\GoodArms{t}}
  \PriorityTime{a}{\SelectedScale{t}},
  \]
  using classification observations in the current phase\;

  \tcp{5. Classification or monitoring}
  Draw
  \(\BranchChoice{t}\) uniformly from
  \(\{\ClassificationBranch,\MonitoringBranch\}\)\;

  \eIf{\(\BranchChoice{t}=\ClassificationBranch\)}{
    Play
    \(\PlayedArm{t}\leftarrow\ClassificationProposal{t}\)
    and observe \(\ObservedReward{\PlayedArm{t}}{t}\)\;
    Add \(t\) to both histories of \(\PlayedArm{t}\)\;
  }{
    Draw \(\MonitorClass{t}\) uniformly from \([\NumClasses]\),
    and then
    \(\MonitorCandidate{t}\) uniformly from
    \(\ClassArms{\MonitorClass{t}}\)\;

    \eIf{\(
    \ServiceableObligationSet{\MonitorCandidate{t}}{t}
    \neq\emptyset
    \)}{
      Play
      \(\PlayedArm{t}\leftarrow\MonitorCandidate{t}\)
      and observe \(\ObservedReward{\PlayedArm{t}}{t}\)\;
      Add \(t\) only to the detection history of \(\PlayedArm{t}\)\;

      \ForEach{serviceable obligation
      \((\PlayedArm{t},j,s,r)\)}{
        Decrease \(r\) by one and delete the obligation if \(r=0\)\;
      }
    }{
      Play
      \(\PlayedArm{t}\leftarrow\ClassificationProposal{t}\)
      and observe its reward, but do not add \(t\) to either history\;
    }
  }
}
\end{algorithm2e}

\paragraph{Run-time complexity.}
As for \texttt{EBF-H}, the exact implementation is dominated by the
all-interval classification step. At round \(t\),
\texttt{ClassifyArms} requires
\(\Oof{\NumArms^2t^4}\) time.
The two-history change-detection test considers only four ordered history
pairs and requires \(\Oof{\NumArms t^4}\) time, which is lower order.

At most one obligation is launched for each arm, scale, and round, so
there are at most
\(
\Oof{\NumArms\NumRounds(1+\log\NumRounds)}
\)
live obligations. Storing them in per-arm lists makes their total scanning
cost
\(
\Oof{\NumArms\NumRounds^2(1+\log\NumRounds)},
\)
which is again lower order. Hence the total running time is
\(
\Oof{\NumArms^2\NumRounds^5}.
\)
Using prefix statistics for both observation histories, the total memory
complexity is
\(
\Oof{\NumArms\NumRounds(1+\log\NumRounds)},
\) hence the algorithm is efficient.

The next theorem states the expected dynamic regret of \cref{alg:adaptive-ebf-h}.
\begin{theorem}[Expected Dynamic Regret for Unknown Class Change Budgets]
\label{thm:adaptive-main}
For every admissible nonstationary environment, the expected dynamic
regret of \texttt{AEBF-H} satisfies
\[
\E[\DynamicRegret]
\le
16384\LogFactor^2
\sqrt{\NumClasses\NumRounds\TotalDifficulty}+3
=
\BigOTilde\!\left(
\sqrt{\NumClasses\NumRounds\TotalDifficulty}
\right).
\]
\end{theorem}

We emphasize that the class change budgets are not supplied to the algorithm and appear only through \(\TotalDifficulty\) in the guarantee. 

Thus prior knowledge of the class change budgets can be removed without
changing the leading regret rate of \texttt{EBF-H}. The proof of
\Cref{thm:adaptive-main} is given next.

\subsection{Regret Analysis}
\label{app:adaptive-analysis}
Our goal is to prove \Cref{thm:adaptive-main}.
The proof follows the same structure as the known-budget analysis in
\Cref{sec:analysis} and Appendix~\ref{app:known-budget-statements}.
We first establish simultaneous concentration and show that every
algorithmic reset can be charged to a true change. This controls the number
of maximal static intervals (MSIs) and, crucially, implies that the number
of detected changes never exceeds the unknown budget.

We then prove the confidence-sum bounds needed in the regret
decomposition. As in the known-budget analysis, every bad arm has a good
blocking arm, so both the played and optimal arms can be represented by
good arms.

Finally, we analyze the three terms in
\eqref{eq:regret-decomposition} separately. Terms (i) and (ii) are close
to their known-budget counterparts. The main new argument is for
Term (iii), delayed change detection. Between two detected resets of a
class, the adaptive launch probability is fixed. We use this fact to bound
the waiting time until a useful monitoring obligation is launched and then
show that sufficiently servicing such an obligation forces a reset. This
replaces the fixed-budget waiting-time argument used for \texttt{EBF-H}.

Throughout the analysis,
\(\PreRoundHistory{t}\) denotes the information revealed before the fresh
random draws at round \(t\).

\subsubsection{Concentration, Reset Counting, and MSIs}
\label{app:proof-adaptive-concentration-counts}

We first extend the concentration and reset-counting arguments of the
known-budget analysis to the two histories maintained by
\texttt{AEBF-H}.

\begin{lemma}[Two-history concentration]
\label{lem:adaptive-concentration}
Let \(\ConcentrationEvent'\) be the event that the following holds
simultaneously for both
\(H\in\{\ClassificationTag,\DetectionTag\}\), every arm \(a\), and every
nonempty interval \(I=[s,v]\) contained in one stationary segment of that
arm:
\[
\left|
\TaggedEmpMean{a}{H}{I}
-
\MeanReward{a}{s}
\right|
\le
\TaggedConfRadius{a}{H}{I}.
\]
Then
\[
\Prob(\ConcentrationEvent')\ge1-\NumRounds^{-1}.
\]
\end{lemma}

\paragraph{Proof intuition.}
The observation times in both histories are adaptive, but whether a reward
is recorded is determined before that reward is revealed. The recorded
centered rewards therefore form a bounded martingale-difference sequence,
so concentration remains valid at these adaptive observation times.

\begin{proof}
Fix \(a\in\ArmSet\),
\(H\in\{\ClassificationTag,\DetectionTag\}\), and
\(s\in[\NumRounds]\).
For \(u\in[\NumRounds]\), the event
\(
u\in\HistoryObservationRounds{a}{H}
\)
means that round \(u\) records a reward of arm \(a\) in history \(H\).

Let \(\PreRewardHistory{u}\) additionally contain all internal random
draws at round \(u\) that determine the played arm and whether its reward
is recorded, but not the reward itself. Then
\[
\PreRoundHistory{u}
\subseteq
\PreRewardHistory{u}
\subseteq
\PreRoundHistory{u+1},
\]
and
\(
\Indicator{u\in\HistoryObservationRounds{a}{H}}
\)
is \(\PreRewardHistory{u}\)-measurable.

Because the reward is sampled only after these decisions, on a stationary
segment of arm \(a\) with mean \(\mu\),
\[
\E\!\left[
\Indicator{u\in\HistoryObservationRounds{a}{H}}
\bigl(\ObservedReward{a}{u}-\mu\bigr)
\,\middle|\,
\PreRewardHistory{u}
\right]
=0.
\]
Thus the recorded centered rewards form a bounded martingale-difference
sequence.

Let \(v\) be the last round in the stationary segment of arm \(a\)
containing \(s\). For \(n\in[\NumRounds]\), let \(u_n\) be the round of
the \(n\)-th reward recorded in history \(H\) during \([s,v]\), with
\(u_n=\infty\) if there are fewer than \(n\) such rewards. Let
\(
\mu :=\MeanReward{a}{s}.
\)
For every fixed \(\lambda\in\mathbb R\), the conditional form of
Hoeffding's lemma \citep[Section~5.4, Note~4]{lattimore2020bandit} gives
the nonnegative supermartingale
\[
\exp\!\left(
\lambda
\sum_{u=s}^{t}
\Indicator{u\in\HistoryObservationRounds{a}{H}}
(\ObservedReward{a}{u}-\mu)
-
\frac{\lambda^2}{8}
\sum_{u=s}^{t}
\Indicator{u\in\HistoryObservationRounds{a}{H}}
\right),
\qquad
s\le t\le v.
\]
Stop it at \(\min\{u_n,v\}\), using bounded optional stopping
\citep[Theorem~3.8]{lattimore2020bandit}. Optimizing \(\lambda\) for each
of the two tails gives
\[
\Prob\!\left(
u_n\le v,\quad
\left|
\TaggedEmpMean{a}{H}{[s,u_n]}-\mu
\right|
>
\sqrt{
\frac{8\log(2\NumRounds\NumArms)}{n}
}
\right)
\le
2\exp\!\bigl(
-16\log(2\NumRounds\NumArms)
\bigr).
\]

The empirical statistic in this event is evaluated only when \(u_n\) is
finite. In particular, the bound is on the joint event of observing \(n\)
rewards and violating concentration; it does not condition on the random
observation count.

Now take any interval \(I=[s,t]\) contained in that stationary segment.
If \(I\) contains \(n\ge1\) rewards of arm \(a\) recorded in history
\(H\), and \(u_n\le t\) is the round of the last such reward, then
\[
\TaggedNumPulls{a}{H}{I}
=
\TaggedNumPulls{a}{H}{[s,u_n]}
=
n,
\]
and
\[
\TaggedEmpMean{a}{H}{I}
=
\TaggedEmpMean{a}{H}{[s,u_n]}.
\]
The two intervals also have the same constant mean. Hence every
nonzero-count interval statistic is determined by its arm, history,
starting round, and final tagged-observation count.

A union bound over at most
\(2\NumArms\NumRounds^2\)
such choices has failure probability at most
\[
4\NumArms\NumRounds^2
\exp\!\bigl(
-16\log(2\NumRounds\NumArms)
\bigr)
\le
\NumRounds^{-1}.
\]
When an interval contains no reward recorded in history \(H\), its
empirical mean is zero, its true mean lies in \([0,1]\), and its confidence
radius is at least one. This proves simultaneous concentration.
\end{proof}

We next show that concentration prevents spurious resets.

\begin{lemma}[Reset counting]
\label{lem:adaptive-reset-counting}
On \(\ConcentrationEvent'\), if
\((a,H,H',I_1,I_2)\)
witnesses a class-\(m\) reset at round \(t\) through
\eqref{eq:adaptive-reset-test}, then arm \(a\) has a mean change strictly
after the phase start immediately preceding that reset and before \(t\).
Consequently, for every class \(m\) and round \(t\),
\[
\DetectedChangeCount{m}{t}
\le
\bigl|\{s\in\ChangeRounds{m}:s<t\}\bigr|
\le
|\ChangeRounds{m}|
\le
\ChangeBudget{m}.
\]
\end{lemma}

\paragraph{Proof intuition.}
Once both histories concentrate, a reset cannot occur unless a genuine
change has occurred since the beginning of the current phase: otherwise
all confidence intervals for the same arm contain the same constant mean.
Moreover, after a reset the phase start moves forward, so the same true
change cannot be used to justify a later reset.

\begin{proof}
Fix \(m\in[\NumClasses]\) and consider a reset of class \(m\)
at round \(t\). Let \(\PhaseStart{m}\) denote its previous phase start.
By the reset rule, there are
\(a\in\ClassArms{m}\),
\(H,H'\in\{\ClassificationTag,\DetectionTag\}\), and intervals
\[
I_1,I_2
\subseteq
[\PhaseStart{m},t-1]
\]
satisfying \eqref{eq:adaptive-reset-test}.

Suppose, toward a contradiction, that arm \(a\) has no mean change in
rounds
\[
\PhaseStart{m}+1,\ldots,t-1.
\]
Its mean is then
\(\MeanReward{a}{\PhaseStart{m}}\)
throughout both witness intervals.

The two witness intervals each contain at least one observation in their
respective histories. With zero observations, the tagged lower bound is
negative and the tagged upper bound exceeds one, so a zero-observation
interval cannot occur on either side of the strict reset inequality.

On \(\ConcentrationEvent'\),
\[
\TaggedEmpBound{a}{H}{-1}{I_1}
\le
\MeanReward{a}{\PhaseStart{m}}
\le
\TaggedEmpBound{a}{H'}{+1}{I_2},
\]
contradicting \eqref{eq:adaptive-reset-test}. Hence every reset can be
charged to a class change strictly after the preceding phase start.

After the reset, the new phase starts at \(t\), so a later reset must be
charged to a later change. Therefore, for every round \(u\),
\[
\DetectedChangeCount{m}{u}
\le
\bigl|\{v\in\ChangeRounds{m}:v<u\}\bigr|
\le
|\ChangeRounds{m}|
\le
\ChangeBudget{m}.
\]
\end{proof}

The reset-counting bound immediately controls the number of MSIs.

\begin{lemma}[Number of MSIs]
\label{lem:adaptive-msi-count}
On \(\ConcentrationEvent'\), the number of class-\(m\) MSIs is at most
\[
|\ChangeRounds{m}|
+
\DetectedChangeCount{m}{\NumRounds}
+1
\le
2\ChangeBudget{m}+1.
\]
\end{lemma}

\begin{proof}
The first class-\(m\) MSI starts at round \(1\), and every subsequent MSI
starts at either a true class-\(m\) change or an algorithmic reset of
class \(m\). Hence the number of class-\(m\) MSIs is at most
\[
|\ChangeRounds{m}|
+
\DetectedChangeCount{m}{\NumRounds}
+1.
\]
By \Cref{lem:adaptive-reset-counting},
\[
\DetectedChangeCount{m}{\NumRounds}
\le
\ChangeBudget{m},
\]
while by definition
\[
|\ChangeRounds{m}|
\le
\ChangeBudget{m}.
\]
Therefore
\[
|\ChangeRounds{m}|
+
\DetectedChangeCount{m}{\NumRounds}
+1
\le
2\ChangeBudget{m}+1.
\]
\end{proof}

\paragraph{The stopped process.}
To avoid conditioning fresh random draws on the future event
\(\ConcentrationEvent'\), let \(\AdaptiveStopTime\) be the first round in
\([\NumRounds]\) at whose start one of the concentration inequalities in
\Cref{lem:adaptive-concentration} fails for either history on an
arm-stationary interval of preceding rounds. Set
\(
\AdaptiveStopTime=\NumRounds+1
\)
if no such failure occurs, and define
\(
\AdaptiveStoppedRounds
:=
\{t\in[\NumRounds]:t<\AdaptiveStopTime\}.
\)
We call the execution on \(\AdaptiveStoppedRounds\) the
\emph{stopped process}. This stopping rule is used only in the analysis.

For every \(t\in\AdaptiveStoppedRounds\), all classification- and
detection-history statistics formed from observations available at the
start of round \(t\) satisfy the concentration inequalities of
\Cref{lem:adaptive-concentration}. Moreover, whether
\(t\in\AdaptiveStoppedRounds\) is determined before the fresh random draws
and reward at round \(t\). Consequently, all algorithmic random variables
retain their stated conditional distributions on the stopped process.

Applying the reset-counting argument of
\Cref{lem:adaptive-reset-counting} to the concentrated observations
available before each stopped round gives
\[
\DetectedChangeCount{m}{t}
\le
\ChangeBudget{m},
\qquad
m\in[\NumClasses],\quad
t\in\AdaptiveStoppedRounds.
\]

Finally, the omitted suffix
\([\NumRounds]\setminus\AdaptiveStoppedRounds\) is nonempty only if
\(\ConcentrationEvent'\) fails. By
\Cref{lem:adaptive-concentration},
\[
\Prob\bigl((\ConcentrationEvent')^c\bigr)
\le
\NumRounds^{-1}.
\]
Since the regret over the omitted suffix is at most \(\NumRounds\), its
expected contribution is at most one.

\paragraph{Convention for the remainder of the proof.}
From this point onward, all regret bounds are proved for the stopped
process. Thus, whenever \(t\in\AdaptiveStoppedRounds\), every
classification- or detection-history statistic formed from observations
available at the start of round \(t\) satisfies the concentration bounds
of \Cref{lem:adaptive-concentration}. We therefore invoke
concentration freely on stopped rounds without conditioning on the global
event \(\ConcentrationEvent'\).
This distinction is useful because membership in
\(\AdaptiveStoppedRounds\) is determined before the fresh random draws at
round \(t\). Hence the launch, branch, class, and arm-selection coins retain
their stated conditional distributions throughout the stopped-process
analysis.

\subsubsection{Blocking Arms Are Good}
\label{app:proof-adaptive-blocking-is-good}

The classification mechanism is unchanged from \texttt{EBF-H}, except
that it uses only the classification history. The blocking-arm property
therefore continues to hold.

\begin{proposition}[Adaptive blocking arms are good]
\label{prop:adaptive-blocking-is-good}
For every round \(t\) and arm \(a\in\BadArms{t}\),
\(\BlockingArm{t}{a}\in\GoodArms{t}\).
Consequently, \(\GoodArms{t}\neq\emptyset\).
\end{proposition}

\begin{proof}
Fix a round \(t\) and an arm \(a\in\BadArms{t}\), and write
\(b:=\BlockingArm{t}{a}\).
Suppose, toward a contradiction, that \(b\in\BadArms{t}\).
Let the selected certificate for \(a\) be 
\(
\left(
b,\,
\ArmCertificateInterval{t}{a},\,
\BlockingCertificateInterval{t}{a}
\right),
\)
where \(\ArmCertificateInterval{t}{a}\) is the observation interval for
\(a\), and \(\BlockingCertificateInterval{t}{a}\) is the observation
interval for \(b\). Hence
\begin{equation}
\label{eq:adaptive-blocking-cert-a}
\EmpBound{b}{-3}{\BlockingCertificateInterval{t}{a}}
>
\EmpBound{a}{+3}{\ArmCertificateInterval{t}{a}}.
\end{equation}

Since \(b\) is also bad, let \(c:=\BlockingArm{t}{b}\).
Let its selected certificate be
\(
\left(
c,\,
\ArmCertificateInterval{t}{b},\,
\BlockingCertificateInterval{t}{b}
\right),
\) 
and therefore
\begin{equation}
\label{eq:adaptive-blocking-cert-b}
\EmpBound{c}{-3}{\BlockingCertificateInterval{t}{b}}
>
\EmpBound{b}{+3}{\ArmCertificateInterval{t}{b}}.
\end{equation}

\paragraph{Step 1: Construct a second certificate for \(a\).}
We claim that
\(
\left(
c,\,
\ArmCertificateInterval{t}{a},\,
\BlockingCertificateInterval{t}{b}
\right)
\) 
is also a valid certificate for \(a\).

Because \(b\) is bad after the reset step at round \(t\), its class was
not reset at that round. Indeed, immediately after a reset the new phase
contains no classification observations, so no three-confidence
certificate for \(b\) can exist.

Thus the change-detection test did not fire for the class of \(b\).
Applying its
\((\ClassificationTag,\ClassificationTag)\) specialization to the two
classification intervals of \(b\),
\(\BlockingCertificateInterval{t}{a}\) and
\(\ArmCertificateInterval{t}{b}\), gives
\begin{equation}
\label{eq:adaptive-blocking-no-reset}
\EmpBound{b}{-1}{\BlockingCertificateInterval{t}{a}}
\le
\EmpBound{b}{+1}{\ArmCertificateInterval{t}{b}}.
\end{equation}
Equivalently,
\[
\EmpMean{b}{\ArmCertificateInterval{t}{b}}
\ge
\EmpMean{b}{\BlockingCertificateInterval{t}{a}}
-
\ConfRadius{b}{\BlockingCertificateInterval{t}{a}}
-
\ConfRadius{b}{\ArmCertificateInterval{t}{b}}.
\]

Combining this inequality with
\eqref{eq:adaptive-blocking-cert-a} and
\eqref{eq:adaptive-blocking-cert-b} yields
\begin{align*}
\EmpBound{c}{-3}{\BlockingCertificateInterval{t}{b}}
&>
\EmpBound{b}{+3}{\ArmCertificateInterval{t}{b}}
\\
&\ge
\EmpMean{b}{\BlockingCertificateInterval{t}{a}}
-
\ConfRadius{b}{\BlockingCertificateInterval{t}{a}}
+
2\ConfRadius{b}{\ArmCertificateInterval{t}{b}}
\\
&\ge
\EmpBound{b}{-1}{\BlockingCertificateInterval{t}{a}}
\\
&=
\EmpBound{b}{-3}{\BlockingCertificateInterval{t}{a}}
+
2\ConfRadius{b}{\BlockingCertificateInterval{t}{a}}
\\
&>
\EmpBound{a}{+3}{\ArmCertificateInterval{t}{a}}.
\end{align*}
Hence \(c\), together with
\(\ArmCertificateInterval{t}{a}\) and
\(\BlockingCertificateInterval{t}{b}\), satisfies the three-confidence
certificate inequality for \(a\).

It remains to check that \(c\neq a\). If \(c=a\), the preceding strict
inequality implies
\[
\EmpBound{a}{-3}{\BlockingCertificateInterval{t}{b}}
>
\EmpBound{a}{+3}{\ArmCertificateInterval{t}{a}},
\]
and therefore
\[
\EmpBound{a}{-1}{\BlockingCertificateInterval{t}{b}}
>
\EmpBound{a}{+1}{\ArmCertificateInterval{t}{a}}.
\]
Both intervals belong to the current phase of \(a\), so this would have
triggered the change-detection test before classification. Thus
\(c\neq a\), and the displayed tuple is a valid candidate certificate
for \(a\).

\paragraph{Step 2: Compare the certificate separations.}
The original selected certificate for \(a\) and the candidate constructed
above use the same interval \(\ArmCertificateInterval{t}{a}\) for the
blocked arm \(a\). Therefore, to compare their one-confidence
separations, it suffices to compare the lower confidence bounds of their
blocking arms.

The inequalities from Step~1 imply
\[
\EmpBound{c}{-3}{\BlockingCertificateInterval{t}{b}}
>
\EmpBound{b}{-1}{\BlockingCertificateInterval{t}{a}},
\]
and hence
\[
\EmpBound{c}{-1}{\BlockingCertificateInterval{t}{b}}
>
\EmpBound{b}{-1}{\BlockingCertificateInterval{t}{a}}.
\]
Consequently,
\begin{align*}
\EmpBound{c}{-1}{\BlockingCertificateInterval{t}{b}}
-
\EmpBound{a}{+1}{\ArmCertificateInterval{t}{a}}
>
\EmpBound{b}{-1}{\BlockingCertificateInterval{t}{a}}
-
\EmpBound{a}{+1}{\ArmCertificateInterval{t}{a}}.
\end{align*}
Thus the candidate certificate with blocking arm \(c\) has strictly larger
one-confidence separation than the selected certificate with blocking arm
\(b=\BlockingArm{t}{a}\). This contradicts the maximizing definition of
\(\BlockingArm{t}{a}\).

Therefore \(\BlockingArm{t}{a}\in\GoodArms{t}\). Finally, if some arm is
bad, its blocking arm is good; if no arm is bad, every arm is good. Hence
\(\GoodArms{t}\neq\emptyset\).
\end{proof}

\subsubsection{Confidence-Sum Bounds}
\label{app:proof-adaptive-confidence-sums}

We use the current-MSI statistics of
\Cref{app:known-budget-statements} for both histories. Thus, if \(\tau\)
is the first round of the current MSI of arm \(a\)'s class, then
\(\ConfRadius{a}{t}=\ConfRadius{a}{[\tau,t-1]}\), and analogously for
the detection-history statistics.

We first bound the simpler confidence sum over serviced monitoring
observations. Define
\[
\ServedMonitorRounds
:=
\left\{
t\in\AdaptiveStoppedRounds:
\BranchChoice{t}=\MonitoringBranch,\,
\ServiceableObligationSet{\MonitorCandidate{t}}{t}\neq\emptyset
\right\}.
\]
On every \(t\in\ServedMonitorRounds\), the algorithm plays
\(\PlayedArm{t}=\MonitorCandidate{t}\) and adds its reward to the
detection history.

\begin{lemma}[Detection-history confidence sum]
\label{lem:adaptive-detection-confidence-sum}
For the stopped process,
\begin{equation}
\label{eq:adaptive-played-cb}
\sum_{t\in\ServedMonitorRounds}
\TaggedConfRadius{\PlayedArm{t}}{\DetectionTag}{t}
\le
8\sqrt{\LogFactor\NumRounds\TotalDifficulty}.
\end{equation}
\end{lemma}

\paragraph{Proof intuition.}
Within an MSI, every serviced monitoring observation increases the
detection-history count of the played arm. We can therefore sum the
confidence radii separately for every arm--MSI pair and then combine the
pairs using Cauchy--Schwarz, exactly as for the played-arm confidence sum
in the known-budget analysis.

\begin{proof}
By \Cref{lem:adaptive-msi-count}, class \(m\) has at most
\(2\ChangeBudget{m}+1\) MSIs intersecting
\(\AdaptiveStoppedRounds\). Hence the number of relevant arm--MSI pairs is
at most
\[
\sum_{m=1}^{\NumClasses}
(2\ChangeBudget{m}+1)\ClassSize{m}
\le
2\TotalDifficulty.
\]


Fix one such arm--MSI pair. If it contains no rounds from
$\mathcal T_M$, its confidence contribution is zero. Otherwise,
let $n\ge1$ be the number of rounds from $\mathcal T_M$ in this pair.

Before its \(r\)-th serviced monitoring
observation, the detection history contains at least \(r-1\) observations
of that arm in the MSI. Its total confidence contribution is therefore at
most
\[
\sqrt{8\LogFactor}
\left(
1+\sum_{r=1}^{n-1}r^{-1/2}
\right)
\le
2\sqrt{8\LogFactor n}.
\]
Thus the confidence contribution is at most $2\sqrt{8\Lambda n}$
for every $n\ge0$, using the zero contribution when $n=0$.

Let \(n_{a,I}\) denote the number of serviced monitoring rounds assigned
to arm--MSI pair \((a,I)\). Summing over all pairs and applying
Cauchy--Schwarz gives
\begin{align*}
\sum_{t\in\ServedMonitorRounds}
\TaggedConfRadius{\PlayedArm{t}}{\DetectionTag}{t}
\le
2\sqrt{8\LogFactor}
\sum_{(a,I)}\sqrt{n_{a,I}}
\le
2\sqrt{8\LogFactor}
\sqrt{
\left|\{(a,I)\}\right|
\sum_{(a,I)}n_{a,I}
}.
\end{align*}
There are at most \(2\TotalDifficulty\) pairs and at most
\(\NumRounds\) serviced monitoring rounds. Therefore
\[
\sum_{t\in\ServedMonitorRounds}
\TaggedConfRadius{\PlayedArm{t}}{\DetectionTag}{t}
\le
8\sqrt{\LogFactor\NumRounds\TotalDifficulty}.
\]
\end{proof}

The preceding argument applies only to observations that are actually
collected. We also need to control the confidence radius of a good arm
even when that arm is not selected. For this we use the randomized
class--scale selection rule, as in the eligible-arm confidence bound of
the known-budget analysis.

However, we note that the corresponding bound in the known-budget analysis is stated over all
eligible arms, because \texttt{EBF-H} uses a single selection rule in which
bad arms with active monitoring counters remain eligible. The adaptive
algorithm separates these two roles. The classification branch of
\texttt{AEBF-H} selects only from \(\GoodArms{t}\), while bad arms are
sampled separately by the monitoring branch and are controlled through
their detection-history confidence radii. Moreover, all representatives
appearing in the regret decomposition are good by
\Cref{prop:adaptive-blocking-is-good}. It is therefore enough here to
control the largest classification confidence radius among good arms.

\begin{lemma}[Good-arm confidence sum]
\label{lem:adaptive-good-confidence-sum}
For the stopped process,
\begin{equation}
\label{eq:adaptive-good-cb}
\E\!\left[
\sum_{t\in\AdaptiveStoppedRounds}
\max_{a\in\GoodArms{t}}
\ConfRadius{a}{t}
\right]
\le
128
\sqrt{
\NumClasses(\MaxScale+1)\LogFactor
\NumRounds\TotalDifficulty
}.
\end{equation}
\end{lemma}

\paragraph{Proof intuition.}
A good arm with a large confidence radius need not itself be selected.
At each round, we therefore compare the widest good arm with the arm that
the classification rule would choose from the same class at an appropriate
scale. This comparison arm has a confidence radius within a factor two of
the widest arm and is selected with probability at least
\(1/[2\NumClasses(\MaxScale+1)]\). Each time it is selected, its
classification count increases. Counting how often each observation count
can occur then bounds the cumulative confidence radius.

\begin{proof}
Write $\ell:=\log(2TK)$. For each stopped round $t\in\mathcal T_A$,
choose
\[
 x_t\in\arg\max_{a\in\mathcal G_t}\operatorname{CB}_a(t),\qquad
 k_t:=N_{x_t}(t),\qquad
 h_t:=\lceil\log_2(k_t+1)\rceil,
\]
using the fixed tie-breaking rule. Define
\[
 b_t\in\arg\min_{a\in\mathcal G_t\cap
              \mathcal A_{\operatorname{class}(x_t)}}\tau_a^{h_t},
\]
with the same tie-breaking rule as the classification proposal.
The arm $x_t$ has the largest classification confidence radius, whereas
$b_t$ is the arm the classification rule would select from its class at
scale $h_t$. The symbol $j_t$ continues to denote the algorithm's fresh,
uniformly sampled scale; it is not redefined in this proof.

The quantities $x_t,k_t,h_t,b_t$ are determined before the fresh
round-$t$ draws, and $0\le h_t\le J_{\max}$. Let $u_t$ be the common
current-MSI start of $x_t$ and $b_t$. Since
$k_t<2^{h_t}\le 2(k_t+1)$, the $2^{h_t}$-th most recent classification
observation of $x_t$ in its current phase precedes $u_t$, if it exists.
The recency rule therefore gives
\[
 \tau_{b_t}^{h_t}\le\tau_{x_t}^{h_t}<u_t,
 \qquad N_{b_t}(t)<2^{h_t}\le 2(k_t+1).
\]
If $k_t=0$, then $h_t=0$ and $N_{b_t}(t)=0$; if $k_t\ge1$, then
$2(k_t+1)\le4k_t$. Thus
\[
 \max\{N_{b_t}(t),1\}\le4\max\{k_t,1\},
 \qquad
 \max_{a\in\mathcal G_t}\operatorname{CB}_a(t)
 =\operatorname{CB}_{x_t}(t)
 \le2\operatorname{CB}_{b_t}(t).
\]

Conditional on $\mathcal F_{t-1}$, the event
\[
 A_t:=\{\chi_t=\mathrm{cls},\;
        m_t=\operatorname{class}(x_t),\;j_t=h_t\}
\]
has probability at least
\[
 p_{\mathrm{cls}}:=\frac{1}{2M(J_{\max}+1)}.
\]
On $A_t$, the algorithm plays $b_t$ and records a classification
observation of that arm. For $n\in\{0,\ldots,T-1\}$, set
\[
 V_n:=\sum_{t\in\mathcal T_A}\mathbf 1\{N_{b_t}(t)=n\},
 \qquad
 S_n:=\sum_{t\in\mathcal T_A}\mathbf 1\{N_{b_t}(t)=n\}\mathbf 1\{A_t\}.
\]
Every round counted by $S_n$ supplies the $(n+1)$-st classification
observation of a particular arm in a particular MSI. No arm--MSI pair
can do this twice. The stopped reset-counting argument bounds the number
of such pairs by $2D$, so $S_n\le2D$ pathwise. Since membership in
$\mathcal T_A$ and the condition $N_{b_t}(t)=n$ are determined before
the fresh draws,
\[
 p_{\mathrm{cls}}\mathbb E[V_n]\le\mathbb E[S_n]\le2D.
\]

The exact confidence-sum identity is
\[
 \sum_{t\in\mathcal T_A}\operatorname{CB}_{b_t}(t)
 =\sqrt{8\ell}\sum_{n=0}^{T-1}
       \frac{V_n}{\sqrt{\max\{n,1\}}}.
\]
Set $\beta:=Tp_{\mathrm{cls}}/(2D)$. If $\beta<1$, then
\[
 \sum_{t\in\mathcal T_A}\max_{a\in\mathcal G_t}
       \operatorname{CB}_a(t)
 \le T\sqrt{8\ell}
 \le4\sqrt{\frac{\ell TD}{p_{\mathrm{cls}}}}.
\]
For $\beta\ge1$, the inequalities
$\mathbb E[V_n]\le2D/p_{\mathrm{cls}}$ and
$\sum_{0\le n<\beta}1/\sqrt{\max\{n,1\}}\le2\sqrt\beta$
bound the small-count contribution by
$8\sqrt{\ell TD/p_{\mathrm{cls}}}$. On the remaining rounds,
each radius is at most $\sqrt{8\ell/\beta}$, so their contribution
is at most $4\sqrt{\ell TD/p_{\mathrm{cls}}}$. Therefore
\[
 \mathbb E\!\left[\sum_{t\in\mathcal T_A}
                  \max_{a\in\mathcal G_t}\operatorname{CB}_a(t)\right]
 \le24\sqrt{\frac{\ell TD}{p_{\mathrm{cls}}}}
 =24\sqrt{2M(J_{\max}+1)\ell TD}
 \le128\sqrt{M(J_{\max}+1)\Lambda TD},
\]
where the final step uses $\ell\le\Lambda$ and $24\sqrt2\le128$.
This also covers $\beta<1$ and proves the stated bound.
\end{proof}

\subsubsection{Regret Decomposition}
\label{app:adaptive-regret-decomposition}

We begin by defining the two good arms that will serve as reference points
in the regret decomposition. For the played arm, define
\[
\RepresentativeArm{t}
:=
\begin{cases}
\PlayedArm{t},
&
\PlayedArm{t}\in\GoodArms{t},
\\
\BlockingArm{t}{\PlayedArm{t}},
&
\PlayedArm{t}\in\BadArms{t}.
\end{cases}
\]
Thus, if the played arm is good, it represents itself; if it is bad, we
replace it by the blocking arm selected by its classification certificate.

Similarly, for the optimal arm define
\[
\RepresentativeOptimalArm{t}
:=
\begin{cases}
\OptimalArm{t},
&
\OptimalArm{t}\in\GoodArms{t},
\\
\BlockingArm{t}{\OptimalArm{t}},
&
\OptimalArm{t}\in\BadArms{t}.
\end{cases}
\]
Hence the optimal arm also represents itself whenever it is good, while a
bad optimal arm is represented by the arm that currently certifies it as
bad.

By \Cref{prop:adaptive-blocking-is-good}, the blocking arm of every bad
arm is good. Consequently,
\(\RepresentativeArm{t}\in\GoodArms{t}\) and
\(\RepresentativeOptimalArm{t}\in\GoodArms{t}\) at every round. This is
the purpose of the representatives: they replace any bad arm appearing in
the regret comparison by a good arm whose current confidence radius can be
controlled. Importantly, when an arm is bad, its representative is chosen
using its historical classification certificate; we do not assume that
the representative's current mean is larger than the mean of the arm it
represents.

Adding and subtracting the means of these two representatives gives
\[
\MeanReward{\OptimalArm{t}}{t}
-
\MeanReward{\PlayedArm{t}}{t}
=
\underbrace{
\MeanReward{\RepresentativeArm{t}}{t}
-
\MeanReward{\PlayedArm{t}}{t}
}_{\text{Term (i)}}
+
\underbrace{
\MeanReward{\RepresentativeOptimalArm{t}}{t}
-
\MeanReward{\RepresentativeArm{t}}{t}
}_{\text{Term (ii)}}
+
\underbrace{
\MeanReward{\OptimalArm{t}}{t}
-
\MeanReward{\RepresentativeOptimalArm{t}}{t}
}_{\text{Term (iii)}}.
\]
The three terms isolate the same statistical effects as in the
known-budget analysis. Term (i) is nonzero only when the algorithm plays a
bad arm and measures the cost of monitoring that arm relative to its good
representative. Term (ii) compares two good arms and is therefore
controlled by their current confidence radii. Term (iii) is nonzero only
when the optimal arm is bad and measures the loss caused by delayed
detection of a stale optimal-arm certificate.

Define the two representatives as in \Cref{def:representatives}, using the
adaptive good and bad sets. By \Cref{prop:adaptive-blocking-is-good}, both
representatives are good. Hence the algebraic decomposition
\eqref{eq:regret-decomposition} applies unchanged:
\[
\MeanReward{\OptimalArm{t}}{t}
-
\MeanReward{\PlayedArm{t}}{t}
=
\underbrace{
\MeanReward{\RepresentativeArm{t}}{t}
-
\MeanReward{\PlayedArm{t}}{t}
}_{\text{Term (i)}}
+
\underbrace{
\MeanReward{\RepresentativeOptimalArm{t}}{t}
-
\MeanReward{\RepresentativeArm{t}}{t}
}_{\text{Term (ii)}}
+
\underbrace{
\MeanReward{\OptimalArm{t}}{t}
-
\MeanReward{\RepresentativeOptimalArm{t}}{t}
}_{\text{Term (iii)}}.
\]
As in the known-budget analysis, Term (i) is the cost of playing a bad
arm for monitoring, Term (ii) compares the two good representatives, and
Term (iii) captures the delay in recognizing an optimal arm whose
certificate has become stale. We bound the three terms separately.

\paragraph{Term (i): monitoring bad arms.}

Term (i) vanishes whenever the played arm is good. If a bad arm is played,
it must be servicing a monitoring obligation. We charge the resulting
dominance-gap contribution to that obligation.

\begin{lemma}[Bounding Term (i)]
\label{lem:adaptive-term-one}
For the stopped process,
\[
\E\!\left[
\sum_{t\in\AdaptiveStoppedRounds}
\left(
\MeanReward{\RepresentativeArm{t}}{t}
-
\MeanReward{\PlayedArm{t}}{t}
\right)
\right]
\le
4096\LogFactor^2
\sqrt{\NumClasses\NumRounds\TotalDifficulty}.
\]
\end{lemma}

\paragraph{Proof intuition.}
When a bad arm is played, its selected certificate compares it with a good
blocking arm. The no-reset condition lets us compare the historical
certificate with the current classification and detection statistics.
This bounds the regret by two confidence terms and the arm's dominance
gap. The confidence terms are controlled by
\Cref{lem:adaptive-detection-confidence-sum,lem:adaptive-good-confidence-sum}.
For the dominance gap, a scale-\(j\) obligation is charged at most
\(\ServiceTarget{j}\) times, and the expected number of launched
obligations is controlled by the adaptive launch probability.

\begin{proof}
Fix \(t\in\AdaptiveStoppedRounds\). The term is zero if
\(\PlayedArm{t}\in\GoodArms{t}\). Otherwise, let
\(a:=\PlayedArm{t}\) and
\(b:=\RepresentativeArm{t}=\BlockingArm{t}{a}\).

The selected three-confidence certificate for \(a\) implies
\[
\DominanceGap{t}{a}
\ge
2\ConfRadius{b}{\BlockingCertificateInterval{t}{a}}
+
2\ConfRadius{a}{\ArmCertificateInterval{t}{a}}.
\]
Neither certificate interval can be empty: an empty-history
three-confidence lower bound is negative, whereas an empty-history
three-confidence upper bound exceeds one.

Since these certificates exist after the round-\(t\) reset step, the
corresponding classes were not reset at round \(t\). Hence the reset test
did not invoke on the relevant intervals. For the blocking arm \(b\), its
\((\ClassificationTag,\ClassificationTag)\) specialization gives
\[
\EmpBound{b}{-1}{t}
\le
\EmpBound{b}{+1}{\BlockingCertificateInterval{t}{a}},
\]
while for \(a\), its
\((\ClassificationTag,\DetectionTag)\) specialization gives
\[
\EmpBound{a}{-1}{\ArmCertificateInterval{t}{a}}
\le
\TaggedEmpBound{a}{\DetectionTag}{+1}{t}.
\]
If either current-MSI interval is empty, the corresponding comparison
follows directly from the empty-history convention.

Since \(t\in\AdaptiveStoppedRounds\), all statistics available at the
start of round \(t\) satisfy the concentration inequalities of
\Cref{lem:adaptive-concentration}. Combining concentration with the
preceding no-reset comparisons gives
\[
\MeanReward{b}{t}
-
\left[
\EmpBound{b}{-1}{\BlockingCertificateInterval{t}{a}}
+
\DominanceGap{t}{a}
\right]
\le
2\ConfRadius{b}{t},
\]
and
\[
\left[
\EmpBound{a}{+1}{\ArmCertificateInterval{t}{a}}
-
\DominanceGap{t}{a}
\right]
-
\MeanReward{a}{t}
\le
2\TaggedConfRadius{a}{\DetectionTag}{t}.
\]
By definition,
\[
\EmpBound{b}{-1}{\BlockingCertificateInterval{t}{a}}
-
\EmpBound{a}{+1}{\ArmCertificateInterval{t}{a}}
=
\DominanceGap{t}{a}.
\]
Adding the three contributions yields
\begin{equation}
\label{eq:adaptive-term-one}
\MeanReward{\RepresentativeArm{t}}{t}
-
\MeanReward{\PlayedArm{t}}{t}
\le
2\ConfRadius{\RepresentativeArm{t}}{t}
+
3\DominanceGap{t}{\PlayedArm{t}}
+
2\TaggedConfRadius{\PlayedArm{t}}{\DetectionTag}{t}.
\end{equation}

We next bound the dominance-gap contribution. Every bad arm played by
\texttt{AEBF-H} has a serviceable obligation. Assign
\(3\DominanceGap{t}{\PlayedArm{t}}\) to the first such obligation under a
fixed deterministic ordering, and let \(j\) be its scale. Since
\(j\le\DominanceScale{t}{\PlayedArm{t}}\), we have
\(\DominanceGap{t}{\PlayedArm{t}}\le2^{1-j}\). Thus the obligation
receives a charge of at most \(6\cdot2^{-j}\) at this round.

Each service decreases the obligation's remaining count by one, so a
scale-\(j\) obligation receives at most \(\ServiceTarget{j}\) charges
before completion. Its total charge is therefore at most
\(6\cdot2^{-j}\ServiceTarget{j}\).

Fix \(m\in[\NumClasses]\) and \(j\in\{0,\ldots,\MaxScale\}\). For
\(a\in\ClassArms{m}\), let \(\AdaptiveLaunchIndicator{a}{j}{t}\) denote
the indicator that a scale-\(j\) obligation is launched for \(a\) at
round \(t\). Since the classification and launch probability are
determined before the launch seed is examined,
\[
\E\!\left[
\AdaptiveLaunchIndicator{a}{j}{t}
\mid
\PreRoundHistory{t}
\right]
=
\Indicator{
a\in\BadArms{t},\,
j\le\min\{\DominanceScale{t}{a},\MaxScale\}
}
\AdaptiveLaunchProbability{m}{j}{t}
\le
\AdaptiveLaunchProbability{m}{j}{t}.
\]

On every stopped round,
\(\DetectedChangeCount{m}{t}+1\le\ChangeBudget{m}+1\). Moreover,
\(\ServiceTarget{j}=2^{2j+7}\LogFactor\le129\LogFactor2^{2j}\).
Membership in \(\AdaptiveStoppedRounds\) is
\(\PreRoundHistory{t}\)-measurable, so the tower property gives
\begin{align*}
&\E\!\left[
\sum_{t\in\AdaptiveStoppedRounds}
\sum_{a\in\ClassArms{m}}
6\cdot2^{-j}\ServiceTarget{j}
\AdaptiveLaunchIndicator{a}{j}{t}
\right]
\\
&\qquad\le
\E\!\left[
\sum_{t\in\AdaptiveStoppedRounds}
\sum_{a\in\ClassArms{m}}
6\cdot2^{-j}\ServiceTarget{j}
\AdaptiveLaunchProbability{m}{j}{t}
\right]
\\
&\qquad\le
774
\sqrt{
\LogFactor\NumRounds
(\ChangeBudget{m}+1)\ClassSize{m}
}.
\end{align*}
Summing over the \(\MaxScale+1\) scales and applying Cauchy--Schwarz over
classes bounds all dominance-gap charges by
\[
774(\MaxScale+1)
\sqrt{
\LogFactor\NumClasses\NumRounds\TotalDifficulty
}.
\]

Finally, the first confidence term in \eqref{eq:adaptive-term-one} is
controlled by \eqref{eq:adaptive-good-cb}, since the representative is
good, and the detection-history term is controlled by
\eqref{eq:adaptive-played-cb}. Using
\(\MaxScale+1\le3\LogFactor\), \(\LogFactor\ge1\), and
\(\NumClasses\ge1\) proves the lemma.
\end{proof}

\paragraph{Term (ii): comparing good representatives.}

Both representatives are good, so this term is controlled entirely by
their current classification confidence radii.

\begin{lemma}[Bounding Term (ii)]
\label{lem:adaptive-term-two}
For the stopped process,
\[
\E\!\left[
\sum_{t\in\AdaptiveStoppedRounds}
\left(
\MeanReward{\RepresentativeOptimalArm{t}}{t}
-
\MeanReward{\RepresentativeArm{t}}{t}
\right)
\right]
\le
4096\LogFactor^2
\sqrt{\NumClasses\NumRounds\TotalDifficulty}.
\]
\end{lemma}

\paragraph{Proof intuition.}
Since both representatives are good, their current-MSI intervals cannot
form a three-confidence certificate. Concentration converts this empirical
comparison into a bound by their confidence radii, whose cumulative sum is
controlled by \Cref{lem:adaptive-good-confidence-sum}.

\begin{proof}
Fix \(t\in\AdaptiveStoppedRounds\), and let
\(a':=\RepresentativeOptimalArm{t}\) and
\(a:=\RepresentativeArm{t}\). Both arms belong to \(\GoodArms{t}\).

If \(a=a'\), the contribution is zero. Otherwise, the classification rule
does not declare \(a\) bad using \(a'\) and their two current-MSI
intervals. Hence
\[
\EmpMean{a'}{t}-\EmpMean{a}{t}
\le
3\ConfRadius{a'}{t}
+
3\ConfRadius{a}{t}.
\]
If either current-MSI history is empty, the same inequality follows
directly from the empty-history convention.

Since \(t\in\AdaptiveStoppedRounds\), concentration holds for both
current-MSI statistics. Therefore
\[
\MeanReward{a'}{t}-\MeanReward{a}{t}
\le
4\ConfRadius{a'}{t}
+
4\ConfRadius{a}{t}.
\]
Both radii are at most
\(\max_{u\in\GoodArms{t}}\ConfRadius{u}{t}\). Applying
\eqref{eq:adaptive-good-cb} gives
\[
\E\!\left[
\sum_{t\in\AdaptiveStoppedRounds}
\left(
\MeanReward{\RepresentativeOptimalArm{t}}{t}
-
\MeanReward{\RepresentativeArm{t}}{t}
\right)
\right]
\le
1024
\sqrt{
\NumClasses(\MaxScale+1)\LogFactor
\NumRounds\TotalDifficulty
}.
\]
Using \(\MaxScale+1\le3\LogFactor\) and \(\LogFactor\ge1\) proves the
lemma.
\end{proof}

\paragraph{Term (iii): delayed detection.}

Term (iii) is nonzero only when the optimal arm is classified as bad.
In this case its classification certificate may contain observations from
before a recent change, so the historical certificate need not reflect the
arm's current mean.

\begin{lemma}[Bounding Term (iii)]
\label{lem:adaptive-term-three}
For the stopped process,
\[
\E\!\left[
\sum_{t\in\AdaptiveStoppedRounds}
\left(
\MeanReward{\OptimalArm{t}}{t}
-
\MeanReward{\RepresentativeOptimalArm{t}}{t}
\right)
\right]
\le
8192\LogFactor^2
\sqrt{\NumClasses\NumRounds\TotalDifficulty}
+2.
\]
\end{lemma}

\paragraph{Proof intuition.}
Suppose the optimal arm is currently classified as bad. Its representative
is its good blocking arm. We compare the current mean of the optimal arm
with the historical upper confidence bound that caused it to be classified
as bad. The difference between these two quantities is the
\emph{certificate residual}.

Small residuals can be absorbed directly into confidence and
class-dependent threshold terms. For a large residual at scale \(j\), the
algorithm launches scale-\(j\) monitoring obligations for the optimal arm.
While class \(m\) has detected \(e-1\) changes, the launch probability is
fixed and gives a geometric waiting time of order \(e^{-1/2}\). Once a
useful obligation is launched, sufficiently many fresh monitoring samples
make the detection interval incompatible with the stale classification
certificate and force a reset.

We first introduce the notation used in this argument and then prove the
three ingredients: completion of an obligation forces detection, a useful
obligation is launched quickly enough, and the resulting large residuals
have the desired cumulative bound.

\paragraph{Residuals and hard rounds.}

Let
\(
\HardRounds
:=
\left\{
t\in\AdaptiveStoppedRounds:
\OptimalArm{t}\in\BadArms{t}
\right\}.
\)
Thus \(\HardRounds\) contains exactly the stopped rounds on which
Term (iii) may be nonzero.

Fix \(t\in\HardRounds\), and write
\(a^\star:=\OptimalArm{t}\). Since \(a^\star\) is bad, its selected
classification certificate contains an interval
\(\ArmCertificateInterval{t}{a^\star}\) for \(a^\star\). Define its
\emph{certificate residual} by
\begin{equation}
\label{eq:adaptive-residual-definition}
r_t := \mu_{a_t^*}(t)
       - \widehat\mu_{a_t^*}^{(+1)}
         \bigl(I_t(a_t^*)\bigr),
\qquad t\in\mathcal U.
\end{equation}
Hence a positive residual means that the current mean of the optimal arm
lies above the historical upper confidence bound used in its certificate.

We also introduce a class-dependent threshold for separating residuals
that can be summed directly from those that require monitoring. For each
class \(m\in[\NumClasses]\), let
\[
\ClassOptimalRounds{m}
:=
\left|
\left\{
t\in[\NumRounds]:
\OptimalArm{t}\in\ClassArms{m}
\right\}
\right|
\]
be the number of rounds in the full horizon whose optimal arm belongs to
class \(m\). The fixed tie-breaking rule makes the optimal arm unique, so
\(\sum_{m=1}^{\NumClasses}\ClassOptimalRounds{m}=\NumRounds\).

Define
\[
\ClassThreshold{m}
:=
\begin{cases}
\displaystyle
\LogFactor
\sqrt{
\frac{
\NumClasses\ClassSize{m}(\ChangeBudget{m}+1)
}{
\ClassOptimalRounds{m}
}
},
&
\ClassOptimalRounds{m}>0,
\\[2ex]
0,
&
\ClassOptimalRounds{m}=0.
\end{cases}
\]
These thresholds are used only in the analysis; the algorithm does not
know the change budgets.

We call a stopped round a \emph{large-residual round} if the optimal arm is
bad and its certificate residual exceeds both its dominance gap and its
class threshold:
\[
\LargeResidualRounds
:=
\left\{
t\in\HardRounds:
\KnownBudgetOptimalResidual{t}
>
\max\left\{
\DominanceGap{t}{\OptimalArm{t}},
\ClassThreshold{\ArmClass{\OptimalArm{t}}}
\right\}
\right\}.
\]

To match such residuals with the monitoring scales, for
\(m\in[\NumClasses]\) and \(j\in\{1,\ldots,\MaxScale\}\), define
\[
\HardScaleRounds{m}{j}
:=
\left\{
t\in\LargeResidualRounds:
\OptimalArm{t}\in\ClassArms{m},\;
2^{-j}
<
\KnownBudgetOptimalResidual{t}
\le
2^{1-j}
\right\}.
\]
Thus \(\HardScaleRounds{m}{j}\) consists of the large-residual rounds for
class \(m\) whose residual is at resolution \(2^{-j}\).

Residuals in \(\LargeResidualRounds\) that are at most
\(2^{-\MaxScale}\) are not assigned to any of these scales. Their total
contribution is at most
\(\NumRounds 2^{-\MaxScale}\le1\).

\paragraph{First reduction of Term (iii).}

Fix \(t\in\HardRounds\), write
\(a^\star:=\OptimalArm{t}\), and let
\(b:=\RepresentativeOptimalArm{t}=\BlockingArm{t}{a^\star}\).
By \Cref{prop:adaptive-blocking-is-good}, \(b\in\GoodArms{t}\).

Let
\(\BlockingCertificateInterval{t}{a^\star}\) be the interval for \(b\)
in the selected certificate of \(a^\star\). By the definition of the
dominance gap,
\[
\DominanceGap{t}{a^\star}
=
\EmpBound{b}{-1}
{\BlockingCertificateInterval{t}{a^\star}}
-
\EmpBound{a^\star}{+1}
{\ArmCertificateInterval{t}{a^\star}}.
\]
Using \eqref{eq:adaptive-residual-definition}, we therefore have the exact
identity
\begin{align*}
\MeanReward{a^\star}{t}-\MeanReward{b}{t}
&=
\KnownBudgetOptimalResidual{t}
-
\DominanceGap{t}{a^\star}
\\
&\quad+
\EmpBound{b}{-1}
{\BlockingCertificateInterval{t}{a^\star}}
-
\MeanReward{b}{t}.
\end{align*}

The class of \(b\) was not reset at the start of round \(t\), because the
certificate interval displayed above is still available. Since the reset
test did not fire, comparing this certificate interval with the current
classification history of \(b\) gives
\[
\EmpBound{b}{-1}
{\BlockingCertificateInterval{t}{a^\star}}
\le
\EmpBound{b}{+1}{t}.
\]
Since \(t\in\AdaptiveStoppedRounds\), concentration holds for the current
MSI statistic of \(b\). Hence
\[
\EmpBound{b}{-1}
{\BlockingCertificateInterval{t}{a^\star}}
-
\MeanReward{b}{t}
\le
2\ConfRadius{b}{t}.
\]
Consequently,
\begin{equation}
\label{eq:adaptive-term-three-pointwise}
\MeanReward{\OptimalArm{t}}{t}
-
\MeanReward{\RepresentativeOptimalArm{t}}{t}
\le
\KnownBudgetOptimalResidual{t}
-
\DominanceGap{t}{\OptimalArm{t}}
+
2\ConfRadius{\RepresentativeOptimalArm{t}}{t}.
\end{equation}

If
\(\KnownBudgetOptimalResidual{t}
\le\DominanceGap{t}{\OptimalArm{t}}\),
the first two terms on the right are nonpositive. If the residual exceeds
the dominance gap but is at most
\(\ClassThreshold{\ArmClass{\OptimalArm{t}}}\), their difference is at
most that threshold. All remaining rounds belong to
\(\LargeResidualRounds\). Summing \eqref{eq:adaptive-term-three-pointwise}
therefore gives
\begin{align}
\label{eq:adaptive-first-split}
&\sum_{t\in\AdaptiveStoppedRounds}
\left[
\MeanReward{\OptimalArm{t}}{t}
-
\MeanReward{\RepresentativeOptimalArm{t}}{t}
\right]
\\
&\quad\le
2
\sum_{t\in\AdaptiveStoppedRounds}
\ConfRadius{\RepresentativeOptimalArm{t}}{t}
+
\sum_{m=1}^{\NumClasses}
\ClassOptimalRounds{m}\ClassThreshold{m}
+
\sum_{t\in\LargeResidualRounds}
\KnownBudgetOptimalResidual{t}.
\nonumber
\end{align}

The first sum is controlled by
\Cref{lem:adaptive-good-confidence-sum}, because every representative is
good. For the threshold contribution, the definition above and
Cauchy--Schwarz give
\begin{align}
\label{eq:adaptive-class-threshold}
\sum_{m=1}^{\NumClasses}
\ClassOptimalRounds{m}\ClassThreshold{m}
&=
\LogFactor\sqrt{\NumClasses}
\sum_{\substack{m\in[\NumClasses]:\\
\ClassOptimalRounds{m}>0}}
\sqrt{
\ClassOptimalRounds{m}
\ClassSize{m}
(\ChangeBudget{m}+1)
}
\nonumber\\
&\le
\LogFactor\sqrt{\NumClasses}
\sqrt{
\sum_{m=1}^{\NumClasses}\ClassOptimalRounds{m}
}
\sqrt{
\sum_{m=1}^{\NumClasses}
\ClassSize{m}(\ChangeBudget{m}+1)
}
\nonumber\\
&=
\LogFactor
\sqrt{
\NumClasses\NumRounds\TotalDifficulty
}.
\end{align}
It remains to control the large-residual sum.

\paragraph{Monitoring opportunities and completion.}

Recall that a scale-\(j\) monitoring obligation requires
\[
\ServiceTarget{j}
=
2^{2j+7}\LogFactor
\]
services before it is completed.

Fix a class \(m\) and an arm \(a\in\ClassArms{m}\). We call round \(t\)
a \emph{monitoring slot for \(a\)} if
\(\BranchChoice{t}=\MonitoringBranch\) and
\(\MonitorCandidate{t}=a\). Conditional on the information available
before these two fresh random draws, the probability of such a slot is
\[
\MonitoringSlotProbability{m}
:=
\frac{1}{2\NumClasses\ClassSize{m}}.
\]
If \(a\) has a serviceable obligation, a monitoring slot supplies one new
detection-history observation of \(a\) and advances every serviceable
obligation of that arm.

For \(m\in[\NumClasses]\) and
\(j\in\{1,\ldots,\MaxScale\}\), define
\begin{equation}
\label{eq:adaptive-service-window}
\CompletionOpportunityCount{m}{j}
:=
8
\NumClasses
\ClassSize{m}
\ServiceTarget{j}.
\end{equation}
Using the definition of \(\ServiceTarget{j}\),
\[
\CompletionOpportunityCount{m}{j}
\le
1033
\NumClasses
\ClassSize{m}
\LogFactor
2^{2j}.
\]
Moreover,
\[
\CompletionOpportunityCount{m}{j}
\MonitoringSlotProbability{m}
=
4\ServiceTarget{j}.
\]
Thus a block of
\(\CompletionOpportunityCount{m}{j}\) hard opportunities contains four
times the required number of monitoring slots in expectation.

\begin{lemma}[Completion within a stationary segment]
\label{lem:adaptive-clean-completion}
Fix \(m\in[\NumClasses]\),
\(j\in\{1,\ldots,\MaxScale\}\), and
\(s\in\HardScaleRounds{m}{j}\). Let
\(a:=\OptimalArm{s}\). Suppose a scale-\(j\) obligation for \(a\) is
launched at round \(s\) and receives \(\ServiceTarget{j}\) services
before the next true class-\(m\) change. Then, at the start of the round
immediately following completion, class \(m\) is reset or that round does
not belong to \(\AdaptiveStoppedRounds\).
\end{lemma}

\paragraph{Proof intuition.}
At the launch round \(s\), the residual is larger than \(2^{-j}\), so the
current mean of \(a\) lies noticeably above its stale classification
upper bound. After \(\ServiceTarget{j}\) fresh detection observations,
the detection confidence radius is at most \(2^{-j-2}\). If the class has
not changed again, the fresh detection lower bound must therefore exceed
the stale classification upper bound, which triggers the cross-history
reset test.

\begin{proof}
Fix \(m,j,s\) as in the lemma and let \(a:=\OptimalArm{s}\). If the
stopped process ends before the obligation completes, the second
alternative in the statement already holds.

Otherwise, let \(u\) be the round in which the obligation receives its
\(\ServiceTarget{j}\)-th service. By assumption, no true class-\(m\)
change occurs after \(s\) and through \(u\), so \([s,u]\) lies inside one
true class-\(m\) stationary segment.

A class-\(m\) reset deletes every live obligation of class \(m\).
Therefore, because the obligation launched at \(s\) survives until its
completion at \(u\), no class-\(m\) reset occurs between \(s\) and \(u\).
Hence both the detection interval \([s,u]\) and the original
classification interval
\(\ArmCertificateInterval{s}{a}\) belong to the same class-\(m\) phase
at the start of round \(u+1\).

The interval \([s,u]\) contains at least \(\ServiceTarget{j}\) detection
observations of \(a\). Consequently,
\[
\TaggedConfRadius{a}{\DetectionTag}{[s,u]}
\le
\sqrt{
\frac{
8\LogFactor
}{
2^{2j+7}\LogFactor
}
}
=
2^{-j-2}.
\]

If the observation at round \(u\) creates the first failure of the
concentration inequalities, then \(u+1=\AdaptiveStopTime\), so the
following round is outside the stopped process.

Otherwise, concentration holds for the detection statistic on \([s,u]\).
Because \(s\in\HardScaleRounds{m}{j}\),
\(\KnownBudgetOptimalResidual{s}>2^{-j}\). By the definition of the
residual,
\[
\MeanReward{a}{s}
>
\EmpBound{a}{+1}{\ArmCertificateInterval{s}{a}}
+
2^{-j}.
\]
Since the mean of \(a\) is constant on \([s,u]\), concentration gives
\begin{align*}
\TaggedEmpBound{a}{\DetectionTag}{-1}{[s,u]}
&\ge
\MeanReward{a}{s}
-
2\TaggedConfRadius{a}{\DetectionTag}{[s,u]}
\\
&>
\EmpBound{a}{+1}{\ArmCertificateInterval{s}{a}}
+
2^{-j}
-
2^{-j-1}
\\
&>
\EmpBound{a}{+1}{\ArmCertificateInterval{s}{a}}.
\end{align*}
If \(u=\NumRounds\), there is no following round in the stopped process.
Otherwise, the last inequality is precisely a
\((\DetectionTag,\ClassificationTag)\) witness for the reset condition
\eqref{eq:adaptive-reset-test}. Thus class \(m\) resets at the start of
round \(u+1\).
\end{proof}

\paragraph{Detected episodes and their launch probabilities.}

We next control how long the process waits before such a useful
obligation is launched.

For a class \(m\in[\NumClasses]\) and an integer \(e\ge1\), define its
\(e\)-th \emph{detected episode} by
\[
\DetectedEpisode{m}{e}
:=
\left\{
t\in\AdaptiveStoppedRounds:
\DetectedChangeCount{m}{t}+1=e
\right\}.
\]
If nonempty, this is the consecutive set of stopped rounds after the
\((e-1)\)-st class-\(m\) reset and before the next class-\(m\) reset, or
before the stopped process ends. Throughout this episode,
\(\DetectedChangeCount{m}{t}=e-1\).

Fix a scale \(j\). Since the detected-change count is constant throughout
the episode, the class-\(m\), scale-\(j\) launch probability is also
constant. Denote this probability by
\[
\AdaptiveEpisodeLaunchProbability{m}{j}{e}
:=
2^{-j}
\sqrt{
\frac{
e
}{
\NumRounds
\ClassSize{m}
\LogFactor
}
}.
\]
For every \(t\in\DetectedEpisode{m}{e}\),
\[
\AdaptiveLaunchProbability{m}{j}{t}
=
\AdaptiveEpisodeLaunchProbability{m}{j}{e}.
\]
To see that the cap in \eqref{eq:adaptive-launch-rate} is inactive, note
that on a nonempty episode
\(e=\DetectedChangeCount{m}{t}+1\le t\le\NumRounds\), while
\(\ClassSize{m}\ge1\), \(\LogFactor\ge1\), and \(2^{-j}\le1\).

\paragraph{An auxiliary hard-round sequence.}

A subtlety is that a successful launch may later cause a reset, which
changes future classifications and hence changes which future rounds are
hard. We therefore define an auxiliary trajectory whose hard rounds can be
fixed without looking at the class-\(m\), scale-\(j\) launch seeds.

Fix \(m\in[\NumClasses]\),
\(j\in\{1,\ldots,\MaxScale\}\), and \(e\ge1\), and suppose first that
\(\DetectedEpisode{m}{e}\neq\emptyset\). Let
\(
s_e:=\min\DetectedEpisode{m}{e}
\)
be the first round of the episode.

Starting from the state immediately after the reset step at round \(s_e\),
construct an auxiliary execution using the same round-indexed rewards and
the same algorithmic random seeds as the actual process, except that every
class-\(m\) reset update is suppressed. Thus, when the class-\(m\) reset
test fires in the auxiliary execution, we keep \(\PhaseStart{m}\) fixed,
do not increment \(\DetectedChangeCount{m}{t}\), and do not delete
class-\(m\) obligations. Resets of all other classes are processed
normally.

The auxiliary execution continues through round $T$ without imposing any
concentration stopping rule. Neither membership in $\mathcal T_A$ nor an
auxiliary counterpart of that condition is used to define its frozen hard
sequence. Actual stopping enters only through the definition of reached
rounds below.

A suppressed reset may make the auxiliary good set empty. To keep the
auxiliary recursion well defined in that case, define its classification
proposal to be the first arm under the fixed arm ordering. This convention
is needed only after the auxiliary and actual executions have diverged; it
is never invoked while the actual detected episode is still active.

We now define three auxiliary objects.

\paragraph{Frozen hard sequence.}

Use the superscript $\mathrm{aux}$ for classifications and certificate
statistics computed along the auxiliary execution. Whenever
$a_t^*\in\mathcal B_t^{\mathrm{aux}}$, define
\[
 r_t^{\mathrm{aux}}
 := \mu_{a_t^*}(t)
    - \widehat\mu_{a_t^*}^{\mathrm{aux},(+1)}
      \bigl(I_t^{\mathrm{aux}}(a_t^*)\bigr).
\]
Let $\mathcal H_{m,j,e}^{\mathrm{aux}}$ be the ordered sequence of all
rounds $t\in\{s_e,\ldots,T\}$ satisfying
\[
 \begin{aligned}
 a_t^*
   &\in \mathcal A_m\cap\mathcal B_t^{\mathrm{aux}},\\
 r_t^{\mathrm{aux}}
   &> \max\{\Delta_t^{\mathrm{aux}}(a_t^*),\zeta_m\},\\
 2^{-j}
   &< r_t^{\mathrm{aux}}\le 2^{1-j}.
 \end{aligned}
\]
In particular, this definition does not require $t\in\mathcal T_A$.

We say that the actual execution \emph{reaches} an element
\(t\in\FrozenHardSequence{m}{j}{e}\) if
\(t\in\DetectedEpisode{m}{e}\). While the detected episode is active, no
class-\(m\) reset has yet separated the actual and auxiliary trajectories,
so their classification histories agree. Hence the reached elements are
exactly
\[
\HardScaleRounds{m}{j}
\cap
\DetectedEpisode{m}{e}.
\]

\paragraph{Core opportunity sequence.}
Partition \(\FrozenHardSequence{m}{j}{e}\) according to the true
class-\(m\) stationary segments. In each part, remove its final
\(\CompletionOpportunityCount{m}{j}\) elements, or remove the whole part
if it contains fewer than this many elements. Concatenate the remaining
elements in their original order and call the resulting sequence
\(\CoreOpportunitySequence{m}{j}{e}\).

The purpose of this trimming is that every element of
\(\CoreOpportunitySequence{m}{j}{e}\) has at least
\(\CompletionOpportunityCount{m}{j}\) later frozen hard rounds in the
same true stationary segment. Thus a useful obligation launched at a core
round has enough future hard opportunities to be completed before the
next true class change.

\paragraph{Core waiting count.}
Recall that \(\LaunchSeed{a}{j}{t}\sim\UniformUnitInterval\) is the latent
uniform seed used to decide whether a scale-\(j\) obligation for arm \(a\)
is launched at round \(t\).

Let \(\CoreWaitingCount{m}{j}{e}\) be the number of reached elements of
\(\CoreOpportunitySequence{m}{j}{e}\), through and including the first
reached core round \(t\) satisfying
\[
\LaunchSeed{\OptimalArm{t}}{j}{t}
\le
\AdaptiveEpisodeLaunchProbability{m}{j}{e}.
\]
If no such successful launch occurs before the detected episode ends,
\(\CoreWaitingCount{m}{j}{e}\) is the total number of reached core
elements.

Finally, if \(\DetectedEpisode{m}{e}=\emptyset\), we use the convention
\[
\FrozenHardSequence{m}{j}{e}
=
\CoreOpportunitySequence{m}{j}{e}
=
\emptyset,
\qquad
\CoreWaitingCount{m}{j}{e}=0.
\]

\begin{lemma}[Waiting time on hard class--scale rounds]
\label{lem:adaptive-coin-invariant-core}
Fix \(m\in[\NumClasses]\),
\(j\in\{1,\ldots,\MaxScale\}\), and \(e\ge1\). Then
\[
\E\!\left[
\CoreWaitingCount{m}{j}{e}
\right]
\le
\AdaptiveEpisodeLaunchProbability{m}{j}{e}^{-1}.
\]

Moreover, suppose a reached core round
\(s\in\CoreOpportunitySequence{m}{j}{e}\) successfully launches a
scale-\(j\) obligation for \(a:=\OptimalArm{s}\). If the actual process
reaches the next \(\CompletionOpportunityCount{m}{j}\) frozen hard rounds
in the same true class-\(m\) stationary segment and at least
\(\ServiceTarget{j}\) of them are monitoring slots for \(a\), then class
\(m\) resets no later than the start of the following round, or the
stopped process has ended by then.
\end{lemma}

\paragraph{Proof intuition.}
We expose all randomness except the class-\(m\), scale-\(j\) launch seeds.
The auxiliary hard sequence is then already fixed. Consequently, each
reached core round reveals a fresh independent launch coin with the same
success probability. A reset can only stop the sequence early, so it
cannot increase the waiting time.

\begin{proof}
Fix \(m,j,e\). The result is immediate when
\(\DetectedEpisode{m}{e}=\emptyset\), so assume the episode is nonempty.

Let \(s\in[\NumRounds]\) be a possible deterministic episode start. The
event
\[
\left\{
\min\DetectedEpisode{m}{e}=s
\right\}
\]
is \(\PreRoundHistory{s}\)-measurable: whether \(s\) belongs to the
stopped process, whether a class-\(m\) reset occurs at its start, and the
detected-change count after that reset step are all determined before the
fresh random draws of round \(s\).

Recall the latent-seed representation of the launch rule: before the game,
sample mutually independent
\(\LaunchSeed{a}{h}{u}\sim\UniformUnitInterval\) for every arm \(a\),
scale \(h\), and round \(u\). Whenever the algorithm considers a
scale-\(h\) obligation for \(a\) at round \(u\), it launches the
obligation exactly when
\[
\LaunchSeed{a}{h}{u}
\le
\AdaptiveLaunchProbability{\ArmClass{a}}{h}{u}.
\]

On the event
\(\{\min\DetectedEpisode{m}{e}=s\}\), expose
\(\PreRoundHistory{s}\), the entire round-indexed reward array
\(\{\ObservedReward{a}{u}\}_{a,u}\), and all future algorithmic random
seeds except
\[
\left\{
\LaunchSeed{a}{j}{u}:
a\in\ClassArms{m},\;
u\in[s,\NumRounds]
\right\}.
\]
The index set of these omitted seeds is deterministic once \(m,j,s\) are
fixed and does not depend on the future hard sequence.

To make the independence explicit, construct the auxiliary execution
described above while additionally suppressing every class-\(m\),
scale-\(j\) obligation whose launch seed belongs to the omitted
collection. We claim that this additional suppression does not change the
auxiliary classification trajectory.

Indeed, argue inductively over rounds. On a classification branch,
identical phase starts and classification histories produce the same good
and bad sets, the same class and scale choices, the same classification
proposal, and the same recorded classification reward. On a monitoring
branch, no classification reward is recorded, including on a fallback
play, and the monitoring candidate is sampled independently of the live
obligations.

Changing the omitted launch seeds can therefore affect only the
class-\(m\), scale-\(j\) obligations and the class-\(m\) detection
history. It does not alter any classification history. Resets of classes
other than \(m\) are therefore unchanged, while class-\(m\) resets are
suppressed by construction and hence cannot alter its phase start.
Consequently, all classifications, certificates, and residuals in the
auxiliary trajectory are unchanged.




This invariance applies to the entire unstopped auxiliary classification
trajectory. Conditional on the exposed episode-start history, reward
array, and other seeds, both $\mathcal H_{m,j,e}^{\mathrm{aux}}$ and
$\mathcal H_{m,j,e}^{\circ}$ are therefore fixed independently of the
omitted launch seeds. Those seeds may change the actual detection
history, concentration stopping time, and next class-$m$ reset, but
these affect only which frozen rounds are reached, not the frozen
sequences themselves.

Until the actual detected episode ends, its classification trajectory
agrees with the auxiliary one. At every reached frozen hard round $t$,
\[
 \Delta_t(a_t^*) < r_t \le 2^{1-j},
\]
so $j\le i_t(a_t^*)$ and the algorithm examines the scale-$j$ launch
seed of the optimal arm. Throughout the episode, its success
probability is $q:=q_{m,j,e}$.

Conditional on the exposed randomness, the core rounds and their
associated optimal arms are fixed. If $G_{m,j,e}>r$, then the first
$r$ core rounds exist, are reached, and all their launch seeds exceed
$q$. These are distinct independent uniform seeds. Thus the conditional
probability of this event is at most $(1-q)^r$, regardless of any
earlier termination caused by noncore launches or concentration
failure. Averaging over the additionally exposed randomness gives,
for every integer $r\ge0$,

\[
\Prob\!\left(
\left\{
\min\DetectedEpisode{m}{e}=s
\right\}
\cap
\left\{
\CoreWaitingCount{m}{j}{e}>r
\right\}
\,\middle|\,
\PreRoundHistory{s}
\right)
\le
\Indicator{\min\DetectedEpisode{m}{e}=s}
(1-q)^r.
\]
A launch at a noncore round may cause an earlier reset and shorten the
episode, but it cannot make the event on the left more likely.

Summing the geometric tail probabilities gives
\[
\E\!\left[
\CoreWaitingCount{m}{j}{e}
\Indicator{\min\DetectedEpisode{m}{e}=s}
\right]
\le
\frac{
\Prob(\min\DetectedEpisode{m}{e}=s)
}{q}.
\]
The possible episode-start events are disjoint, so summing over
\(s\in[\NumRounds]\) yields
\[
\E\!\left[
\CoreWaitingCount{m}{j}{e}
\right]
\le
q^{-1}.
\]

For the second claim, suppose a reached core round \(s\) successfully
launches a scale-\(j\) obligation for \(a=\OptimalArm{s}\). By the
definition of the core sequence, there are at least
\(\CompletionOpportunityCount{m}{j}\) later frozen hard rounds in the
same true stationary segment.

Unless a class-\(m\) reset or \(\AdaptiveStopTime\) occurs first, these
rounds are also reached by the actual process. On every such hard round,
the global optimal arm belongs to class \(m\). Since the means of all
class-\(m\) arms and the deterministic tie-breaking rule are unchanged
inside the stationary segment, this optimal arm is the same arm \(a\).
Moreover, the hard-scale condition gives
\(j\le\DominanceScale{t}{a}\), so the obligation remains serviceable.

Hence \(\ServiceTarget{j}\) monitoring slots for \(a\) complete the
obligation. \Cref{lem:adaptive-clean-completion} then implies that class
\(m\) resets by the following round or the stopped process has already
ended.
\end{proof}

\paragraph{Bounding the large residuals.}

We can now combine the waiting-time and completion arguments.

\begin{lemma}[Regret on hard class--scale rounds]
\label{lem:adaptive-hard-residual}
For the stopped process,
\[
\E\!\left[
\sum_{m=1}^{\NumClasses}
\sum_{j=1}^{\MaxScale}
\sum_{t\in\HardScaleRounds{m}{j}}
\KnownBudgetOptimalResidual{t}
\right]
\le
4096\LogFactor^2
\sqrt{\NumClasses\NumRounds\TotalDifficulty}
+1.
\]
\end{lemma}

\paragraph{Proof intuition.}
Fix a class \(m\), scale \(j\), and detected episode \(e\). Before the
first useful launch, each hard round contributes at most \(2^{1-j}\), while
the expected number of such rounds is at most the inverse launch
probability. The factor \(2^{-j}\) in the launch probability cancels the
residual scale, leaving a contribution proportional to \(e^{-1/2}\).

After a successful launch, a block of
\(\CompletionOpportunityCount{m}{j}\) hard rounds contains enough
monitoring slots with very high probability. The obligation then completes
and forces a reset. We separately charge the few hard rounds removed near
stationary-segment boundaries.

\begin{proof}
Fix \(m\in[\NumClasses]\) and
\(j\in\{1,\ldots,\MaxScale\}\).

Every \(t\in\HardScaleRounds{m}{j}\) satisfies
\[
\KnownBudgetOptimalResidual{t}
\le
2^{1-j}
\]
and
\[
\DominanceGap{t}{\OptimalArm{t}}
<
\KnownBudgetOptimalResidual{t}
\le
2^{1-j}.
\]
Hence
\(j\le\DominanceScale{t}{\OptimalArm{t}}\). In particular, a scale-\(j\)
obligation for the optimal arm is serviceable on every such round.

Within a true class-\(m\) stationary segment, all class-\(m\) means are
fixed. Since every hard \((m,j)\) round has its global optimum in class
\(m\), the deterministic tie-breaking rule selects the same class-\(m\)
optimal arm on all such rounds in that segment.

\paragraph{Enough monitoring slots after a launch.}

Fix a deterministic candidate launch round \(s\in[\NumRounds]\), and let
\(a:=\OptimalArm{s}\). If \(a\notin\ClassArms{m}\), then \(s\) cannot be
a class-\(m\) core launch and there is nothing to consider.

Otherwise, follow the actual execution after \(s\) until the first of:
the next class-\(m\) reset, the next true class-\(m\) change, or
\(\AdaptiveStopTime\). Among the rounds before this boundary, mark the
successive rounds belonging to \(\HardScaleRounds{m}{j}\), stopping after
\(\CompletionOpportunityCount{m}{j}\) marked rounds have been reached.

Whether a round is marked is determined by \(\PreRoundHistory{t}\).
Conditional on the information available before the fresh branch and
monitor-candidate draws at a marked round, the probability that the round
is a monitoring slot for \(a\) is
\(p_m:=\MonitoringSlotProbability{m}\).

Let \(X\in\{0,1\}\) be the indicator of such a monitoring slot. For every
\(\theta>0\),
\[
\E[e^{-\theta X}\mid\text{past}]
=
1-p_m+p_me^{-\theta}
\le
\exp\!\left(
-p_m(1-e^{-\theta})
\right).
\]
Therefore the exponential of minus \(\theta\) times the number of
monitoring slots, plus
\(p_m(1-e^{-\theta})\) times the number of reached marked rounds, is a
nonnegative supermartingale. Hold both counts fixed after the stopping
boundary.

Let \(W:=\CompletionOpportunityCount{m}{j}\),
\(n:=\ServiceTarget{j}\), and
\(p:=\MonitoringSlotProbability{m}\). Recall that \(Wp=4n\).

Fix the candidate launch round \(s\) and launch arm
\(a:=\OptimalArm{s}\). Starting after round \(s\), follow the actual
execution until the first of the following events: a class-\(m\) reset, a
true class-\(m\) change, \(\AdaptiveStopTime\), or the \(W\)-th subsequent
round in \(\HardScaleRounds{m}{j}\). Let \(\tau\) denote this stopping
round.

For \(u>s\), let \(N_u\) be the number of marked hard \((m,j)\) rounds
reached by time \(u\), capped at \(W\), and let \(S_u\) be the number
among these reached marked rounds that are monitoring slots for \(a\).
Both processes are kept constant after \(\tau\).

Consider a reached marked round \(u\). Whether \(u\) is marked is
\(\PreRoundHistory{u}\)-measurable. Conditional on
\(\PreRoundHistory{u}\), the fresh branch and monitoring-candidate draws
are independent of the past, and
\[
\Prob\!\left(
\BranchChoice{u}=\MonitoringBranch,\,
\MonitorCandidate{u}=a
\,\middle|\,
\PreRoundHistory{u}
\right)
=
p.
\]
Thus, if \(X_u\) denotes the indicator that the marked round \(u\) is a
monitoring slot for \(a\), then for every \(\theta>0\),
\[
\E\!\left[
e^{-\theta X_u}
\,\middle|\,
\PreRoundHistory{u}
\right]
=
1-p+pe^{-\theta}
\le
\exp\!\left(-p(1-e^{-\theta})\right).
\]

Consequently,
\[
M_u
:=
\exp\!\left(
-\theta S_u
+
p(1-e^{-\theta})N_u
\right)
\]
is a nonnegative supermartingale with \(M_s=1\). Since
\(\tau\le\NumRounds\), bounded optional stopping gives
\(\E[M_\tau]\le1\).

Let \(\mathcal F_{s,m,j}\) be the event that all \(W\) marked hard rounds
are reached before the stopping boundary, but fewer than \(n\) of them
are monitoring slots for \(a\). On this event,
\(N_\tau=W\) and \(S_\tau<n\). Therefore
\[
M_\tau
=
\exp\!\left(
-\theta S_\tau
+
p(1-e^{-\theta})W
\right)
\ge
\exp\!\left(
-\theta n
+
p(1-e^{-\theta})W
\right).
\]
Hence
\[
1
\ge
\E[M_\tau]
\ge
\Prob(\mathcal F_{s,m,j})
\exp\!\left(
-\theta n
+
p(1-e^{-\theta})W
\right),
\]
and therefore
\[
\Prob(\mathcal F_{s,m,j})
\le
\exp\!\left(
\theta n
-
p(1-e^{-\theta})W
\right).
\]

Choosing \(\theta=\log4\), so that
\(1-e^{-\theta}=3/4\), and using \(Wp=4n\), gives
\begin{align*}
\Prob(\mathcal F_{s,m,j})
&\le
\exp\!\left(
n\log4-\frac34Wp
\right)
\\
&=
\exp\!\left(
-(3-\log4)n
\right)
\\
&\le
e^{-n}
=
e^{-\ServiceTarget{j}},
\end{align*}
where the last inequality uses \(3-\log4>1\).

Importantly, this bounds the \emph{joint} event that all \(W\) hard
opportunities are reached and that fewer than \(n\) monitoring slots
occur among them. We do not condition on the future event that these
\(W\) opportunities are reached; their adaptive reachability is handled
directly by the stopped counting processes \(N_u\) and \(S_u\).

Define \(\ServiceSuccessEvent\) to be the event that the preceding service
condition succeeds for every reached core launch, over all classes, scales,
and candidate launch rounds. Since
\(\ServiceTarget{j}\ge128\LogFactor\), a union bound gives
\[
\Prob(\ServiceSuccessEvent^c)
\le
\NumClasses\MaxScale\NumRounds
e^{-128\LogFactor}
\le
\NumRounds^{-2}.
\]
When \(\NumRounds=1\), there are no scales
\(j\in\{1,\ldots,\MaxScale\}\), so the event holds trivially.

\paragraph{Waiting before the first core launch.}

Fix a nonempty detected episode
\(\DetectedEpisode{m}{e}\). By
\Cref{lem:adaptive-coin-invariant-core},
\[
\E[
\CoreWaitingCount{m}{j}{e}
]
\le
\AdaptiveEpisodeLaunchProbability{m}{j}{e}^{-1}.
\]
Every reached core round in
\(\HardScaleRounds{m}{j}\) contributes at most \(2^{1-j}\) residual.
Therefore the expected residual accumulated through the first successful
core launch is at most
\begin{align*}
\frac{
2^{1-j}
}{
\AdaptiveEpisodeLaunchProbability{m}{j}{e}
}
=
\frac{
2^{1-j}
}{
2^{-j}
\sqrt{
e/(
\NumRounds\ClassSize{m}\LogFactor
)
}
}
=
2
\sqrt{
\frac{
\NumRounds\ClassSize{m}\LogFactor
}{
e
}
}.
\end{align*}

By \Cref{lem:adaptive-reset-counting}, the detected-change count never
exceeds \(\ChangeBudget{m}\) on the stopped process. Thus there are at
most \(\ChangeBudget{m}+1\) nonempty detected episodes. Summing the
preceding bound over them gives
\begin{align*}
2
\sqrt{
\NumRounds\ClassSize{m}\LogFactor
}
\sum_{e=1}^{\ChangeBudget{m}+1}
e^{-1/2}
\le
4
\sqrt{
\NumRounds
\ClassSize{m}
\LogFactor
(\ChangeBudget{m}+1)
},
\end{align*}
where we used
\(\sum_{e=1}^{L+1}e^{-1/2}\le2\sqrt{L+1}\).

\paragraph{Post-launch and boundary rounds.}

On \(\ServiceSuccessEvent\), once the first successful core launch occurs
in an episode, at most
\(\CompletionOpportunityCount{m}{j}\) further hard rounds can be reached
in the same stationary segment before a class-\(m\) reset or
\(\AdaptiveStopTime\). Indeed, if all these opportunities are reached,
the service-success event supplies at least \(\ServiceTarget{j}\)
monitoring slots, and
\Cref{lem:adaptive-coin-invariant-core,lem:adaptive-clean-completion}
forces the reset.

Moreover, on $\mathcal E_{\mathrm{srv}}$, an episode with a successful
core launch cannot remain active into the next true class-$m$
stationary segment. Otherwise it would reach all of the $W_{m,j}$
later frozen hard rounds guaranteed by the core definition in the
launch segment. The service condition and completion lemma would
then end the episode, or the stopped process, no later than the first
round of the next segment. Thus the $W_{m,j}$ post-launch charge is
paid at most once per detected episode.

There are also hard rounds deliberately removed when constructing the core
sequence. Each intersection of a detected episode with a true class-\(m\)
stationary segment contributes at most
\(\CompletionOpportunityCount{m}{j}\) removed rounds.

Before \(\AdaptiveStopTime\), class \(m\) has at most
\(\ChangeBudget{m}\) resets by
\Cref{lem:adaptive-reset-counting} and at most
\(\ChangeBudget{m}\) true changes. The union of the reset and change
boundaries therefore partitions the stopped horizon into at most
\(2\ChangeBudget{m}+1\) detected-episode/stationary-segment
intersections.

Consequently, the union of the post-launch hard rounds and the hard rounds
removed from the core contains at most
\[
3(\ChangeBudget{m}+1)
\CompletionOpportunityCount{m}{j}
\]
rounds. It also contains at most
\(\ClassOptimalRounds{m}\) rounds, because every hard
\((m,j)\) round has its optimal arm in class \(m\).

Let \(\mathcal R_{m,j}\) denote the hard \((m,j)\) rounds that are not
already charged to the core waiting contribution. By the preceding
argument,
\[
|\mathcal R_{m,j}|
\le
3(\ChangeBudget{m}+1)\CompletionOpportunityCount{m}{j},
\]
while also
\[
|\mathcal R_{m,j}|
\le
\ClassOptimalRounds{m},
\]
because every \(t\in\mathcal R_{m,j}\) has
\(\OptimalArm{t}\in\ClassArms{m}\).

Moreover, every \(t\in\mathcal R_{m,j}\) belongs to
\(\HardScaleRounds{m}{j}\), and therefore
\(\KnownBudgetOptimalResidual{t}\le 2^{1-j}\). Hence
\begin{align*}
\sum_{t\in\mathcal R_{m,j}}
\KnownBudgetOptimalResidual{t}
&\le
2^{1-j}|\mathcal R_{m,j}|
\\
&\le
2^{1-j}
\min\!\left\{
\ClassOptimalRounds{m},\,
3(\ChangeBudget{m}+1)
\CompletionOpportunityCount{m}{j}
\right\}
\\
&\le
6\cdot2^{-j}
\min\!\left\{
\ClassOptimalRounds{m},\,
(\ChangeBudget{m}+1)
\CompletionOpportunityCount{m}{j}
\right\}.
\end{align*}
The last inequality uses
\(\min\{x,3y\}\le 3\min\{x,y\}\).

Using \(\min\{x,y\}\le\sqrt{xy}\) and
\eqref{eq:adaptive-service-window}, this is at most
\[
256
\sqrt{
\NumClasses
\ClassSize{m}
\LogFactor
(\ChangeBudget{m}+1)
\ClassOptimalRounds{m}
},
\]
where we used \(6\sqrt{1033}\le256\).

\paragraph{Summing over classes and scales.}




Write $\Gamma:=\sqrt{MTD}$ and define
\[
 \begin{aligned}
 H
   &:= \sum_{m=1}^{M}\sum_{j=1}^{J_{\max}}
       \sum_{t\in\mathcal U_{m,j}} r_t,\\
 R_{\mathrm{wait}}
   &:= \sum_{m=1}^{M}\sum_{j=1}^{J_{\max}}
       \sum_{e=1}^{L_m+1} 2^{1-j}G_{m,j,e}.
 \end{aligned}
\]
Here $R_{\mathrm{wait}}$ is a nonnegative random upper bound on the
core waiting contribution, with $G_{m,j,e}=0$ for empty episodes.
The waiting-time bound of Lemma D.12 and Cauchy--Schwarz give the
unconditional estimate
\[
 \mathbb E[R_{\mathrm{wait}}]
 \le 4J_{\max}\sqrt{\Lambda}
       \sum_{m=1}^{M}\sqrt{TK_m(L_m+1)}
 \le 4J_{\max}\sqrt{\Lambda}\,\Gamma.
\]
The preceding post-launch and boundary calculation gives, pathwise
on $\mathcal E_{\mathrm{srv}}$,
\[
 H \le R_{\mathrm{wait}}
       +256J_{\max}\sqrt{\Lambda}\,\Gamma.
\]
On every realization, the sets $\mathcal U_{m,j}$ are disjoint and
every residual in them lies in $(0,1]$, so $0\le H\le T$.
Consequently, on every realization,
\[
 H \le R_{\mathrm{wait}}
       +256J_{\max}\sqrt{\Lambda}\,\Gamma
       +T\mathbf 1\{\mathcal E_{\mathrm{srv}}^c\}.
\]
Taking unconditional expectations and using
$\mathbb P(\mathcal E_{\mathrm{srv}}^c)\le T^{-2}$,
$J_{\max}\le3\Lambda$, and $\Lambda\ge1$, we obtain
\[
 \begin{aligned}
 \mathbb E[H]
 &\le 260J_{\max}\sqrt{\Lambda}\,\Gamma+1\\
 &\le 780\Lambda^{3/2}\Gamma+1\\
 &\le 4096\Lambda^2\Gamma+1.
 \end{aligned}
\]
No independence between $R_{\mathrm{wait}}$ and
$\mathcal E_{\mathrm{srv}}$ is needed. This proves the lemma.
\end{proof}

\begin{proof}[Proof of \Cref{lem:adaptive-term-three}]
We first recall why the three contributions in
\eqref{eq:adaptive-first-split} cover all of Term~(iii).

If \(t\notin\HardRounds\), then
\(\OptimalArm{t}\in\GoodArms{t}\), so by definition
\(\RepresentativeOptimalArm{t}=\OptimalArm{t}\). Hence the Term~(iii)
summand is zero.

Now fix \(t\in\HardRounds\). By
\eqref{eq:adaptive-term-three-pointwise},
\[
\MeanReward{\OptimalArm{t}}{t}
-
\MeanReward{\RepresentativeOptimalArm{t}}{t}
\le
2\ConfRadius{\RepresentativeOptimalArm{t}}{t}
+
\KnownBudgetOptimalResidual{t}
-
\DominanceGap{t}{\OptimalArm{t}}.
\]
Write \(m=\ArmClass{\OptimalArm{t}}\). There are three possibilities for
the residual term.

First, if
\(\KnownBudgetOptimalResidual{t}
\le
\DominanceGap{t}{\OptimalArm{t}}\),
then
\[
\KnownBudgetOptimalResidual{t}
-
\DominanceGap{t}{\OptimalArm{t}}
\le0,
\]
so only the confidence term remains.

Second, suppose
\[
\DominanceGap{t}{\OptimalArm{t}}
<
\KnownBudgetOptimalResidual{t}
\le
\ClassThreshold{m}.
\]
Then
\[
\KnownBudgetOptimalResidual{t}
-
\DominanceGap{t}{\OptimalArm{t}}
\le
\ClassThreshold{m}.
\]
Summing these contributions over rounds whose optimal arm lies in class
\(m\) costs at most
\(\ClassOptimalRounds{m}\ClassThreshold{m}\).

Finally, if
\[
\KnownBudgetOptimalResidual{t}
>
\max\left\{
\DominanceGap{t}{\OptimalArm{t}},
\ClassThreshold{m}
\right\},
\]
then, by definition, \(t\in\LargeResidualRounds\). On these rounds we
simply upper-bound
\[
\KnownBudgetOptimalResidual{t}
-
\DominanceGap{t}{\OptimalArm{t}}
\le
\KnownBudgetOptimalResidual{t}.
\]

These three cases partition all rounds in \(\HardRounds\). Together with
the fact that Term~(iii) vanishes outside \(\HardRounds\), they give
\eqref{eq:adaptive-first-split}:
\begin{align*}
&\sum_{t\in\AdaptiveStoppedRounds}
\left(
\MeanReward{\OptimalArm{t}}{t}
-
\MeanReward{\RepresentativeOptimalArm{t}}{t}
\right)
\\
&\qquad\le
2
\sum_{t\in\AdaptiveStoppedRounds}
\ConfRadius{\RepresentativeOptimalArm{t}}{t}
+
\sum_{m=1}^{\NumClasses}
\ClassOptimalRounds{m}\ClassThreshold{m}
+
\sum_{t\in\LargeResidualRounds}
\KnownBudgetOptimalResidual{t}.
\end{align*}

It remains to bound these three sums.

For the first sum,
\(\RepresentativeOptimalArm{t}\in\GoodArms{t}\) by
\Cref{prop:adaptive-blocking-is-good}. Hence
\Cref{lem:adaptive-good-confidence-sum} gives
\[
2
\E\!\left[
\sum_{t\in\AdaptiveStoppedRounds}
\ConfRadius{\RepresentativeOptimalArm{t}}{t}
\right]
\le
256
\sqrt{
\NumClasses(\MaxScale+1)
\LogFactor
\NumRounds
\TotalDifficulty
}.
\]

For the second sum, \eqref{eq:adaptive-class-threshold} gives
\[
\sum_{m=1}^{\NumClasses}
\ClassOptimalRounds{m}\ClassThreshold{m}
\le
\LogFactor
\sqrt{
\NumClasses
\NumRounds
\TotalDifficulty
}.
\]

It remains to bound the residual sum over
\(\LargeResidualRounds\). Split these rounds according to whether their
residual is at most \(2^{-\MaxScale}\).

The rounds satisfying
\(\KnownBudgetOptimalResidual{t}\le2^{-\MaxScale}\) contribute at most
\[
\sum_{\substack{t\in\LargeResidualRounds:\\
\KnownBudgetOptimalResidual{t}\le2^{-\MaxScale}}}
\KnownBudgetOptimalResidual{t}
\le
\NumRounds\,2^{-\MaxScale}
\le1.
\]

Every remaining large-residual round satisfies
\(\KnownBudgetOptimalResidual{t}>2^{-\MaxScale}\). Such a round has a
unique optimal-arm class \(m\), and there is a unique
\(j\in\{1,\ldots,\MaxScale\}\) such that
\[
2^{-j}
<
\KnownBudgetOptimalResidual{t}
\le
2^{1-j}.
\]
Hence these rounds are partitioned by the sets
\(\HardScaleRounds{m}{j}\). Therefore
\[
\sum_{\substack{t\in\LargeResidualRounds:\\
\KnownBudgetOptimalResidual{t}>2^{-\MaxScale}}}
\KnownBudgetOptimalResidual{t}
=
\sum_{m=1}^{\NumClasses}
\sum_{j=1}^{\MaxScale}
\sum_{t\in\HardScaleRounds{m}{j}}
\KnownBudgetOptimalResidual{t}.
\]
Applying \Cref{lem:adaptive-hard-residual} gives
\[
\E\!\left[
\sum_{\substack{t\in\LargeResidualRounds:\\
\KnownBudgetOptimalResidual{t}>2^{-\MaxScale}}}
\KnownBudgetOptimalResidual{t}
\right]
\le
4096
\LogFactor^2
\sqrt{
\NumClasses
\NumRounds
\TotalDifficulty
}
+1.
\]

Combining the confidence, threshold, small-residual, and hard-residual
contributions yields
\begin{align*}
&\E\!\left[
\sum_{t\in\AdaptiveStoppedRounds}
\left(
\MeanReward{\OptimalArm{t}}{t}
-
\MeanReward{\RepresentativeOptimalArm{t}}{t}
\right)
\right]
\\
&\qquad\le
256
\sqrt{
\NumClasses(\MaxScale+1)
\LogFactor
\NumRounds
\TotalDifficulty
}
\\
&\qquad\quad+
\LogFactor
\sqrt{
\NumClasses
\NumRounds
\TotalDifficulty
}
\\
&\qquad\quad+
4096
\LogFactor^2
\sqrt{
\NumClasses
\NumRounds
\TotalDifficulty
}
+2.
\end{align*}
Here one additive \(1\) comes from the residuals below the smallest
maintained scale, and the other comes from the failure probability in
\Cref{lem:adaptive-hard-residual}.

Finally, using
\(\MaxScale+1\le3\LogFactor\) and \(\LogFactor\ge1\), we obtain
\[
\E\!\left[
\sum_{t\in\AdaptiveStoppedRounds}
\left(
\MeanReward{\OptimalArm{t}}{t}
-
\MeanReward{\RepresentativeOptimalArm{t}}{t}
\right)
\right]
\le
8192
\LogFactor^2
\sqrt{
\NumClasses
\NumRounds
\TotalDifficulty
}
+2.
\]
This proves \Cref{lem:adaptive-term-three}.
\end{proof}

\subsubsection{Proof of the Unknown-Budget Expected Dynamic Regret Theorem}
\label{app:proof-adaptive-main}

We now combine the three regret contributions and account for the rounds
discarded by the stopped-process argument.

\begin{theorem}[Restatement of \Cref{thm:adaptive-main}]
For every admissible instance, \texttt{AEBF-H} satisfies
\[
\E[\DynamicRegret]
\le
16384
\LogFactor^2
\sqrt{
\NumClasses
\NumRounds
\TotalDifficulty
}
+3.
\]
Equivalently,
\[
\E[\DynamicRegret]
=
\BigOTilde\!\left(
\sqrt{
\NumClasses
\NumRounds
\TotalDifficulty
}
\right).
\]
\end{theorem}

\begin{proof}
Recall the three-term decomposition from
\Cref{app:adaptive-regret-decomposition}. For every round \(t\),
\begin{align*}
\MeanReward{\OptimalArm{t}}{t}
-
\MeanReward{\PlayedArm{t}}{t}
&=
\left(
\MeanReward{\RepresentativeArm{t}}{t}
-
\MeanReward{\PlayedArm{t}}{t}
\right)
\\
&\quad+
\left(
\MeanReward{\RepresentativeOptimalArm{t}}{t}
-
\MeanReward{\RepresentativeArm{t}}{t}
\right)
\\
&\quad+
\left(
\MeanReward{\OptimalArm{t}}{t}
-
\MeanReward{\RepresentativeOptimalArm{t}}{t}
\right).
\end{align*}
These are Terms (i), (ii), and (iii), respectively.

By \Cref{lem:adaptive-term-one}, the expected stopped contribution of
Term~(i) is at most
\[
4096
\LogFactor^2
\sqrt{
\NumClasses
\NumRounds
\TotalDifficulty
}.
\]
By \Cref{lem:adaptive-term-two}, the expected stopped contribution of
Term~(ii) is at most
\[
4096
\LogFactor^2
\sqrt{
\NumClasses
\NumRounds
\TotalDifficulty
}.
\]
Finally, by \Cref{lem:adaptive-term-three}, the expected stopped
contribution of Term~(iii) is at most
\[
8192
\LogFactor^2
\sqrt{
\NumClasses
\NumRounds
\TotalDifficulty
}
+2.
\]

Adding the three bounds gives
\begin{align*}
&\E\!\left[
\sum_{t\in\AdaptiveStoppedRounds}
\left(
\MeanReward{\OptimalArm{t}}{t}
-
\MeanReward{\PlayedArm{t}}{t}
\right)
\right]
\\
&\qquad\le
16384
\LogFactor^2
\sqrt{
\NumClasses
\NumRounds
\TotalDifficulty
}
+2.
\end{align*}

It remains to account for the rounds after the stopping time. By
\Cref{lem:adaptive-concentration},
\[
\Prob\!\left((\ConcentrationEvent')^c\right)
\le
\NumRounds^{-1}.
\]
By construction of \(\AdaptiveStopTime\), the omitted suffix
\([\NumRounds]\setminus\AdaptiveStoppedRounds\) is nonempty only if
\(\ConcentrationEvent'\) fails. Since the instantaneous regret is at most
one, the regret over this suffix is at most \(\NumRounds\). Therefore
\[
\E\!\left[
\sum_{t\in[\NumRounds]\setminus\AdaptiveStoppedRounds}
\left(
\MeanReward{\OptimalArm{t}}{t}
-
\MeanReward{\PlayedArm{t}}{t}
\right)
\right]
\le
\NumRounds
\Prob\!\left((\ConcentrationEvent')^c\right)
\le
1.
\]
Combining the stopped and omitted contributions yields
\[
\E[\DynamicRegret]
\le
16384
\LogFactor^2
\sqrt{
\NumClasses
\NumRounds
\TotalDifficulty
}
+3.
\]
By $\LogFactor$ definition, the theorem applies.
\end{proof}

\section{Additional Related Work}
\label{app:related-work}

\noindent\textbf{Adversarial online learning and reinforcement learning.}
Adversarial online learning allows arbitrary loss sequences and measures
adaptation through quantities such as comparator switches, interval-wise
regret, or comparator path length~\citep{auer2002nonstochastic,cutkosky2020parameter}.
In adversarial reinforcement learning, \citet{jin2020adversarial} establish
regret against the best fixed policy for episodic Markov decision processes
(MDPs) with adversarial
losses, unknown transitions, and bandit feedback, while
\citet{zhao2022dynamic} study dynamic regret against changing policies in
online MDPs with adversarial losses. Related nonstationary
reinforcement-learning reductions give
dynamic-regret guarantees in terms of global change counts or variation
budgets~\citep{wei2021non}. These works adapt to global temporal or
comparator complexity, whereas our model localizes nonstationarity across
known classes of arms.

\noindent\textbf{Learning with corrupted feedback.}
Corruption-robust bandit models start from a stochastic instance but allow
an adaptive adversary to alter rewards subject to a total corruption budget;
their guarantees degrade with this budget while retaining stochastic behavior
when it is zero~\citep{lykouris2018stochastic,gupta2019better}. This framework
has been extended to linear bandits and episodic reinforcement learning with
corrupted rewards or transitions~\citep{bogunovic2021stochastic,lykouris2021corruption}.
Corruption budgets measure the total magnitude of adversarial alterations
around a fixed stochastic instance, whereas our class-specific budgets count
persistent mean changes and localize them by class.

\noindent\textbf{Additional related work} exploits structure in evolving environments. \emph{Contextual} and \emph{linear} nonstationary bandits use additional information in the form of contexts, policy classes, or linear reward models~\citep{DBLP:conf/icml/RakhlinS16,luo2018efficient,DBLP:conf/nips/RussacVC19}. These works exploit structure in the reward model or context space, whereas our focus is on the \emph{location of nonstationarity}: different arm groups may evolve at different rates.
Restless bandits provide another model of evolving rewards, with different
assumptions on how arm states change over time~\citep{slivkins2008adapting,wang2020restless,seznec2020single,Jiang2023online}.

\end{document}

%% file: math_commands.tex
\usepackage{amsmath,amsfonts,bm}

\def\eqref#1{equation~\ref{#1}}

\def\1{\bm{1}}

\DeclareMathAlphabet{\mathsfit}{\encodingdefault}{\sfdefault}{m}{sl}
\SetMathAlphabet{\mathsfit}{bold}{\encodingdefault}{\sfdefault}{bx}{n}

\newcommand{\E}{\mathbb{E}}

\DeclareMathOperator*{\argmax}{arg\,max}
\DeclareMathOperator*{\argmin}{arg\,min}

%% file: iclr2027_conference.bib
@article{hoeffding1963probability,
  title={Probability Inequalities for Sums of Bounded Random Variables},
  author={Hoeffding, Wassily},
  journal={Journal of the American Statistical Association},
  volume={58},
  number={301},
  pages={13--30},
  year={1963},
  doi={10.1080/01621459.1963.10500830}
}

@book{lattimore2020bandit,
  title={Bandit algorithms},
  author={Lattimore, Tor and Szepesv{\'a}ri, Csaba},
  year={2020},
  publisher={Cambridge University Press}
}

@article{garivier2008upper,
  title={On upper-confidence bound policies for non-stationary bandit problems},
  author={Garivier, Aur{\'e}lien and Moulines, Eric},
  journal={arXiv preprint arXiv:0805.3415},
  year={2008}
}

@inproceedings{auer2019adaptively,
  title={Adaptively tracking the best bandit arm with an unknown number of distribution changes},
  author={Auer, Peter and Gajane, Pratik and Ortner, Ronald},
  booktitle={Conference on Learning Theory},
  pages={138--158},
  year={2019},
  organization={PMLR}
}

@inproceedings{chen2019new,
  title={A new algorithm for non-stationary contextual bandits: Efficient, optimal and parameter-free},
  author={Chen, Yifang and Lee, Chung-Wei and Luo, Haipeng and Wei, Chen-Yu},
  booktitle={Conference on Learning Theory},
  pages={696--726},
  year={2019},
  organization={PMLR}
}

@inproceedings{wei2021non,
  title={Non-stationary reinforcement learning without prior knowledge: An optimal black-box approach},
  author={Wei, Chen-Yu and Luo, Haipeng},
  booktitle={Conference on learning theory},
  pages={4300--4354},
  year={2021},
  organization={PMLR}
}

@inproceedings{cutkosky2020parameter,
  title={Parameter-free, dynamic, and strongly-adaptive online learning},
  author={Cutkosky, Ashok},
  booktitle={Proceedings of the 37th International Conference on Machine Learning},
  pages={2250--2259},
  year={2020},
  organization={PMLR}
}

@inproceedings{jin2020adversarial,
  title={Learning adversarial {M}arkov decision processes with bandit feedback and unknown transition},
  author={Jin, Chi and Jin, Tiancheng and Luo, Haipeng and Sra, Suvrit and Yu, Tiancheng},
  booktitle={Proceedings of the 37th International Conference on Machine Learning},
  pages={4860--4869},
  year={2020},
  organization={PMLR}
}

@inproceedings{zhao2022dynamic,
  title={Dynamic regret of online {M}arkov decision processes},
  author={Zhao, Peng and Li, Long-Fei and Zhou, Zhi-Hua},
  booktitle={Proceedings of the 39th International Conference on Machine Learning},
  pages={26865--26894},
  year={2022},
  organization={PMLR}
}

@inproceedings{lykouris2018stochastic,
  title={Stochastic bandits robust to adversarial corruptions},
  author={Lykouris, Thodoris and Mirrokni, Vahab S. and Paes Leme, Renato},
  booktitle={Proceedings of the 50th Annual ACM SIGACT Symposium on Theory of Computing},
  pages={114--122},
  year={2018},
  organization={ACM},
  doi={10.1145/3188745.3188918}
}

@inproceedings{gupta2019better,
  title={Better algorithms for stochastic bandits with adversarial corruptions},
  author={Gupta, Anupam and Koren, Tomer and Talwar, Kunal},
  booktitle={Proceedings of the Thirty-Second Conference on Learning Theory},
  pages={1562--1578},
  year={2019},
  organization={PMLR}
}

@inproceedings{maillard2014latent,
  title={Latent Bandits.},
  author={Maillard, Odalric-Ambrym and Mannor, Shie},
  booktitle={International Conference on Machine Learning},
  pages={136--144},
  year={2014},
  organization={PMLR}
}

@inproceedings{slivkins2008adapting,
  title={Adapting to a Changing Environment: the Brownian Restless Bandits.},
  author={Slivkins, Aleksandrs and Upfal, Eli},
  booktitle={COLT},
  pages={343--354},
  year={2008}
}

@inproceedings{combes2014unimodal,
  title={Unimodal bandits: Regret lower bounds and optimal algorithms},
  author={Combes, Richard and Proutiere, Alexandre},
  booktitle={International Conference on Machine Learning},
  pages={521--529},
  year={2014},
  organization={PMLR}
}

@article{abbasi2011improved,
  title={Improved algorithms for linear stochastic bandits},
  author={Abbasi-Yadkori, Yasin and P{\'a}l, D{\'a}vid and Szepesv{\'a}ri, Csaba},
  journal={Advances in neural information processing systems},
  volume={24},
  year={2011}
}

@inproceedings{li2010contextual,
  title={A contextual-bandit approach to personalized news article recommendation},
  author={Li, Lihong and Chu, Wei and Langford, John and Schapire, Robert E},
  booktitle={Proceedings of the 19th international conference on World wide web},
  pages={661--670},
  year={2010}
}

@inproceedings{seznec2020single,
  title={A single algorithm for both restless and rested rotting bandits},
  author={Seznec, Julien and Menard, Pierre and Lazaric, Alessandro and Valko, Michal},
  booktitle={International Conference on Artificial Intelligence and Statistics},
  pages={3784--3794},
  year={2020},
  organization={PMLR}
}

@inproceedings{bogunovic2021stochastic,
  title={Stochastic linear bandits robust to adversarial attacks},
  author={Bogunovic, Ilija and Losalka, Arpan and Krause, Andreas and Scarlett, Jonathan},
  booktitle={Proceedings of the 24th International Conference on Artificial Intelligence and Statistics},
  pages={991--999},
  year={2021},
  organization={PMLR}
}

@inproceedings{gentile2014online,
  title={Online clustering of bandits},
  author={Gentile, Claudio and Li, Shuai and Zappella, Giovanni},
  booktitle={International conference on machine learning},
  pages={757--765},
  year={2014},
  organization={PMLR}
}

@inproceedings{lykouris2021corruption,
  title={Corruption-robust exploration in episodic reinforcement learning},
  author={Lykouris, Thodoris and Simchowitz, Max and Slivkins, Alex and Sun, Wen},
  booktitle={Proceedings of the Thirty-Fourth Conference on Learning Theory},
  pages={3242--3245},
  year={2021},
  organization={PMLR}
}

@article{auer2002nonstochastic,
  title={The nonstochastic multiarmed bandit problem},
  author={Auer, Peter and Cesa-Bianchi, Nicolo and Freund, Yoav and Schapire, Robert E},
  journal={SIAM journal on computing},
  volume={32},
  number={1},
  pages={48--77},
  year={2002},
  publisher={SIAM}
}

@article{besbes2014stochastic,
  title={Stochastic multi-armed-bandit problem with non-stationary rewards},
  author={Besbes, Omar and Gur, Yonatan and Zeevi, Assaf},
  journal={Advances in neural information processing systems},
  volume={27},
  year={2014}
}

@inproceedings{luo2018efficient,
  title={Efficient contextual bandits in non-stationary worlds},
  author={Luo, Haipeng and Wei, Chen-Yu and Agarwal, Alekh and Langford, John},
  booktitle={Conference On Learning Theory},
  pages={1739--1776},
  year={2018},
  organization={PMLR}
}

@article{wang2020restless,
  title={Restless-UCB, an efficient and low-complexity algorithm for online restless bandits},
  author={Wang, Siwei and Huang, Longbo and Lui, John},
  journal={Advances in Neural Information Processing Systems},
  volume={33},
  pages={11878--11889},
  year={2020}
}

@InProceedings{Jiang2023online,
  title = 	 {Online Restless Bandits with Unobserved States},
  author =       {Jiang, Bowen and Jiang, Bo and Li, Jian and Lin, Tao and Wang, Xinbing and Zhou, Chenghu},
  booktitle = 	 {Proceedings of the 40th International Conference on Machine Learning},
  pages = 	 {15041--15066},
  year = 	 {2023},
  editor = 	 {Krause, Andreas and Brunskill, Emma and Cho, Kyunghyun and Engelhardt, Barbara and Sabato, Sivan and Scarlett, Jonathan},
  volume = 	 {202},
  series = 	 {Proceedings of Machine Learning Research},
  month = 	 {23--29 Jul},
  publisher =    {PMLR},
}

@inproceedings{DBLP:conf/icml/RakhlinS16,
  author       = {Alexander Rakhlin and
                  Karthik Sridharan},
  editor       = {Maria{-}Florina Balcan and
                  Kilian Q. Weinberger},
  title        = {{BISTRO:} An Efficient Relaxation-Based Method for Contextual Bandits},
  booktitle    = {Proceedings of the 33nd International Conference on Machine Learning,
                  {ICML} 2016, New York City, NY, USA, June 19-24, 2016},
  series       = {{JMLR} Workshop and Conference Proceedings},
  pages        = {1977--1985},
  publisher    = {JMLR.org},
  year         = {2016},
  url          = {http://proceedings.mlr.press/v48/rakhlin16.html},
  bibsource    = {dblp computer science bibliography, https://dblp.org}
}

@inproceedings{DBLP:conf/nips/RussacVC19,
  author       = {Yoan Russac and
                  Claire Vernade and
                  Olivier Capp{\'{e}}},
  title        = {Weighted Linear Bandits for Non-Stationary Environments},
  booktitle    = {Advances in Neural Information Processing Systems},
  volume       = {32},
  pages        = {12017--12026},
  year         = {2019},
  url          = {https://proceedings.neurips.cc/paper/2019/hash/263fc48aae39f219b4c71d9d4bb4aed2-Abstract.html},
  bibsource    = {dblp computer science bibliography, https://dblp.org}
}
